\documentclass[10pt,twocolumn,letterpaper]{article}

\usepackage{iccv}
\usepackage{times}
\usepackage{epsfig}
\usepackage{graphicx}
\usepackage{amsmath}
\usepackage{amssymb}
\usepackage{float}
\usepackage{flushend}
\usepackage{arydshln}

\usepackage{caption,subcaption}

\def\minwrt[#1]{\underset{#1}{\text{minimize }}}
\def\maxwrt[#1]{\underset{#1}{\text{maximize }}}
\def\infwrt[#1]{\underset{#1}{\text{inf }}}
\def\argmin[#1]{\underset{#1}{\text{arg min }}}
\def\argmax[#1]{\underset{#1}{\text{arg max }}}

\newcommand{\bJ}{\mathbf{J}}

\newcommand{\norm}[1]{\left\lVert #1 \right\rVert}

\def\bx{{\bf x}}

\def\bz{{\bf z}}
\def\bx{{\bf x}}

\def\bgamma{{\boldsymbol \gamma}}
\def\bI{{\bf I}}
\newcommand{\barray}{\begin{array}}
\newcommand{\earray}{\end{array}}
\newcommand{\bea}{\begin{eqnarray}}
\newcommand{\eea}{\end{eqnarray}}

\def\bz{{\bf z}}

\newcommand{\bM}{\mathbf{M}}

\usepackage[pagebackref=true,breaklinks=true,letterpaper=true,colorlinks,bookmarks=false]{hyperref}

 \iccvfinalcopy % *** Uncomment this line for the final submission

\def\iccvPaperID{} % *** Enter the ICCV Paper ID here

\ificcvfinal\fi

\begin{document}

%%%%%%%%% TITLE
%\title{Latent Space Exploration Across Mismatched Manifolds \\ in Deep Generative Modeling}
\title{Discriminating Against Unrealistic Interpolations\\ in Generative Adversarial Networks}

\author{Henning Petzka\thanks{Equal contribution}\\
Lund University\\
Sweden\\
{\tt\small henning.petzka@math.lth.se}
\and
Ted Kronvall$^*$\\
Lund University\\
Sweden\\
{\tt\small ted.kronvall@math.lth.se}
\and
Cristian Sminchisescu\\
Lund University\\
Sweden\\
{\tt\small cristian.sminchisescu@math.lth.se}
}

\maketitle
\ificcvfinal\thispagestyle{empty}\fi

%%%%%%%%% ABSTRACT
\begin{abstract}
Interpolations in the latent space of deep generative models is one of the standard tools to synthesize semantically meaningful mixtures of generated samples. As the generator function is non-linear, commonly used linear interpolations in the latent space do not yield the shortest paths in the sample space, resulting in non-smooth interpolations. Recent work has therefore equipped the latent space with a suitable metric to enforce shortest paths on the manifold of generated samples. These are often, however, susceptible of veering away from the manifold of real samples, resulting in smooth but unrealistic generation that requires an additional method to assess the sample quality along paths. Generative Adversarial Networks (GANs), by construction, measure the sample quality using its discriminator network. In this paper, we establish that the discriminator can be used effectively to avoid regions of low sample quality along shortest paths. By reusing the discriminator network to modify the metric on the latent space, we propose a lightweight solution for improved interpolations in pre-trained GANs.

\end{abstract}

%%%%%%%%% BODY TEXT
%------------------------------------------------------------------------

\section{Introduction}

    Deep generative models, a class of deep learning methods used to synthesize data representative of some observed but typically unknown data distribution, have in recent years seen a terrific increase in active research. As a consequence, some of these methods can now achieve truly stunning results, e.g., very realistic images depicting people or objects which don't exist in real life. 
    %Deep generative models, a class of deep learning methods used to synthesize data representative of some observed but typically unsupervised dataset, has in recent years seen a terrific increase in active research. As a consequence, some of these methods can now achieve truly stunning results, e.g., very realistic images depicting people or objects which don't exist in real life. 
    %
    At the forefront among these are generative adversarial networks (GANs), a quickly expanding subclass of deep generative models following the seminal work by Goodfellow \etal \cite{GoodfellowPMXWOCB14}, which are distinguished by their ability to not only learn to synthesize data, but also learn to assess how representative of the dataset the synthetic data is. These two learning processes are pitched against each other in an adversarial game, with the goal of the former being able to synthesize data so well that the latter can't tell it apart from real data.
%------------------------------
\begin{figure}[t] 
\centering
\includegraphics[width=0.45\textwidth]{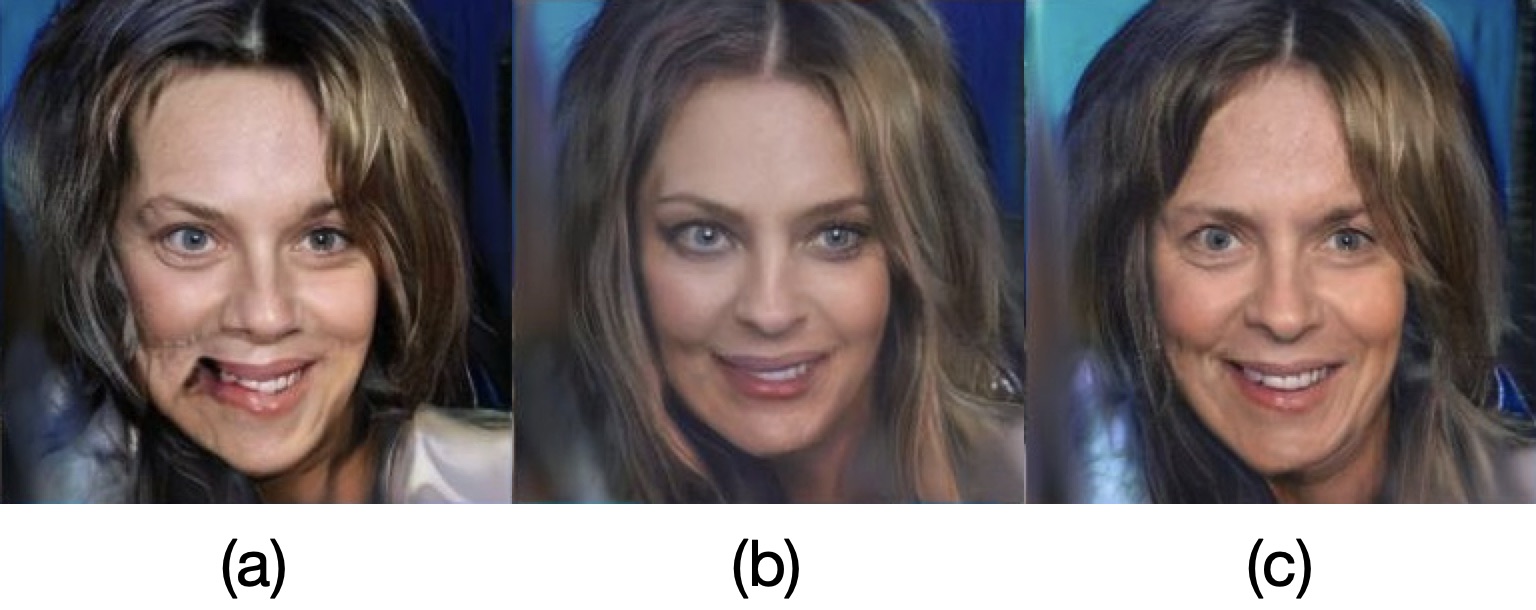} \caption{\label{fig:interpolation_point_example}Single points on interpolated paths for ProGAN by Karras \etal \cite{KarrasALL18} on CelebA-HQ using (a) Linear, (b) shortest path and (c) our approach. Full paths in Figure~\ref{fig:CelebAlinearfails}.
} \vspace{-3mm}
\end{figure}
%------------------------------

    For
    many generative models, including GANs,
    %use an input from some lower-dimensional latent space to produce a sample in the high-dimensional sample space. This 
    the (so called) generator can be seen as producing samples on a lower dimensional manifold immersed in the high dimensional sample space.
    %can be seen as the (so called) generator producing points on a lower dimensional manifold immersed in the high dimensional sample space. %which ideally exactly includes the data points representative of the dataset, and nothing else.
    %
    %It is assumed throughout this paper 
    Similarly, we herein assume that the set of real data points, of which the dataset is a representative example, form a lower dimensional manifold immersed in sample space, however unknown. 
    %lower-dimensional submanifold of the sample space 
    The %fully trained 
    %generative model should thus not only parametrize 
    goal of the generator is thereby not only to produce samples 
    %a function which outputs coincide 
    coinciding with the observed dataset, but to achieve perfect alignment between the generator and real data manifold.%; nothing less, nothing more. 
    %it should also be able to synthesize the infinitely many unobserved points on the real data manifold, and nothing else. 
    
    Due to factors such as finite-size datasets, suboptimal training, choice of network architecture and its capacity, there will in general exist some degree of mismatch between these manifolds. %, see Figure \ref{fig:manifold} for an illustration. %generator image and the true data manifold.
    One type of mismatch is commonly called mode collapse; when all points in latent space map to either a single, or too few, points in sample space, %the generator fails completely is %happens 
    %when all points in the latent space are mapped to a single point in sample space, called complete mode collapse, %, typically called (complete) mode collapse, 
    such that the generator's output is apparently less rich than the real data manifold. In this work, we focus on another type of mismatch; when (the well-trained) generator %GAN
    occasionally produces unrealistic outputs, i.e., samples from outside of the real data manifold, as exemplified in Figure \ref{fig:interpolation_point_example}(a). This mismatch may be less obvious during training, but can be detrimental for interpolations in the latent space; a common post-processing tool for generating semantically meaningful data. E.g.,
    %
    %The consideration of the generator as a map between manifolds becomes important when regarding
    using two latent points %in the latent space 
    known to produce data with certain semantic features, one may use convex combinations (linear interpolations) of these to synthesize data where these features are mixed, see, e.g., Karras \etal \cite{KarrasALL18} and Chen \etal \cite{ChenCDHSSA16}.
    Similarly, %Another approach used to 
    %modify features using linear interpolations 
    %to that end was examined by, e.g., 
    %is to first, post-training, classify a large sample of outputs from generator into different clusters. One may then enhance certain features in the output by linearly extrapolating in the latent space towards the centroids of these feature clusters, see, e.g., 
    Bengio \etal \cite{BengioMDR13} and Radford \etal \cite{RadfordMC15}, interpolate linearly between points and  feature-dependent clusters identified post training.

    If the latent space is equipped with a Euclidean metric, a linear interpolation defines a shortest path, i.e., a geodesic. The image of this path under the generator will however not, in general, be a shortest path on its manifold in sample space.  One may instead equip the generator's output space with a sensible and interpretable metric and pull this back to the latent space, thereby defining a Riemannian metric (see Arvanitidis \cite{ArvanitidisHH18} and Chen \cite{ChenKKJBS18}).  Searching for shortest paths in the latent space with respect to this newly defined metric will then %, by definition, 
    lead to shortest paths on the manifold of the generator, 
    %
    %Interpolation using convex combinations of two latent points defines a shortest path, or geodesic, if the latent space is equipped with the Euclidean metric. The image of this path under the generator, however, will in general not result in a %shortest path
    %geodesic on the image manifold \textit{Remove this?: considered as a submanifold of its ambient space}. As a result, one may instead equip the data space with a sensible and interpretable metric and pull this back to the latent space, similar to Arvanitidis \cite{ArvanitidisHH18} and Chen \cite{ChenKKJBS18}. Searching for shortest paths in the latent space with respect to this newly defined metric will, by definition, lead to shortest paths on the image manifold of the generator. 
    %
    %Using a suitable Riemannian metric as above, the image path follows a %shortest path 
    %geodesic on the image manifold of the generator, 
    but not necessarily on the true data manifold; causing the synthesis of unrealistic data, as illustrated in Figure \ref{fig:interpolation_point_example}(b), where the shortest path becomes blurry.
    To that end, this paper investigates
    % outlines
     a general yet simple approach to modify the Riemannian metric so that the mismatch between the generator and the true data manifold is taken into account, and a shortest path will only traverse regions of high data fidelity. For GANs, we show that a well-trained discriminator has the ability to provide gradient information between real and generated samples, and may thus be used as the modifier of the metric. Formulated as a machine learning method, the proposed method creates short and realistic interpolations between points in the latent space, see Figure \ref{fig:interpolation_point_example}(c).
    
    %A specific choice of a Riemannian metric on the image manifold of a generator coming from a GAN leads to an approach, formulated as a machine learning method, to interpolate between latent points realizing short paths on the image manifold of high quality.
   
%------------------------------
\begin{figure}[t!]
\centering
\begin{subfigure}{.22\textwidth} %.233
  \centering
  \includegraphics[width=\linewidth]{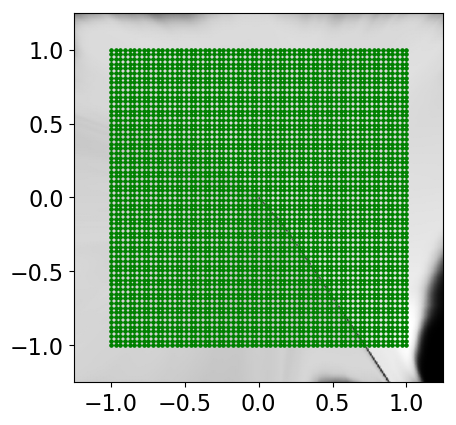}
  %\caption{}
  %\label{}
\end{subfigure}%
\begin{subfigure}{.25\textwidth} % 0.267
  \centering
  \includegraphics[clip, width=\linewidth]{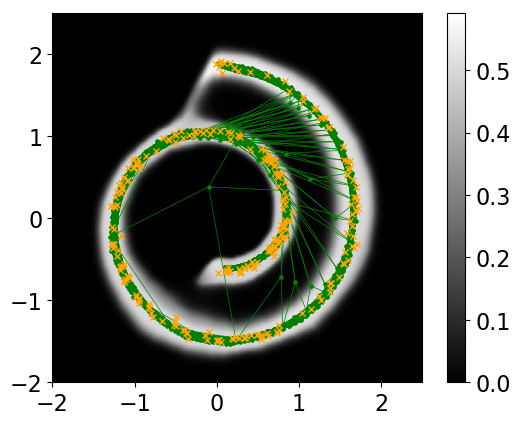}
  %\caption{}
  %\label{}
\end{subfigure} 
\caption{\label{fig:swiss_roll_manifold}Latent space $Z$ (left) and the generator manifold $G(Z)$ (right) for the Swiss roll using a discrete grid of $64^2$ points. Orange $X$s and grayscale background denote real data points and discriminator values, respectively.
} 
\label{fig:Swiss_roll_manifold}
\vspace{-3mm}
\end{figure}
%------------------------------
\label{page:swiss_roll_manifold}

   As a proof of concept, %we show 
   a toy example shows that mismatch of image and true data manifold already appears for simple problems; a GAN trained on the two-dimensional Swiss roll illustrates how the generator might learn a manifold larger than the data manifold (see Figure \ref{fig:swiss_roll_manifold}).  Shortest paths on the image manifold thus fall off the Swiss roll, while the geodesics from our modified metric don't. %focuses the curves to stay on and 
   %follow the Swiss roll. 
   The proposed method is also implemented for ProGAN (by Karras \etal \cite{KarrasALL18}) trained on the CelebA dataset (Liu \etal \cite{LiuLWT15}), illustrating 
   its {ability} to outperform
   %effectiveness  in outperforming 
   both linear interpolations and shortest paths by producing realistic interpolations on the generator manifold.
   %how the realisticness of the interpolations is improved using the proposed metric, compared to both the linear interpolations in the latent spaces as well as shortest paths on the image manifold.
   
%---------------------------------------------------------------------
\section{Related work}

    %Outline:
    %\begin{itemize}
    %\item Riemannian metric, and approximations, Arvanitidis, Shao, Chen.
    %\item Graph approximation, Chen
    %\item Statistics, Kuhnel
    %\item Adaptions, re-training the networks for adapting to the metric, Berthelot, Karras StyleGAN2, more?
    %\item Adapting the metric for realisticness, Hauberg    
   %\end{itemize}
    
   % With regards to the usefulness of interpolation schemes in deep generative modeling, there have been some recent contributions on the geometric interpretation of the latent space. In broad terms, it is established that a proper way to measure distances in the latent space is by pulling back the metric from the sample space, which is typically assumed to be Euclidean on the sub-manifold where all representative sample points lie.
    %
    It has been shown in a number of recent works that the proper way to measure distances in the latent space is to endow it with a Riemannian metric.
    Arvanitidis \etal \cite{ArvanitidisHH18} calculate shortest paths by solving an ordinary differential equation (ODE), applied to a variational autoencoder (or VAE, see \cite{KingmaW13}) to account for the curvature of the latent space. Subsequently, Yang \etal \cite{YangAFLH18_unpublished} show that a shortest path may also be found using a discrete analogue of the ODE in finite many sample points along the path. Both Shao \etal \cite{Shao0F18} and Chen \etal \etal \cite{ChenKKJBS18} show similar results using discrete sample points. The latter work shows that linear interpolations are still serviceable, and hypothesises that the reason is that the curvature of the Riemannian metric is typically quite flat already, i.e., close to Euclidean. In subsequent work, K\"{u}hnel \etal \cite{KuhnelFJS18_unpublished} develop a framework for performing statistics (such as calculating the mean value) in a non-Euclidean latent space. 
    
    A number of works attempt to flatten or address the curvature of the latent space during training of the generator, thus improving linear interpolations. In the work by Berthelot \etal \cite{BerthelotRRG19}, the performance issues of autoencoders (see \cite{RumelhartHW86}) are improved by an additional network measuring image quality that is used to regularize the training of the autoencoder. 
    Similarly, for VAEs, Chen \etal \cite{ChenKFBP20_unpublished} proposed to flatten the generator manifold by replacing the VAEs standard Gaussian prior with a customized type of hierarchical prior.
    Using a different approach from the same authors, Chen \etal  \cite{ChenFKPBS19}, approximate the generator manifold using a finite graph of generated samples and then use classical search algorithms to find the shortest paths.
     In a recent work, Stolberg-Larsen \etal \cite{StolbergLarsenS21} consider interpolations over partially disjoint generator manifolds, which could be used for datasets containing (partially disjoint) subclasses of data, for which Euclidean distances are ill-defined. Using a novel architecture of multiple generators, interpolations are done by matching the generator output using finite sample graphs.
    In two recent papers by Karras \etal \cite{KarrasLA18_unpublished,KarrasLAHLA20}, the authors train a GAN  
%of unconventional architecture 
{using a generator architecture similar to networks performing style transfer} to smoothly mix features between the images of faces or other objects using an intermediate latent space instead of the conventional latent space. To enforce smooth and realistic interpolations, the curvature of the intermediate latent space is regularized during training, thereby encouraging linear interpolations to be shortest paths. {In this work, we will only consider traditional GAN architectures.}

    Two issues with the approaches above arise: Firstly, constraining the curvature might impede the quality of image generation, and secondly, it would be desirable to separate path finding from network training, as one may then reuse pre-trained networks. %, and it would be desirable to separate these two processes to find interpolating curves for any existing generator network.
    %
    %The issue with many of the works above, where new network structures are being trained, is that existing networks known to produce mostly realistic results would go unused. 
    In a paper by Laine \cite{Laine18}, shortest paths are found by pulling back the Euclidean metric on intermediary feature spaces in the VGG19 classification network \cite{Simonyan15}, whose representations have been shown to correlate well with human perception \cite{zhang2018unreasonable}. Similarly, a recent work by Arvanitidis \cite{ArvanitidisHS20_unpublished} 
    studies suitable metrics on the sample space for defining the Riemannian pull-back metric. While their method is targeted toward probabilistic generators, they also consider how certain regions (in their example, blond people) can be effectively avoided in interpolating curves by assigning 
    high cost to those regions. %the use of a suitable metric on the sample space that exhibits high cost around the non-desired regions. 
    While our work was derived independently from their results, it may be considered as a continuation of their work, showing that the existing discriminator of a GAN is a suitable, yet simple ingredient to avoiding %(unlabeled) 
    regions of unrealistic samples in interpolations.

%------------------------------------------------------------------------
\section{Background}
    The mathematical convention around which deep generative modeling is built stems from statistical inference; the dataset serving as inspiration for the data synthesis is assumed to be a set of samples from some data distribution, i.e., $\bx_i \sim p_X, i = 1,\dots,N,$  and the generative process attempts to draw new samples $\bx$ from this distribution. 
  Since the data distribution is generally intractable to model directly, the approach used in deep generative modeling is to construct a method for drawing samples from a conditional probability $p_{X|Z}(\bx | \bz)$, where a latent space $Z$ is equipped with a probability distribution $p_Z(\bz)$ that is simple to sample from. The method for sampling from $p_{X|Z}$ exploits the immense capabilities of deep neural networks to approximate $p_{X|Z}$ with help of a generator function $G: Z \rightarrow X$. While the generator $G$ may be probabilistic for some generative models, for GANs, among others, the generator is constructed using a deterministic feed-forward network resulting in a degenerate conditional probability distribution 
$p_G(\bx | \bz) = \delta \left(\bx - G(\bz)\right)$.  Using the law of total probability, we have
    \begin{align}
         p_G(\bx) = \int_{\bz\in Z} p_G(\bx|\bz) p_Z(\bz) d\bz,
        \label{eq:marginal}
    \end{align}
and it is apparent that for the desired simple latent distributions $p_Z$, the conditional distribution needs to compensate by means of complexity. 

Assuming a smooth generator $G:Z\subseteq \mathbb{R}^d\rightarrow X=\mathbb{R}^D$, $d<D$, the image of $G$ is a $d$-dimensional immersed manifold in the $D$-dimensional sample space, which means that it is locally homeomorphic to a $d$-dimensional Euclidean space, but might intersect itself globally. To find the most direct path on this manifold between two given samples, it must be clarified how to measure distances in the latent space in a meaningful way. In many applications, a sensible metric can be defined on the sample space instead. For images, for example, one may argue that the Euclidean distance over pixel values makes up a meaningful, yet simple, metric. Alternatively, we can measure the distance between images using the Euclidean metric in layers of a network pre-trained on image data (e.g., the VGG19 network \cite{Simonyan15}). % \cite{Simonyan15}, whose representations have been shown to correlate well with human perception \cite{zhang2018unreasonable}. 
Such metric on the sample space may then be pulled back to the latent space, so that the distance between two points in the latent space is defined by the distance between their corresponding points in the sample space along the generator manifold (see, e.g., \cite{ArvanitidisHH18, ArvanitidisHS20_unpublished, Laine18}). To that end, let $\boldsymbol{\gamma}: [0,1] \rightarrow Z$ be a smooth curve between $\bgamma(a)$ and $\bgamma(b)$ in the latent space, which maps to a corresponding curve $G(\boldsymbol{\bgamma}) =(G\circ\bgamma) : [0, 1] \rightarrow X$  on the generator manifold. Let $H:X\rightarrow Y$ denote a possible map from $X$ into another space where Euclidean distance % measuring distance in the Euclidean sense 
aligns better with human perception of distance. The length of $\boldsymbol{\bgamma}$ in latent space $Z$ is measured by 
    % Furthermore, from the modeling assumptions, there is no meaningful way to measure distances in the latent space they give no indication of whether it is Eucliean. Instead, one may pull the distances from the generator manifold back into the latent space, such that the distance between two points in the latent space appropriates the distance between the corresponding points in the data space along the generator manifold. To that end, let $\boldsymbol{\gamma}: [a,b] \rightarrow Z$ be a continuous curve between $a$ and $b$ in the latent space, for which exists a corresponding curve on the generator manifold, $G(\boldsymbol{\gamma}): [G(a), G(b)] \rightarrow X$. The length of $\boldsymbol{\gamma}$ is
    \begin{align}
        L(\boldsymbol{\gamma}) &= \int_{0}^1 \norm{\ \dot{\boldsymbol{\gamma}}(t)}_R dt= \int_{0}^1 \norm{\ \frac{\partial}{\partial t}{( H\circ G \circ \boldsymbol{\gamma}})(t)}_2 dt,
        \label{eq:curve}
    \end{align}
where $\norm{\ \dot{\boldsymbol{\gamma}}(t)}_R dt$ denotes the length of an infinitesmal curve element at $t \in [0,1]$ with respect to the Riemannian metric defined by the pullback metric of the Euclidean metric (see, e.g., DoCarmo \cite{doCarmo92} and Arvanitidis \cite{ArvanitidisHS20_unpublished}). This Riemannian metric is defined by the matrix $\bM_{\boldsymbol{\gamma}(t)}= {\bJ_{\boldsymbol{\gamma}(t)}}^\top \bJ_{\boldsymbol{\gamma}(t)}$, where $\bJ_{\boldsymbol{\gamma}(t)}$ denotes the first-order derivatives (Jacobian) of  $(H\circ G)$ at $\boldsymbol{\gamma}(t) \in Z$, and
        \begin{align}
        \norm{\ \dot{\boldsymbol{\gamma}}(t)}_R dt
        &= \norm{\ \frac{\partial}{\partial t}{( H\circ G \circ \boldsymbol{\gamma}})(t)}_2 dt\\
        &= \norm{\bJ_{\boldsymbol{\gamma}(t)} \ \dot{\boldsymbol{\gamma}}(t)}_2\, dt\\
      %  &= \sqrt{\ \dot{\boldsymbol{\gamma}}(t)^\top \bJ_{\boldsymbol{\gamma}(t)}^\top \bJ_{\boldsymbol{\gamma}(t)} \ \dot{\boldsymbol{\gamma}}(t)}\, dt\\
        &= \sqrt{\ \dot{\boldsymbol{\gamma}}(t)^\top \bM_{\boldsymbol{\gamma}(t)} \ \dot{\boldsymbol{\gamma}}(t)}\,dt.
    \end{align}
%Similarly, measuring distances between images in a layer $\ell$ of the VGG network, $\tilde \bM_{\gamma(t)}=\tilde J_{\gamma(t)}^T \tilde J_{\gamma(t)}$ where $\tilde J_{\gamma(t)}$ denotes the Jacobian of the composition of $\ell$ and $G$.
%
%\textcolor{red}{
For Euclidean distances along the generator manifold, $H$ simply denotes the identity function, while for distances in VGG network layers, $H$ denotes its corresponding feature maps.
%}
%
A shortest path, called geodesic, may thus be found by minimizing \eqref{eq:curve} with respect to $\boldsymbol{\gamma}(t)$ for fixed endpoints $\bgamma(0),\bgamma(1)$. 
     To also enforce constant speed for a parametrization of the geodesic, % i.e., $\dot{\boldsymbol{\gamma}(t)}$ being almost constant over $t$, 
     one may instead minimize the corresponding energy functional (see \cite[p.182]{Petersen16}), i.e., the integral over the square of the curve lengths,
    \begin{align}
    E(\boldsymbol{\gamma}) = \int_{t=a}^b \ \dot{\boldsymbol{\gamma}}(t)^\top \bM_{\boldsymbol{\gamma}(t)} \ \dot{\boldsymbol{\gamma}}(t)\ dt. \label{eq:energy}
    \end{align}
    Both \eqref{eq:curve} and \eqref{eq:energy} may be minimized numerically by solving a set of ordinary differential equations, albeit at a significant computational cost, as this involves calculating the second-order derivative of the generator (see Shao \etal \cite{Shao0F18})

    In this work, we are interested in finding geodesics on the true data distribution, but in the image of a generator of a GAN. A GAN consists of a generator function $G$ together with a %so-called 
    discriminator function $D: X \rightarrow \mathbb{R}$ which is trained to return high values for the samples in (and representative of) the dataset, i.e., $\bx_i \in X$, for $i = 1,\dots,N$, and low values otherwise. Introduced in 2014 by Goodfellow \etal \cite{GoodfellowPMXWOCB14}, GANs have been one of the most successful methods of generative modeling with numerous applications. For a comprehensive overview on their architecture and how to train them, we refer the reader to \cite{WangZSW21}, and the references therein. A variant of the original GAN based on the Wasserstein distance between the data distribution $p_X$ and its approximation $p_G$, called WGAN-GP (see Arjovsky  \cite{arjovsky17} and Gulrajani \cite{gulrajani2017improved}), is of central interest to our work as it has been used to train highly competitive generative models on image data.

%------------------------------------------------------------------------
    %------------------------------------------------------------------------

%------------------------------
\begin{figure}[t!]
\centering
\includegraphics[width=0.35\textwidth]{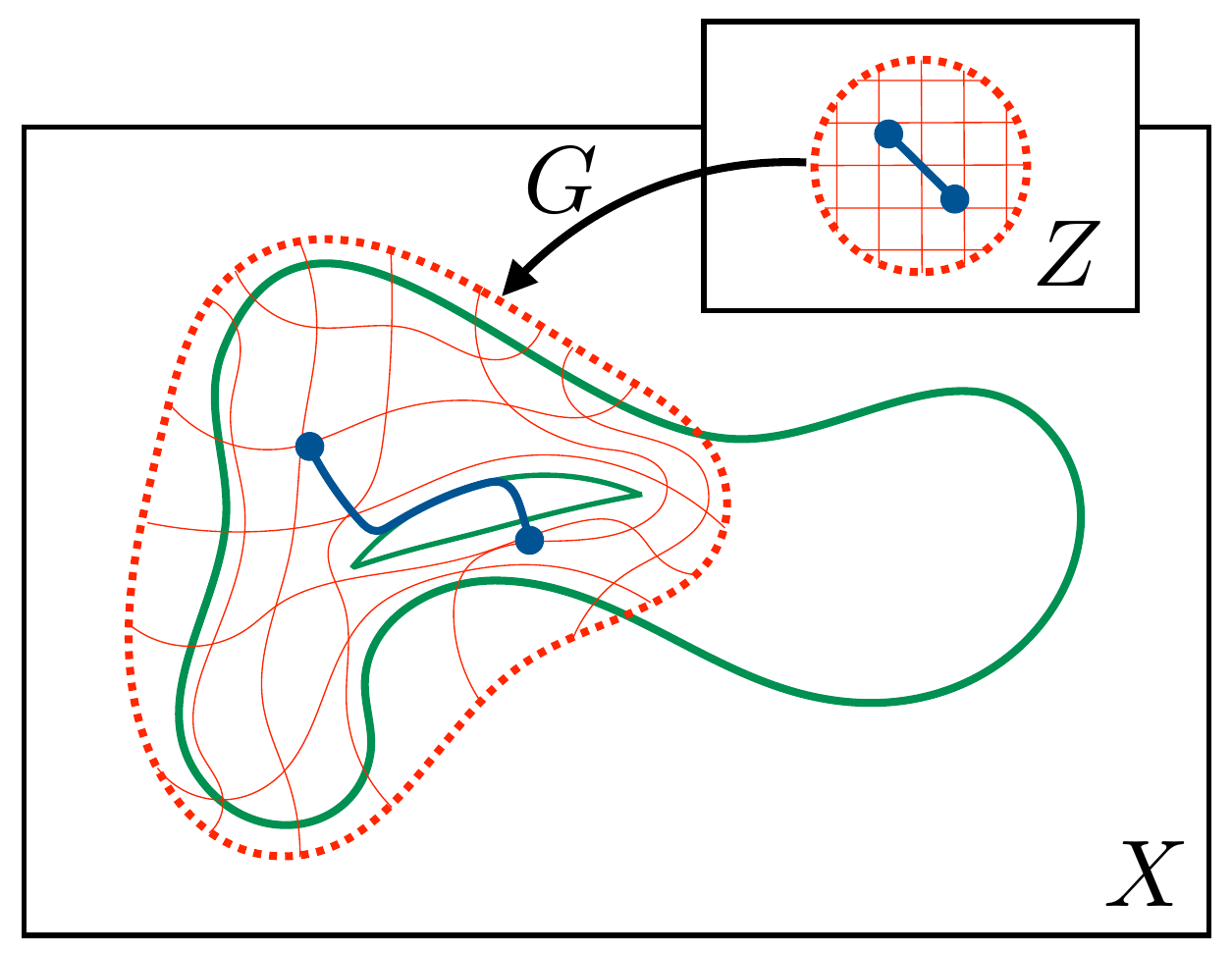}     \caption{\label{fig:manifold}Real data manifold (%the set bordered by the
green) vs. %a part of 
the generator manifold (%the set bordered by 
red dashed), where linear coordinates in $Z$ result in nonlinear coordinates in $X$. Blue line shows a linear interpolation in the latent space mapped to the
%, and the corresponding path in 
sample space. %, which in general is not the shortest path on the data manifold.
} \end{figure}
%------------------------------

\section{An adaptive metric for generative \\adversarial networks}\label{sct:adaptiveMethod}

        %Examining the formulation of the data distribution $p_X$ as marginal in \eqref{eq:marginal}, there are two terms factors to take into consideration; there is the versatility of the latent distribution $p_Z$, versus how well one may sample from the conditional distribution $p_{X|Z}$. In deep generative modeling, $p_Z$ is selected by the user, while the sampling is done using the trained generator $G$, such that $p_G(\bx) \approx p_X(\bx)$. Assume that there is no mismatch between the generator image and the data manifold, implying that $p_G(\bx) = p_X(\bx), \forall \bx \in X$.  
The goal of generative modeling is to match the generated distribution $p_G$ in \eqref{eq:marginal} with the data distribution $p_X$. Assuming that there is no mismatch between the generator image and the data manifold, it must hold for the deterministic GAN approach that,
    \begin{align}\label{eq:mismatch1}
    %\begin{split}
        p_X(\bx) &= p_G(\bx) \\ 
        &= \int_{\bz \in Z} \delta(\bx - G(\bz)) p_Z(\bz) d\bz \\ 
       % &= \int_{\bz \in Z} \delta(G^{-1}(\bx) - \bz) p_Z(\bz) d\bz \\ 
        &= p_Z(G^{-1}(\bx)),
    %    \end{split}
    \label{eq:mismatch4}
    \end{align}
    for the pre-image of $G$, i.e., $G^{-1}(\bx)=\{\bz:G(\bz)=\bx \}$. Since the generator $G$ is modeled as a (continuous) feed-forward network with a pre-determined architecture and a simplistic latent distribution where typically $d < D$, it cannot be guaranteed to find network parameters such that  equations~\eqref{eq:mismatch1}--\eqref{eq:mismatch4} hold in general.
    %provided that the generator $G$ is invertible $\forall \bx \in X$, which is not true for $d<D$ in general.  
    %Thus, there may exist $\bx \in X$ which have no corresponding $\bz$ such that $G(\bx) = \bz$, subsequently causing mismatch between the generator image and the data manifold. 
    Also, since the data manifold is not known everywhere, but only in the sample points included in the finite-sized dataset, the (continuous) generator needs to extrapolate between the observed points on the data manifold; an estimation procedure that forms another source of manifold mismatch.
    One may distinguish between two types of mismatch; firstly when the generator synthesizes unrealistic data, then
    \begin{align}
        \exists \, \bz: G(\bz) \notin X,
    \end{align}
    and, secondly, when the the generator cannot reach all points on the data manifold (mode collapse), then
    \begin{align}
    \exists \, \bx: \{\bz: G(\bz) = \bx\} = \varnothing.
    \end{align}

    %
    %\textcolor{red}{
    An illustration of these two mismatches can be seen in Figure \ref{fig:manifold}, where unrealistic generation happens at the hole in the manifold and outside of the green set on the left, while mode collapse occurs outside of the red dashed set on the right.
    %}
    The mode collapse mismatch is a common issue for GANs; and many approaches exist for circumventing it, see, e.g., \cite{MeschederGN18} and the references therein for a good overview.  
    This paper instead focuses on the first type of mismatch, were the generator synthesizes samples which are not in the (unknown) data manifold, i.e., being unrealistic, and
    describes an approach for circumventing it. To that end, consider the adapted curve length 
    \begin{align}
        L^*(\boldsymbol{\gamma}) = \int_{t=a}^b \norm{\ \dot{\boldsymbol{\gamma}}(t)}_{R^*} dt, \label{eq:curvestar}
    \end{align}
   where the curve length is defined by a proposed metric, which is infinite outside the data manifold. For the Euclidean norm in sample space this is 
    \begin{align}
        \bM^*_{\bz} &= \left\{ \begin{array}{ll}
             \bJ_{\bz}^\top \bJ_{\bz}  & G(\bz) \in X \\
             \infty \cdot \bI & G(\bz) \notin X 
        \end{array} \right.,
    \end{align}
   such that the adapted norm is
    \begin{align}
        \norm{\dot{\bz}}_{R^*} &= \left\{ \begin{array}{ll}
             \norm{\bJ_{\bz} \dot{\bz}}_2 & G(\bz) \in X \\
             \infty & G(\bz) \notin X 
        \end{array} \right.,
    \end{align}
    or equivalently, 
    \begin{align}
        \norm{\dot{\bz}}_{R^*} &= \norm{\bJ_{\bz} \dot{\bz}}_2 + \phi(\bz) \\
        \phi(\bz) &= \left\{ \begin{array}{ll}
             0 & G(\bz) \in X \\
             \infty & G(\bz) \notin X 
        \end{array} \right..
    \end{align}
    In this formulation, $\phi: Z \rightarrow \mathbb{R}$ is a penalty function such that any curve traversing points outside the data manifold has infinite length, ruling these out as shortest paths. However, calculating $\norm{\cdot}_{R^*}$ requires knowledge of the data manifold beforehand, which is infeasible, and to that end, one may approximate $\phi$ using some auxiliary function. Depending on the deep generative modeling method, $\phi$ may be approximated in different ways, however always such that the function has small values for realistic data, and large values for unrealistic data. %In the case of VAE, one might thus use the reconstruction error, while in the case of FLOW, one might use the inverse of the exact likelihood function. 
    For GANs, a natural choice of auxiliary function would be based on the discriminator, $D: X \rightarrow \mathbb{R}$. In the origin work by Goodfellow \etal, \cite{GoodfellowPMXWOCB14} the discriminator maps each data sample into the interval $(0,1)$ and its image $D(\bx)$ on some $\bx\in X$ is interpreted as the probabilty of $\bx$ being a sample from the real data distribution $p_X$. So, to calculate shortest paths for a GAN, the auxiliary function may be approximated as
    \begin{align}\label{eq:phi}
        \phi(\bz) \approx \frac{\lambda}{D(G(\bz)) + \epsilon},      
    \end{align}
    where $\lambda > 0$ is a hyperparameter of the model, and $\epsilon~>~0$ a~ small number added to ensure numerical stability when $D(\bx)\approx 0$. 
    
    For computational reasons, the shortest path may be approximately calculated using a discretized version of $\bgamma(t)$. To that end, consider a sequence of $T+1$ points in the latent space along the curve $\bgamma$, i.e., $\bgamma(t_i), i = 0,\dots,T$, such that $\bgamma(t_0) = \bgamma(0)$ and $\bgamma(t_T) = \bgamma(1)$. In sample space, this sequence forms a corresponding discrete path $(G \circ \bgamma)(t_i), i = 0,\dots,T$, for which the discrete analogue of the energy functional for the proposed metric becomes 
    \begin{align}
        E^*_{t_i}(\bgamma) = \sum_{i=1}^{T} \frac{1}{T} & \norm{(G \circ \bgamma)(t_i) - (G \circ \bgamma)(t_{i-1})}_{R^*}^2,
    \end{align}
    which we approximate, using \eqref{eq:phi}, by
    \begin{align}\label{eq:numericalProblem}
        E^*_{t_i}(\bgamma) \approx \sum_{i=1}^{T} \frac{1}{T} &\bigg( \norm{(H\circ G \circ \bgamma)(t_i) - (H\circ G \circ \bgamma)(t_{i-1})}_2 \nonumber \\  
         &+ \lambda\Big((D \circ G \circ \bgamma)(t_i) + \epsilon \Big)^{-1} \bigg)^2.
    \end{align}
      
    In order to find a geodesic between $\bgamma(0)$ and $\bgamma(1)$, we follow the approach of Yang \etal  \cite{YangAFLH18_unpublished} to model the curve using a polynomial parametrization with trainable parameters $\psi$, for which the geodesic can be found by minimizing $E^*_{t_i}(\bgamma_\psi)$
    %\begin{align}
        %&
        %\underset{\psi}{\text{minimize}} \ %\ E^*_{t_i}(\bgamma_\psi)
    %\end{align}
    %$\text{minimize}\ E^*_{t_i}(\bgamma_\psi)$
    over $\psi$ 
via gradient descent or its variants applied on the discretized objective from \eqref{eq:numericalProblem}.\\
    
\begin{figure}[t]
\centering
  \includegraphics[width=.4\textwidth]{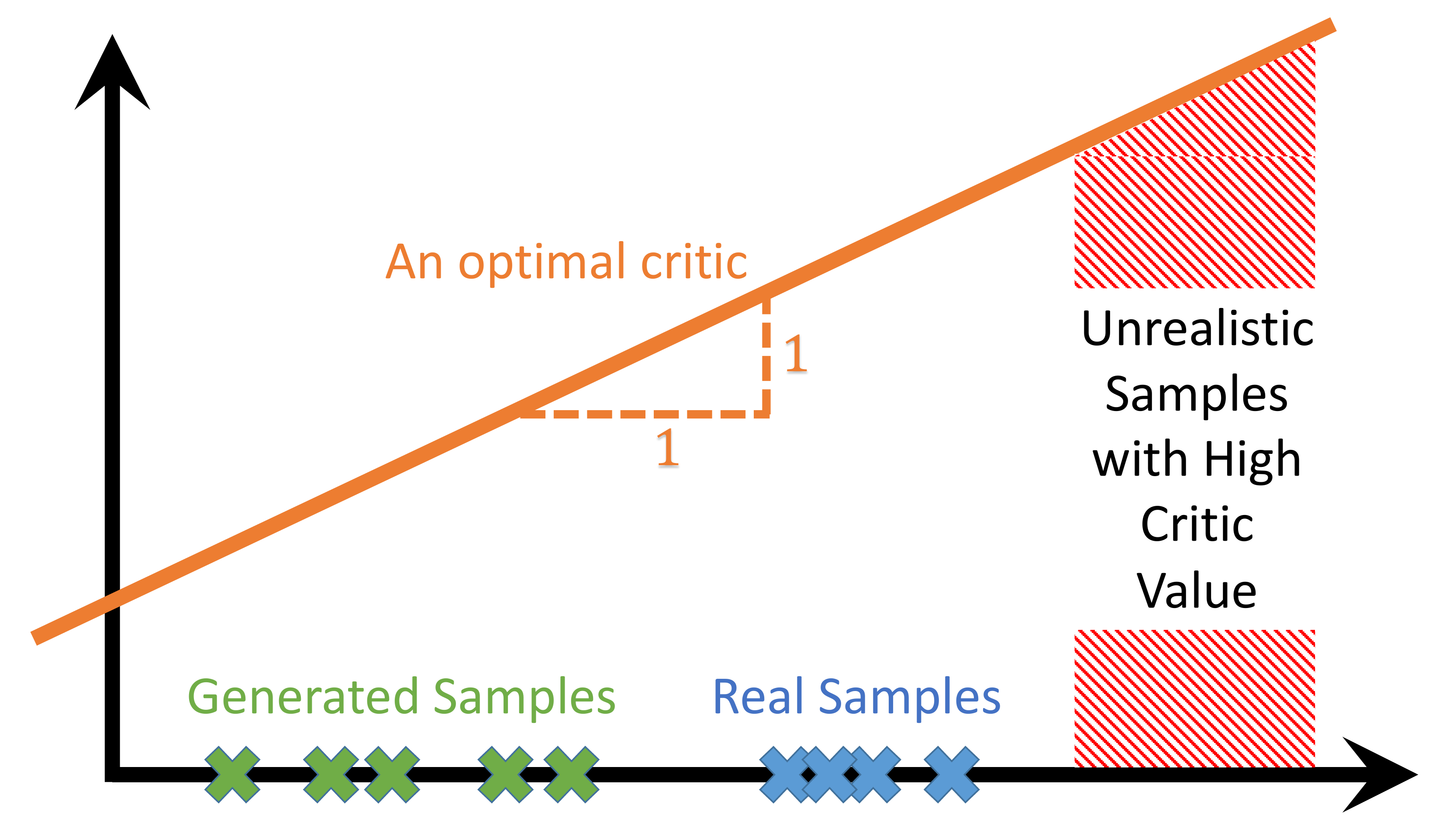}
  \caption{Values of optimal Wasserstein critics do not correspond to realisticness, but the gradient information is useful}
\label{fig:optimalCritic}
\vspace{-3mm}
\end{figure}

\textbf{Adapting our method to Wasserstein GANs.} In order to reuse pre-trained models of high quality generators, we {aim to} apply the proposed method to Wasserstein GANs, where the discriminator is usually called the critic; a terminology adopted herein. The change from discriminator to critic requires some adjustments of our method. A distinguishing difference of Wasserstein GANs to regular GANs is to replace the $(0,1)$-valued discriminator by one that can assign an arbitrary real number. The task of assigning high values to real and low values to generated samples remains unchanged, and a Lipschitz constraint on the discriminator prevents the divergence of their mean difference. With \eqref{eq:phi} relying on the assumption of a $(0,1)$-valued discriminator, but this not holding for the critic, we normalize critic values to $(0.1)$ using an approximate maximum and minimum critic value determined from generated samples. A more difficult issue is the lack of well-defined bounds and, more generally, that critic values cannot be interpreted as a probability of the sample being real. While the critic of a WGAN provides gradient information between real and generated samples toward improved equality, it is possible that non-realistic samples attain high critic values, for example if they fall outside the convex hull of real and generated samples. Figure~\ref{fig:optimalCritic} shows such an example of a real and generated data distribution and a theoretically optimal Wasserstein GAN critic, where the highest critic values are found off the real manifold. (The optimality of the displayed critic follows from \cite[Theorem 5.10]{villani2008optimal}, and Observation~4 in \cite{petzka2018regularization} provides another illustration where an optimal critic assigns a higher critic value to a generated sample than a real).  Therefore, {caution is advised when} 
%we have to be somewhat careful with 
enforcing high critic values, but since our main motivation is to improve the quality of images close to a shortest path, useful gradient information is all we need.

\section{Experiments}
Our experiments compare the following list of interpolation methods: \textbf{(a) Linear} - Linear interpolations in latent space, which are mapped by the generator $G$ onto the image manifold. \textbf{(b) sqDiff} - Shortest path on the image manifold measured with the Euclidean metric in sample space; i.e., the minimum of \eqref{eq:energy} with $H$ being the identity function. This minimizes squared differences of images under $G$ and is a method proposed in \cite{ArvanitidisHH18,ChenKKJBS18,Shao0F18}. \textbf{(c) sqDiff+D} - A herein proposed method using \eqref{eq:numericalProblem}, where the metric combines \textit{sqDiff} with enforcing high discriminator values. \textbf{(d) VGG} - Shortest path on the image manifold as measured by a weighted Euclidean distance between features in different layers of the VGG network (as detailed in \cite{Laine18}). \textbf{VGG+D} - A herein proposed method combining the method by Laine \cite{Laine18} with enforcing high discriminator values; i.e., \eqref{eq:numericalProblem} with $H$ being feature maps of the VGG network. \textbf{(f) Linear in sample space} - As a comparison to \textit{sqDiff} we show the shortest path in sample space with respect to its Euclidean metric, which is a straight line. This path is independent of the generator function, illustrating a completely naive interpolation. 
%\textbf{Disc} -A path on the image manifold that aims to maximize discriminator values along the path; 

%-------------------------------
\begin{figure}
\centering
\begin{subfigure}{.2\textwidth}
  \centering
  \includegraphics[width=\linewidth]{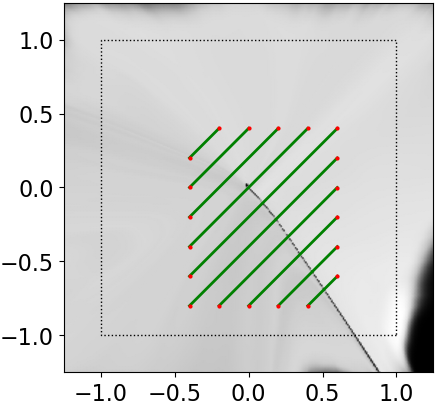}
  %\caption{}
  %\label{}
\end{subfigure}%
\begin{subfigure}{.23\textwidth}
  \centering
  \includegraphics[clip, width=\linewidth]{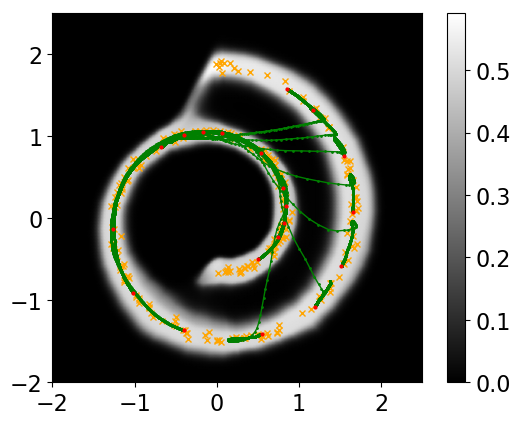}
  %\caption{}
  %\label{}
\end{subfigure} \\

\begin{subfigure}{.2\textwidth}
  \centering
  \includegraphics[width=\linewidth]{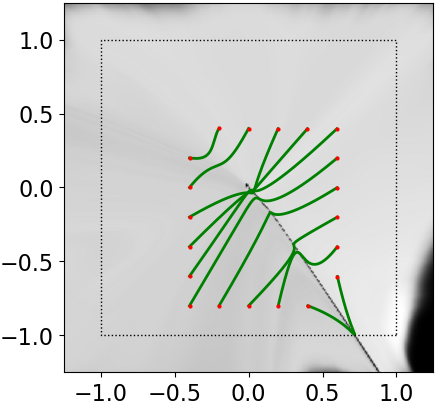}
  %\caption{}
  %\label{}
  \end{subfigure}%
\begin{subfigure}{.23\textwidth}
  \centering
  \includegraphics[width=\linewidth]{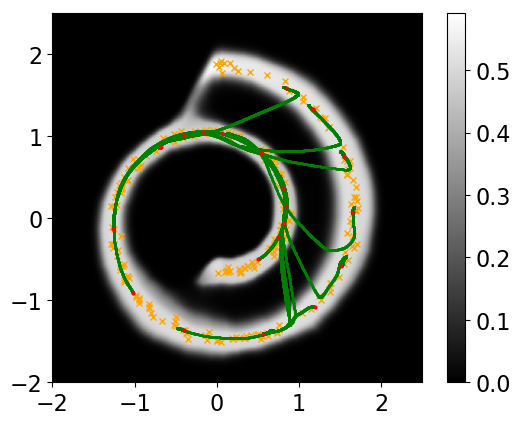}
  %\caption{}
  %\label{}
  \end{subfigure}
  
\begin{subfigure}{.2\textwidth}
  \centering
  \includegraphics[width=\linewidth]{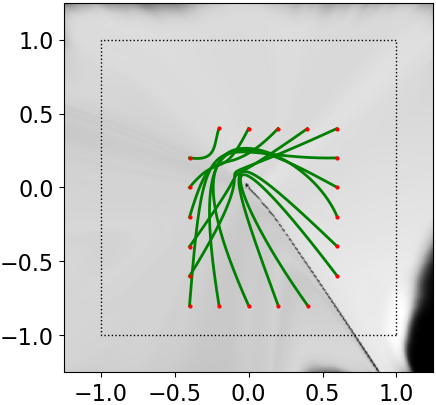}
   %\caption{}
  %\label{}
\end{subfigure}% 
\begin{subfigure}{.23\textwidth}
  \centering
  \includegraphics[width=\linewidth]{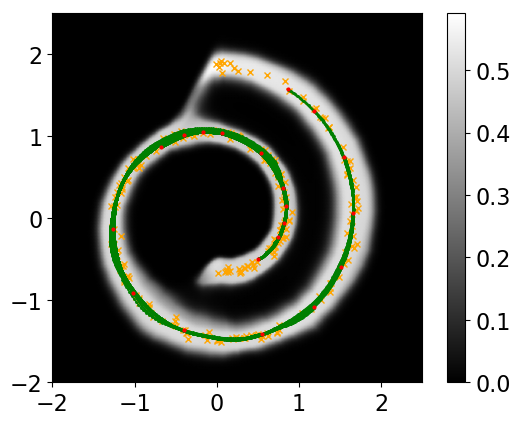}
   %\caption{}
  %\label{} 
\end{subfigure}
\caption{Comparison of interpolation paths (green) in latent space (left) sample space (right) for swiss roll; Top to bottom: (a) \textit{Linear}, (b) \textit{sqDiff}, and (c) \textit{sqDiff+D} (proposed).} 
\label{fig:Swiss_roll_paths}
\vspace{-3mm}
\end{figure}
%-------------------------------

For each of the methods \textit{sqDiff}, \textit{sqDiff+D}, \textit{VGG} and \textit{VGG+D}, we parametrize a path using a multi-dimensional polynomial with fixed endpoints by specifying the coefficients of the constant and the linear part. The coefficients to higher order terms are learned by minimizing the respective objectives. For a fair comparison, all methods are trained for the same number of epochs. We find that all methods benefit from "early stopping", i.e., not training until convergence. This is theoretically justified above for methods using a Wasserstein critic, but we also find that the bias toward blurry images of \textit{sqDiff} and \textit{VGG} (explained below) becomes more severe with more training iterations. We describe all hyperparameter settings and implementation details in Appendix~\ref{app:trainingSetup}. Our code is published under \url{https://github.com/petzkahe/progan_geodesics}.

\subsection{Swiss roll}
As a proof of principle, we consider the two-dimensional toy dataset of the so-called swiss roll. We trained a GAN with a two-dimensional latent space and visualized both generated samples (in green) and discriminator values (grayscale background) in Figure~\ref{fig:Swiss_roll_manifold} on page~\pageref{fig:swiss_roll_manifold}. 
By sampling from a dense grid in latent space, we can visualize how the GAN maps the two-dimensional latent space onto the swiss roll. At the end of training, generated samples stay on the swiss roll with high probability. % onto the swiss roll as desired. %, but some samples do fall off. 
Still, a low density of samples do fall off the swiss roll, implying large gradients of the generator at the corresponding latent points $\bz$, resulting in large metric $\bM_\bz$. Nonetheless, geodesics with respect to the method \textit{sqDiff} traverse this region of unrealistic samples if the resulting path is shorter in the sample space $\mathbb{R}^2$. Figure~\ref{fig:Swiss_roll_paths} compares paths obtained from \textit{Linear}, \textit{sqDiff} and \textit{sqDiff+D}. \textit{sqDiff} results in paths in sample space that are more smooth than the paths of \textit{Linear}, but, as predicted, will often fall off the swiss roll. 
Taking the discriminator information into consideration, our proposed method \textit{sqDiff+D} avoids regions of predicted fake samples and is therefore able to find smooth paths following the swiss roll.

%----------------------------
\begin{figure}[t]
  \centering
  \includegraphics[width=.45\textwidth]{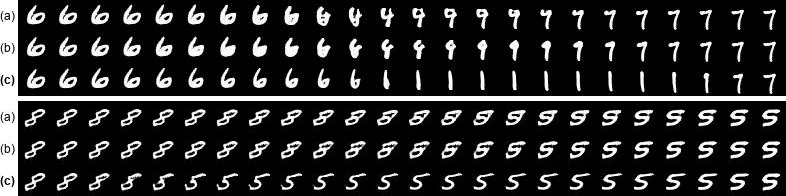}
    \caption{\textit(a) {Linear}, \textit{(b) sqDiff} and \textit{(c) sqDiff+D} (ours) on MNIST showing our method to find a realistic smooth path.}
\label{fig:mnist}
\vspace{-3mm}
\end{figure}
%----------------------------

\subsection{MNIST}
In a typical next step, we apply our method to the MNIST dataset \cite{LecunCB10}. Due to the discrete classes of digits, it seems debatable how a direct path between digits should look. The method \textit{sqDiff} is expected to smoothly traverse through non-realistic samples, whereas \textit{Linear} may traverse many digits on a lengthy path. Figure~\ref{fig:mnist} shows that our method \textit{sqDiff+D} is able to find a short path of high quality with quick transitions between digits.

\subsection{ProGAN on CelebA}
We apply our proposed method to CelebA images \cite{LiuLWT15} using the tensorflow implementation of ProGAN by Karras \etal \cite{KarrasALL18}; a progressively grown Wasserstein GAN (see WGAN-GP~\cite{arjovsky17,gulrajani2017improved}). We download the pre-trained network and reuse it in all of our experiments. As a pre-processing step, we set the feature map used for Minibatch Distcrimination (see \cite{ioffe2015batch}) to zero , which is necessary to obtain well-defined critic values, independent of the batch statistics. Then, as the Wasserstein critic is not limited to $(0,1]$, we sample a large number of points in the latent space, for which we evaluate the critic values and make note of the range of values. In subsequent evaluations, all critic values are scaled by this range in order to obtain values in $(0,1]$, as used in the proposed loss.\\

\begin{figure}[t!]
\centering
\begin{subfigure}{.23\textwidth}
  \centering
  \includegraphics[width=\linewidth]{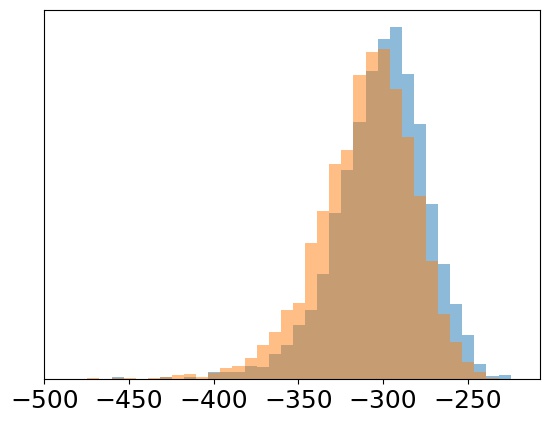}
  %\caption{ProGAN from \cite{KarrasALL18}}
  \label{fig:histoKarras}
\end{subfigure}%
\hfill
\begin{subfigure}{.239\textwidth}
  \centering
  \includegraphics[width=\linewidth]{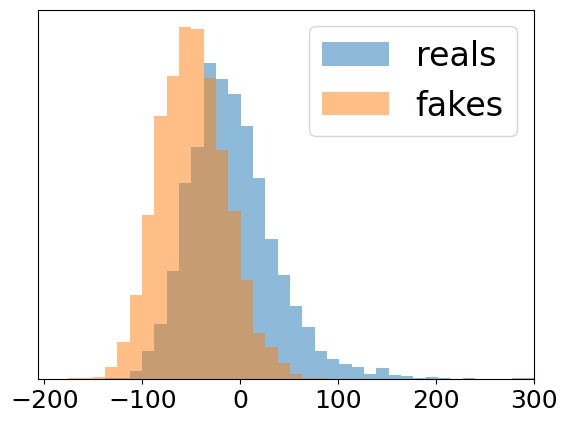}
  %\caption{Our fine-tuned ProGAN}
  \label{fig:histoOur}
\end{subfigure}
\vspace{-3mm}
\caption{Histogram of critic values over each 5000 real (blue) and 5000 generated/fake (orange) samples. Left: ProGAN from \cite{KarrasALL18}. Right: {Our fine-tuned ProGAN}}
\label{fig:histo}
\vspace{-3mm}
\end{figure}

%\paragraph{An evaluation of the pre-trained critic.}
\textbf{An evaluation of the pre-trained critic.}
Our method assumes a critic that is able to differentiate the quality of generated samples. 
Figure~\ref{fig:histo}~(a) shows a histogram of critic values on generated (fake) and real samples for the critic trained in \cite{KarrasALL18}. 
The significant overlap is a proof of the WGAN objective to be at an equilibrium.
However, random samples (shown in Appendix~\ref{app:sampleGrid}) show that, despite the high quality of samples, it is still often easy for a human to tell real apart from generated samples, which ideally should be detected by the critic network. 
To find the relevant gradient information, we fine-tune the provided networks. 
We deviate from an equal learning rate for critic and generator network to a learning rate of $5*10^{-4}$ for the critic and a learning rate of $10^{-4}$ for the generator, allowing the critic to gradually improve in comparison to the generator, but without worsening the generator results 
(We also experimented with training the discriminator only, but found the problem of observing high critic values for unrealistic samples as discussed above to worsen under this training regime. {At the same time, maintaining a positive, but small, learning rate for the generator kept the generating close to constant.}) 
We pick a model after training on 70k images ($<6~{\%}$ additional training). Figure~\ref{fig:histo}~(b) shows the histogram of real and generated samples for this critic, which shows that the new critic has some understanding of differentiating real and fake. 
%
%The persisting overlap in distributions is desirable given that some generated samples of the network cannot be differentiated by humans from real samples either. 

%-----------------------------
%-----------------------------

The necessity to fine-tune a critic before the application of our method {could in principle} 
%can 
be circumvented {by} 
%when 
training 
the ProGAN 
from scratch with an adapted training schedule. Otherwise, as in our case, the extra training is negligible in comparison to the training time of the provided model.\\ %We would highlight this difference between training an entirely new discriminating network to the little to no extra work to reuse the trained critic inherent in the WGAN.

\begin{figure*}
\centering
\begin{subfigure}{.495\textwidth}
  \centering
  \includegraphics[width=\linewidth]{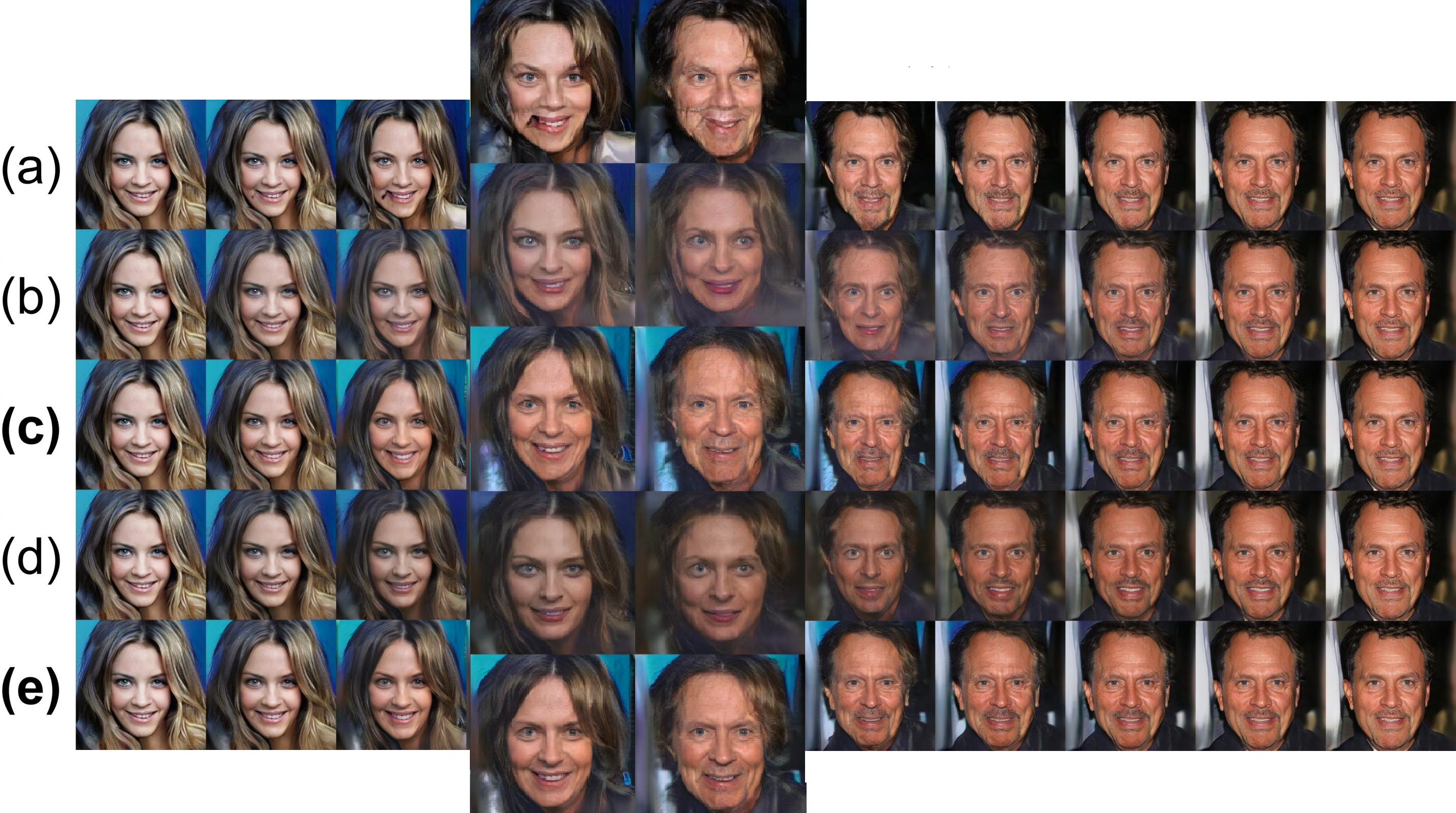}
  %\caption{ProGAN from \cite{KarrasALL18}}
  %\label{fig:histoKarras}
\end{subfigure}%
\hfill
\begin{subfigure}{.495\textwidth}
  \centering
  \includegraphics[width=\linewidth]{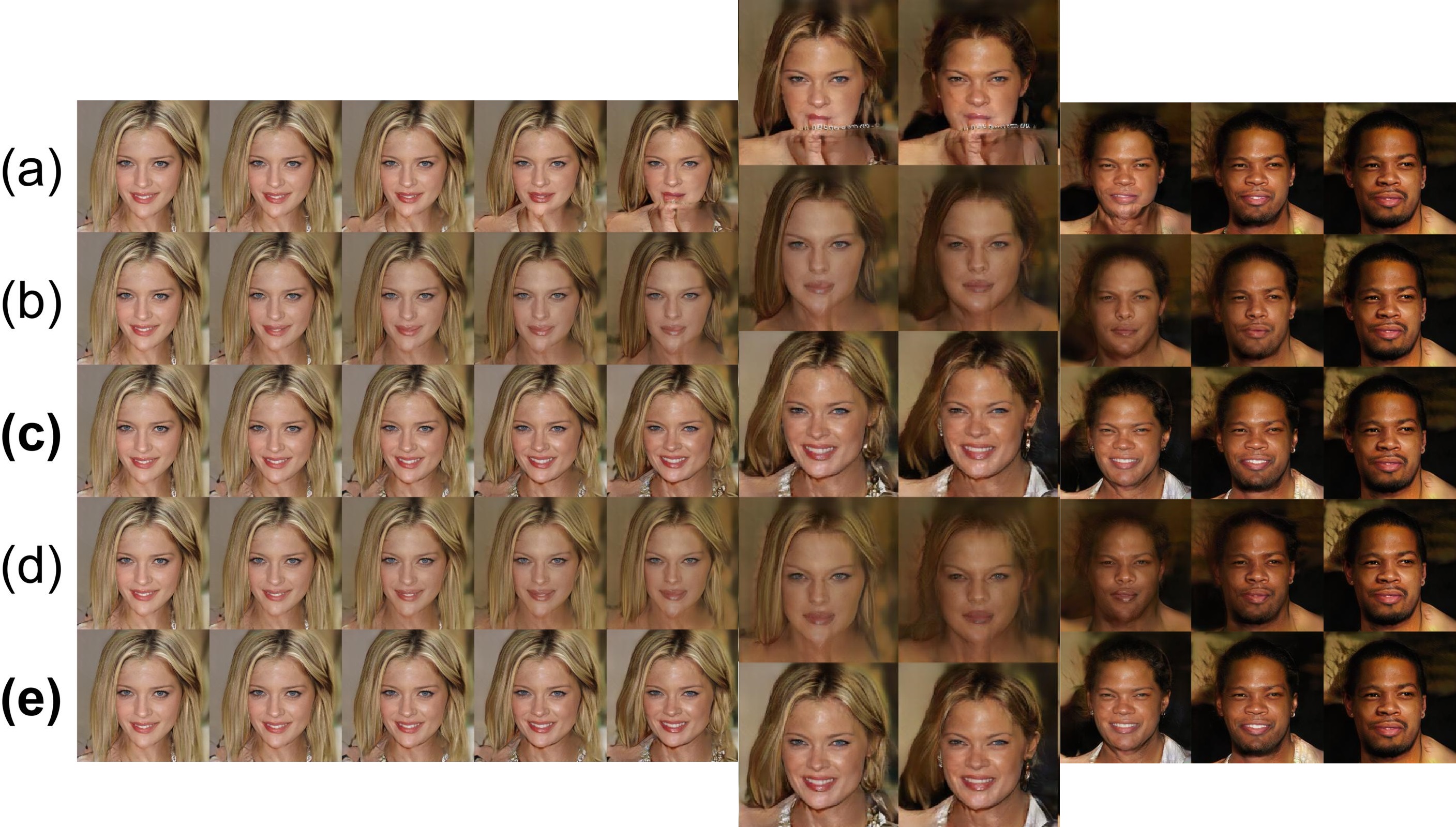}
  %\caption{Our fine-tuned ProGAN}
  %\label{fig:histoOur}
\end{subfigure}
\caption{Comparison of (a) \textit{Linear}, (b) \textit{sqDiff}, (c) \textit{sqDiff+D}, (d) \textit{VGG} and (e) \textit{VGG+D}. \textit{Linear} falls off the real data manifold and \textit{sqDiff} and \textit{VGG} result in blurry images of low quality. Our proposed methods recover crisp images from the short paths of \textit{sqDiff} and \textit{VGG} resulting in a interpolating path of high quality images.} 
\label{fig:CelebAlinearfails}
\end{figure*}

\begin{figure*}
\centering
\begin{subfigure}{.495\textwidth}
  \centering
  \includegraphics[width=\linewidth]{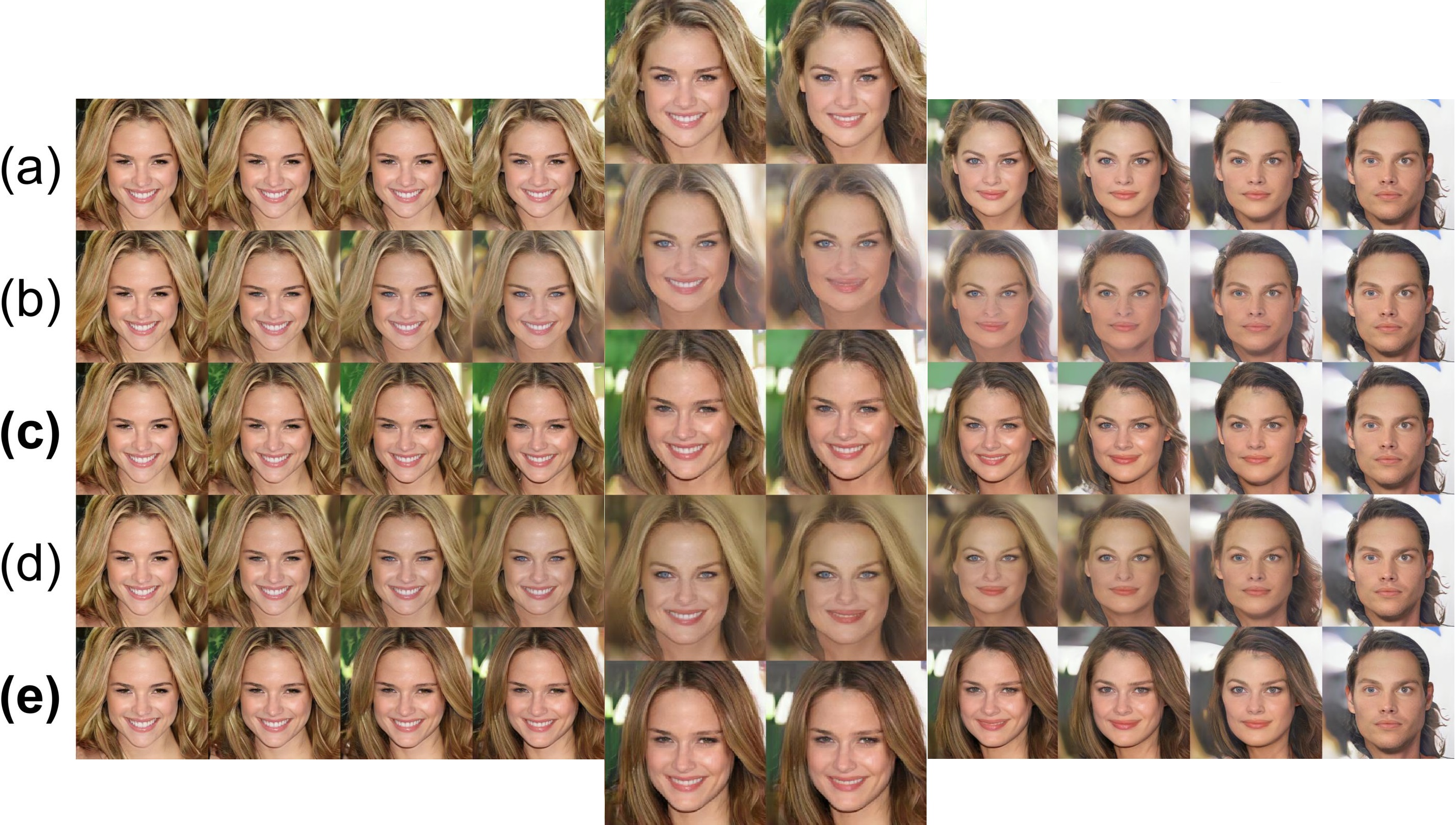}
\end{subfigure}%
\hfill
\begin{subfigure}{.495\textwidth}
  \centering
  \includegraphics[width=\linewidth]{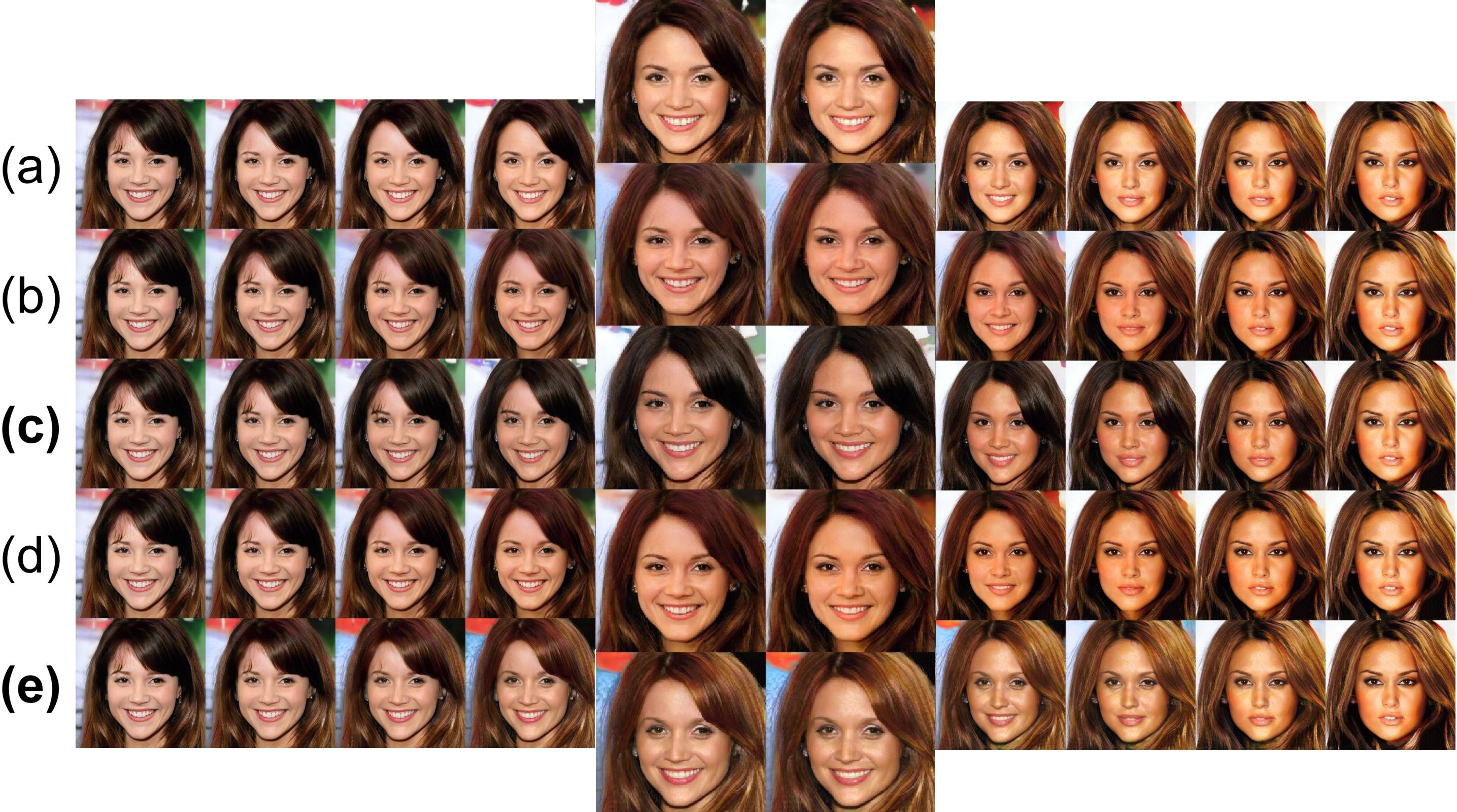}
\end{subfigure}
\caption{Comparison of (a) \textit{Linear}, (b) \textit{sqDiff}, (c) \textit{sqDiff+D}, (d) \textit{VGG} and (e) \textit{VGG+D}. Quality of \textit{Linear} is good  and (left) not close to the short path \textit{dqDiff} so that \textit{sqDiff+D} and \textit{VGG+D} recover from blurry images from \textit{sqDiff} and \textit{VGG} resp., or (right) close to an expected geodesic on the image manifold, in which case all methods approximately agree.} 
\label{fig:CelebAlinearGood}
\vspace{-3mm}
\end{figure*}

\textbf{Comparison of the linear and proposed methods.}
Randomly sampling points from the generator often gives images that can be easily distinguished from real samples (see \ref{app:sampleGrid}). As a result, the linear interpolation shows several artifacts from generation. If, however, two generated samples of high quality are observed, then we find that the linear interpolation in latent space often results in curves of high quality images, a fact that has been previously observed by Chen \etal in \cite{ChenKKJBS18}. We distinguish three scenarios for comparison of our proposed method to the linear interpolation in latent space: (i) Linear interpolation in sample space results in images of low quality, as shown in Figure~\ref{fig:CelebAlinearfails}. In this case, our method is able to find a much better path of high quality.  (ii) Linear interpolation shows samples of high quality, but the path is not a direct path, as shown in Figure~\ref{fig:CelebAlinearGood} (left). In this case, our method finds a more direct path of similarly high quality. (iii) Linear interpolation gives a direct path of high quality, as shown in Figure~\ref{fig:CelebAlinearGood} (right). In this case, our method does not change the path and approximately agrees with the linear method. \\

\begin{figure}[t]
\centering
\begin{subfigure}{.235\textwidth}
  \centering
  \includegraphics[width=\linewidth]{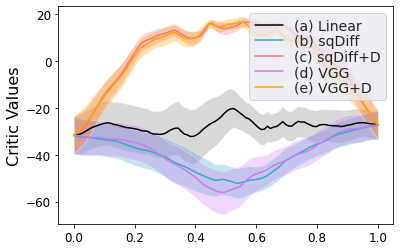}
  %\caption{}
  %\label{fig:histoKarras}
\end{subfigure}%
\begin{subfigure}{.235\textwidth}
  \centering
  \includegraphics[width=\linewidth]{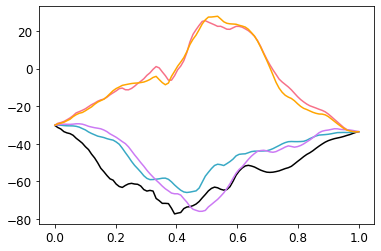}
  %\caption{}
  %\label{fig:histoOur}
\end{subfigure}
\caption{Critic values along paths. Left: Mean $\pm 1$ standard error of mean (12 paths). Right: Paths of Figure~\ref{fig:CelebAlinearfails}~left.}
\label{fig:criticAlongGeodesics}
\vspace{-5mm}
\end{figure}

\textbf{Comparison of \textit{sqDiff} and \textit{VGG} with the proposed methods.}
We find that both \textit{sqDiff} and \textit{VGG} both work well in finding a short and direct path, which can help to avoid bad regions crossed by \textit{Linear}. However, we find that both methods often result in blurry images with \textit{VGG} slightly outperforming \textit{sqDiff}. Despite a shorter path, this often results in paths of lower image quality than linear interpolation in latent space, see Figure~\ref{fig:CelebAlinearfails} and \ref{fig:CelebAlinearGood}~(left). Our method additionally enforces high discriminator values. It thereby follows the short paths obtained from these methods, but results in higher quality samples with images that are more crisp. Figure~\ref{fig:criticAlongGeodesics} shows critic values along the curve of Figure~\ref{fig:CelebAlinearfails}~(left) as well as averaged critic values over several paths. Reflecting visual analysis, the linear method usually obtains higher critic values than both \textit{sqDiff} and \textit{VGG}, but our method improves the poor image quality of the shortest path by realizing higher critic values close by.

\begin{figure}[H]
  \centering
  \includegraphics[width=.45\textwidth]{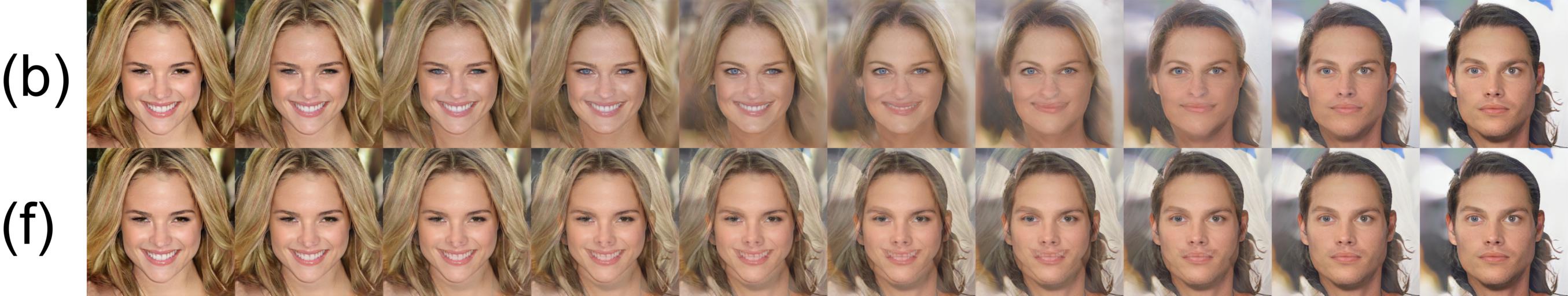}
\caption{\textit{sqDiff} approximates linear interpolation in sample space on the image manifold{, resulting in blurry images that our proposed method can improve upon}.}
\label{fig:mseVsSampleSpace}
\vspace{-3mm}
\end{figure}

%------------------------------------------------------------------------
%-----------------------------
%-----FLOATING FIGURES--------

%-----------------------------
%---------------

We explain the blurriness observed in both \textit{sqDiff} and \textit{VGG} by a curve that follows the requested objective too closely. Suppose the generator would be capable of producing any image in the sample space, then the closest path according to \textit{sqDiff} would be linear interpolation in sample space, which results in a blurry path as visualized in comparison in Figure~\ref{fig:mseVsSampleSpace}. The trained generator is expected to generate real-looking faces from anywhere in latent space, so we cannot find this curve exactly, but due to a mismatch of manifolds of real and generated images, similar paths can be found. Linear interpolation in feature spaces of the VGG network improve upon this, but some blurriness remains.\\ %Therefore, we find that also \textit{sqDiff} and \textit{VGG} benefit from not training until convergence.

\textbf{Limitations} 
The success of the proposed method relies on discriminator (or critic) values that can distinguish samples of good quality from samples with poor quality. For the swiss roll,  the good performance of the discriminator can be seen from the background color in Figure~\ref{fig:mnist}. For the other datasets, we find that the discriminator is not always reliable on single images, discovering a peculiar property of the trained critic. By evaluating the critic on small perturbations in the latent space, the critic values cover a large part critic's range without a significant visual change of the generated image. We showcase this phenomenon in Figure~\ref{fig:localPerturbations} in Appendix~\ref{app:localPerturbations} for the CelebA dataset. 

The reasons are potentially twofold: For one, a general property of Lipschitz-1 functions from a high-dimensional space of dimension $D$ (here $D=3\cdot 1024^2$) to a one-dimensional space is that small changes of size $\epsilon$ to all input dimensions can change the output by $\sqrt(D)\epsilon$. Here, to cover a range of $100$ in critic value, a change of all pixels by $0.0564$ can be sufficient. Secondly, the use of minibatch discrimination may disrupt the correlation between critic value and subjective realisticness, both locally and globally. For consistently penalizing the realisticness of shortest paths, we remove the minibatch standard deviation feature map. But %, as it is used during training of the critic, 
the relative ordering of samples may still not correspond to image quality.

The observation implies that regions of poor image quality can not always be avoided on the CelebA dataset by enforcing high critic values. Indeed, we find that the proposed method can not always recover from poor image quality along the interpolating curves. However, we still find \textit{VGG+D} outperforms all competing methods resulting in paths that are always at least as good as the interpolations found with the other methods. In particular, enforcing high critic values escapes blurry regions traversed by methods \textit{sqDiff} and \textit{VGG}.

Further, we note that some care must be taken in selecting and fine-tuning the critic network to prevent biased interpolations. By only fine-tuning the pre-trained ProGAN's critic network, we found a critic that it favored bright lightening conditions. % and seemed biased toward white faces in the dataset. 
As a direct consequence, our method became biased toward white people and appeared to suffer from the overshooting problem discussed above, converging to extremely bright samples (see \ref{app:failure}).\\

%The success of the proposed method hinges on a discriminator %critic 
%with useful gradient information between generated and real samples. By only fine-tuning the pre-trained ProGAN's critic network, we found a critic that it favored bright lightening conditions. % and seemed biased toward white faces in the dataset. 
%As a direct consequence, our method became biased toward white people and appeared to suffer from the overshooting problem discussed above, converging to extremely bright samples (see \ref{app:failure}). We therefore note that some care must be taken in selecting and fine-tuning the critic network to prevent biased interpolations.

\textbf{Summary of results.} Taken together, we find that our proposed method is always as least as good as linear interpolation in sample space, but
{can improve} 
% improves 
upon it in case of failure. While the competing methods \textit{sqDiff} and \textit{VGG} alone result in direct paths between two endpoints, they result in blurry images. Our method proposes to exploit gradient information from the critic network, enabling us to find direct paths of higher quality. %The supplement contains short videos accompanying all figures and the difference in the methods can be best observed in these.

%\paragraph{Other experiments}
%The supplementary material contains several additional examples comparing all methods. In addition, we show the consequences of changing hyperparameters and polynomial degree of the interpolating curve, where the former provides a smooth transition between the methods mse and disc or vgg and disc respectively, while the latter has little effect.

\section{Discussion and Conclusion}

%\vspace{-1mm}
At a theoretical equilibrium of a GAN objective, there is no gradient information from the discriminator between generated and real images. If the generator optimally matches the true data distribution, then the lack of discriminator information reduces the proposed methods to \textit{sqDiff} and \textit{VGG} respectively. As a result, our method is only interesting for imperfect generators that do not match the true data distribution, which is the practical case for state-of-the-art GANs. For such a generator, our results show that the discriminator information can be used to improve the sample quality of interpolatiog curves. Further improvements with the proposed method can be expected using a tailored training schedules for GANs to produce strong discriminators in the setting of sparse realistic datasets.

%\section{Conclusions}
To conclude, we have presented a method for incorporating both networks of a GAN into a method for learning realistic interpolations on the generator's image manifold. The proposed method is lightweight as it is post-hoc; reusing the discriminator that is learned in parallel to the generator. Incorporating discriminator values together with meaningful distances in a metric can avoid regions where the generator image and real data distribution do not align, resulting in paths that stay on the real data manifold. In its essence, high quality paths were achieved  by discriminating against unrealistic samples.

{\small
\bibliographystyle{ieee_fullname}
\bibliography{tk_refs}

\begin{thebibliography}{10}\itemsep=-1pt

\bibitem{arjovsky17}
Martin Arjovsky, Soumith Chintala, and L{\'e}on Bottou.
\newblock {W}asserstein generative adversarial networks.
\newblock In {\em Proceedings of the 34th International Conference on Machine
  Learning {ICML}}, 2017.

\bibitem{ArvanitidisHH18}
Georgios Arvanitidis, Lars~Kai Hansen, and S{\o}ren Hauberg.
\newblock Latent space oddity: on the curvature of deep generative models.
\newblock In {\em Proceedings of the 6th International Conference on Learning
  Representations {ICLR}}, 2018.

\bibitem{ArvanitidisHS20_unpublished}
Georgios Arvanitidis, S{\o}ren Hauberg, and Bernhard Sch{\"o}lkopf.
\newblock Geometrically enriched latent spaces.
\newblock {\em Proceedings of the 24th International Conference on Artificial
  Intelligence and Statistics}, 2021.

\bibitem{BengioMDR13}
Yoshua Bengio, Gr{\'{e}}goire Mesnil, Yann Dauphin, and Salah Rifai.
\newblock Better mixing via deep representations.
\newblock In {\em Proceedings of the 30th International Conference on Machine
  Learning, {ICML}}, 2013.

\bibitem{BerthelotRRG19}
David Berthelot*, Colin Raffel*, Aurko Roy, and Ian Goodfellow.
\newblock Understanding and improving interpolation in autoencoders via an
  adversarial regularizer.
\newblock In {\em International Conference on Learning Representations}, 2019.

\bibitem{ChenFKPBS19}
Nutan Chen, Francesco Ferroni, Alexej Klushyn, Alexandros Paraschos, Justin
  Bayer, and Patrick van~der Smagt.
\newblock Fast approximate geodesics for deep generative models.
\newblock In {\em Artificial Neural Networks and Machine Learning {ICANN: Deep
  Learning}}, 2019.

\bibitem{ChenKFBP20_unpublished}
Nutan Chen, Alexej Klushyn, Francesco Ferroni, Justin Bayer, and Patrick
  van~der Smagt.
\newblock Learning flat latent manifolds with {VAE}s.
\newblock {\em Preprint arXiv:2002.04881}, 2020.

\bibitem{ChenKKJBS18}
Nutan Chen, Alexej Klushyn, Richard Kurle, Xueyan Jiang, Justin Bayer, and
  Patrick van~der Smagt.
\newblock Metrics for deep generative models.
\newblock In {\em International Conference on Artificial Intelligence and
  Statistics, {AISTATS}}, 2018.

\bibitem{ChenCDHSSA16}
Xi Chen, Yan Duan, Rein Houthooft, John Schulman, Ilya Sutskever, and Pieter
  Abbeel.
\newblock Infogan: Interpretable representation learning by information
  maximizing generative adversarial nets.
\newblock In {\em Advances in Neural Information Processing Systems 29
  {NeurIPS}}, 2016.

\bibitem{doCarmo92}
Manfredo~P. do Carmo.
\newblock {\em Riemannian Geometry}.
\newblock Mathematics (Boston, Mass.). Birkh{\"a}user, 1992.

\bibitem{GoodfellowPMXWOCB14}
Ian Goodfellow, Jean Pouget-Abadie, Mehdi Mirza, Bing Xu, David Warde-Farley,
  Sherjil Ozair, Aaron Courville, and Yoshua Bengio.
\newblock Generative adversarial nets.
\newblock In {\em Advances in Neural Information Processing Systems 27
  {NeurIPS}}, 2014.

\bibitem{gulrajani2017improved}
Ishaan Gulrajani, Faruk Ahmed, Martin Arjovsky, Vincent Dumoulin, and Aaron
  Courville.
\newblock Improved training of {W}asserstein {GAN}s.
\newblock In {\em Advances in on Neural Information Processing Systems 30
  {NeurIPS}}, 2017.

\bibitem{ioffe2015batch}
Sergey Ioffe and Christian Szegedy.
\newblock Batch normalization: Accelerating deep network training by reducing
  internal covariate shift.
\newblock In {\em Proceedings of the 32nd International Conference on Machine
  Learning {ICML}}, 2015.

\bibitem{KarrasALL18}
Tero Karras, Timo Aila, Samuli Laine, and Jaakko Lehtinen.
\newblock Progressive growing of gans for improved quality, stability, and
  variation.
\newblock In {\em Proceedings of the 6th International Conference on Learning
  Representations {ICLR}}, 2018.

\bibitem{KarrasLA18_unpublished}
Tero {Karras}, Samuli {Laine}, and Timo {Aila}.
\newblock A style-based generator architecture for generative adversarial
  networks.
\newblock In {\em Proceedings of the IEEE/CVF Conference on Computer Vision and
  Pattern Recognition {CVPR}}, 2019.

\bibitem{KarrasLAHLA20}
Tero {Karras}, Samuli {Laine}, Miika {Aittala}, Janne {Hellsten}, Jaakko
  {Lehtinen}, and Timo {Aila}.
\newblock Analyzing and improving the image quality of stylegan.
\newblock In {\em Procceedings of the IEEE/CVF Conference on Computer Vision
  and Pattern Recognition {CVPR}}, 2020.

\bibitem{KingmaW13}
Diederik~P. Kingma and Max Welling.
\newblock Auto-encoding variational bayes.
\newblock In {\em Proceedings of the 2nd International Conference on Learning
  Representations, {ICLR}}, 2014.

\bibitem{KuhnelFJS18_unpublished}
Line K{\"{u}}hnel, Tom Fletcher, Sarang~C. Joshi, and Stefan Sommer.
\newblock Latent space non-linear statistics.
\newblock {\em Preprint arXiv:1805.07632}, 2018.

\bibitem{Laine18}
Samuli Laine.
\newblock Feature-based metrics for exploring the latent space of generative
  models.
\newblock In {\em Proceedings of the 6th International Conference on Learning
  Representations {ICLR}}, 2018.

\bibitem{LecunCB10}
Yann LeCun, Corinna Cortes, and CJ Burges.
\newblock {MNIST} handwritten digit database.
\newblock {\em ATT Labs [Online]. Available: http://yann.lecun.com/exdb/mnist},
  2, 2010.

\bibitem{LiuLWT15}
Ziwei Liu, Ping Luo, Xiaogang Wang, and Xiaoou Tang.
\newblock Deep learning face attributes in the wild.
\newblock In {\em Proceedings of International Conference on Computer Vision
  {ICCV}}, December 2015.

\bibitem{MeschederGN18}
Lars~M. Mescheder, Andreas Geiger, and Sebastian Nowozin.
\newblock Which training methods for gans do actually converge?
\newblock In {\em Proceedings of the 35th International Conference on Machine
  Learning, {ICML}}, 2018.

\bibitem{Petersen16}
Peter Petersen.
\newblock {\em Riemannian Geometry}.
\newblock Graduate Texts in Mathematics: 171. Springer International
  Publishing, 2016.

\bibitem{petzka2018regularization}
Henning Petzka, Asja Fischer, and Denis Lukovnikov.
\newblock On the regularization of wasserstein gans.
\newblock In {\em Proceedings of the 6th International Conference on Learning
  Representations {ICLR}}, 2018.

\bibitem{RadfordMC15}
Alec Radford, Luke Metz, and Soumith Chintala.
\newblock Unsupervised representation learning with deep convolutional
  generative adversarial networks.
\newblock In {\em Proceedings of the 4th International Conference on Learning
  Representations, {ICLR}}, 2016.

\bibitem{RumelhartHW86}
David.~E. Rumelhart, Geoffrey.~E. Hinton, and Ronald~J. Williams.
\newblock Parallel distributed processing: Explorations in the microstructure
  of cognition, vol. 1.
\newblock chapter Learning Internal Representations by Error Propagation. MIT
  Press, 1986.

\bibitem{Shao0F18}
Hang Shao, Abhishek Kumar, and P.~Thomas Fletcher.
\newblock The {R}iemannian geometry of deep generative models.
\newblock In {\em 2018 {IEEE} Conference on Computer Vision and Pattern
  Recognition Workshops}, 2018.

\bibitem{Simonyan15}
Karen Simonyan and Andrew Zisserman.
\newblock Very deep convolutional networks for large-scale image recognition.
\newblock In {\em Proceedings of the 3rd International Conference on Learning
  Representations {ICLR}}, 2015.

\bibitem{StolbergLarsenS21}
Jakob Stolberg-Larsen and Stefan Sommer.
\newblock Atlas generative models and geodesic interpolation.
\newblock {\em Preprint arXiv:2102.00264}, 2021.

\bibitem{villani2008optimal}
C{\'e}dric Villani.
\newblock {\em Optimal transport: old and new}, volume 338.
\newblock Springer Science \& Business Media, 2008.

\bibitem{WangZSW21}
Zhengwei Wang, Qi She, and Tom\'{a}s~E. Ward.
\newblock Generative adversarial networks in computer vision: A survey and
  taxonomy.
\newblock 2021.

\bibitem{YangAFLH18_unpublished}
Tao Yang, Georgios Arvanitidis, Dongmei Fu, Xiaogang Li, and S{\o}ren Hauberg.
\newblock Geodesic clustering in deep generative models.
\newblock {\em Preprint arXiv:1809.04747}, 2020.

\bibitem{zhang2018unreasonable}
Richard Zhang, Phillip Isola, Alexei~A Efros, Eli Shechtman, and Oliver Wang.
\newblock The unreasonable effectiveness of deep features as a perceptual
  metric.
\newblock In {\em Proceedings of the IEEE conference on computer vision and
  pattern recognition {CVPR}}, 2018.

\end{thebibliography}
} 
\flushcolsend

\clearpage \newpage
\raggedend
\raggedcolsend

\appendix

\onecolumn

\begin{table*}[t]
\centering
\begin{tabular}{c:c|c:c|c:c}
\multicolumn{2}{c|}{Swiss roll}  &  \multicolumn{2}{c|}{MNIST}  & \multicolumn{2}{c}{CelebA-HQ}  \\ \hline
     \textit{D}  & self-trained &\textit{D}  & trained using \cite{KarrasALL18} &\textit{D}  & fine-tuned from \cite{KarrasALL18}  \\ 
       \textit{n\_interp\_pts}& 1024 & \textit{n\_interp\_pts}  & 24  &\textit{n\_interp\_pts} & 10 \\ 
        \textit{poly\_degree}& 6 & \textit{poly\_degree} & 4 & \textit{poly\_degree} &  3 \\
      \textit{n\_train\_steps} &  1000  &\textit{n\_train\_steps}  & 100 &\textit{n\_train\_steps}  & 300  \\
       \textit{learn\_rate}& $10^{-3}$ &\textit{learn\_rate}  & $10^{-3}$ &\textit{learn\_rate}  & $10^{-3}$ \\
       \textit{hyperparam} $\lambda$ & $50$  &\textit{hyperparam} $\lambda$  & $0.6$  & \textit{hyperparam} $\lambda$ & $0.4$ \\
       \textit{coefficient\_init} & $10^{-1}$ &\textit{coefficient\_init}  &  $10^{-4}$ &\textit{coefficient\_init}  & $10^{-4}$ \\
\end{tabular}
\end{table*}

\begin{table*}[t]
\centering
\begin{tabular}{c:c|c:c}
\multicolumn{2}{c|}{LSUN Cars}  &  \multicolumn{2}{c|}{LSUN bedrooms}  \\ \hline
     \textit{D}  & fine-tuned from \cite{KarrasALL18} &\textit{D}  & fine-tuned from \cite{KarrasALL18}  \\ 
       \textit{n\_interp\_pts}& 10 & \textit{n\_interp\_pts}  & 10  \\ 
        \textit{poly\_degree}& 3 & \textit{poly\_degree} & 3  \\
      \textit{n\_train\_steps} &  300  &\textit{n\_train\_steps}  & 200   \\
       \textit{learn\_rate}& $10^{-3}$ &\textit{learn\_rate}  & $10^{-3}$  \\
       \textit{hyperparam} $\lambda$ & $0.2$  &\textit{hyperparam} $\lambda$   &0.2  \\
       \textit{coefficient\_init} & $10^{-4}$ &\textit{coefficient\_init}  &  $10^{-4}$  \\
\end{tabular}
\end{table*}

\section{Training setup}\label{app:trainingSetup}

\subsection{Polynomial parametrization}
To find shortest paths, i.e., geodesics, with respect to the different metrics, we parametrize a multi-dimensional polynomial \[p(t)=a_0 +a_1t+a_2t^2...+a_n t^n,\]where each $a_i\in\mathbb{R}^d$ for $d$ the dimension of the latent space. When learning shortest paths by minimizing curve lengths over the polynomial coefficients $a_i$, we do not want the start and endpoint to change. To that end, we consider coefficients $a_i, i\geq 2$ as free parameters and calculate the constant part $a_0$ and the linear part $a_1$ to satisfy the constraints at the endpoint.

The standard choice for the parametrization of curves chooses a time-interval of $t\in[0,1]$. For higher degrees $j$ of polynomials, small numbers for $t$ result in vanishing contributions $a_j t^j$. To mitigate this problem, we parametrize all curves using the interval $t\in[1,2]$ instead.

\subsection{Parameters of implementation}
Our implementation allows the selection of several parameters for learning interpolating curves. The following list comprises the most relevant parameters
\begin{itemize}
    \itemsep0em 
    \item \textit{D}; The discriminator model to use for methods \textit{sqDiff+D} and \textit{VGG+D}.
    \item \textit{Start, End}; Seeds for start and endpoint of the path.
    \item \textit{n\_interp\_pts}; The number of interpolating points on the curve
    \item \textit{poly\_degree}; The polynomial degree of the interpolating curve
    \item \textit{n\_train\_steps}; The number of training steps to update the curve parameters using Adam optimizer to minimize the respective objectives for path minimization with respect to the different metrics
    \item \textit{learn\_rate}; The learning rate used for the Adam optimizer minimizing path length.
    \item \textit{hyperparam} $\lambda$; Hyperparameter used for strength of regularization of the curves using discriminator values as in \eqref{eq:numericalProblem} divided by dimension of image space of $H$.
    \item \textit{coefficient\_init}; Initialization of polynomial coefficients to learn is chosen uniformly random in the interval of this size
    \item \textit{Methods};Choice of method to use. Our implementation features \textit{linear} interpolation, \textit{sqDiff}, \textit{sqDiff+D}, \textit{VGG}, \textit{VGG+D}, \textit{Linear} interpolation \textit{in sample space} and looking for high discriminator values only.    
\end{itemize}

For each dataset, we manually searched for suitable hyperparameters based on visual exploration, Once we found a good setting, we used that choice of hyperparameters for the generation of all results. 

Differences in hyperarameter settings between Swiss Roll versus MNIST and CelebA-HQ can be explained by using a regular GAN with $(0,1)$-valued discriminator for the swiss roll, while using a Wasserstein GAN for the other experiments (see Section~\ref{sct:adaptiveMethod},~Adapting our method to Wasserstein GANs.

\subsection{More implementation details}
\subsubsection{Swiss Roll}
Examining the discriminator values for the GAN trained on Swiss roll in Figure \ref{fig:Swiss_roll_paths}, one sees that there is a thin black region of low values between the origin and the south-east corner. Following the, e.g., linear, paths crossing over this region one sees (in sample space) that these paths fall off the swiss roll's real data manifold, by jumping across the roll instead of following along it. When training the GAN, this bad region of latent points is apparently not removed, but instead shrunk, becoming very very thin. Nonetheless, when interpolating across it, the results are detrimental. During training of \textit{sqDiff} and the proposed \textit{sqDiff+D}, we found that, starting from linear interpolation, there was not always enough gradient information available to escape the unrealistic path. Instead, we trained an ensamble of paths, having the same endpoints but different (and larger) initialization, where after training the path with smallest objective was chosen. In addition, because the bad region of latent points was so thin, to make the selection of hyperparameter less important, we replaced the discrete discriminator values along the paths by the geometric average of discriminator values between adjacent points, thus widening and flattening the region somewhat.

\section{Additional Experiments}
\subsection{CelebA-HQ}\label{app:CelebA}
\subsubsection{A grid of samples}\label{app:sampleGrid}
\begin{figure}[H]
\centering
  \includegraphics[width=0.99\linewidth]{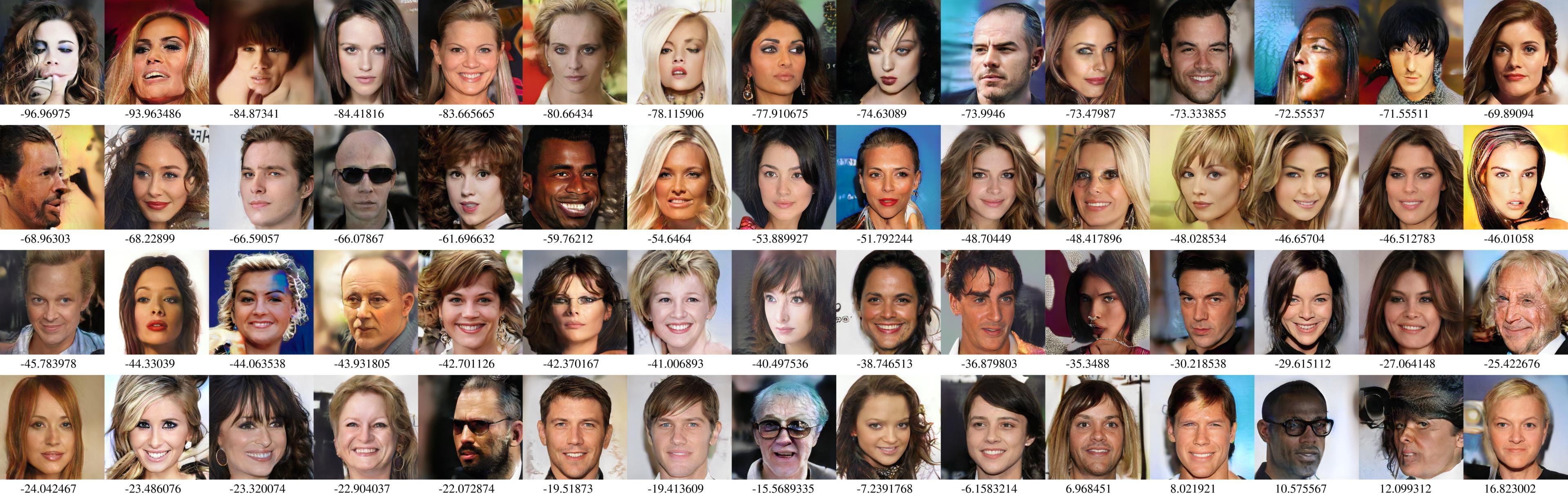}
\caption{Randomly generated CelebA-HQ images using ProGAN.} \label{fig:random_grid} 
\end{figure}

\begin{figure}[H]
\centering
  \includegraphics[width=0.99\linewidth]{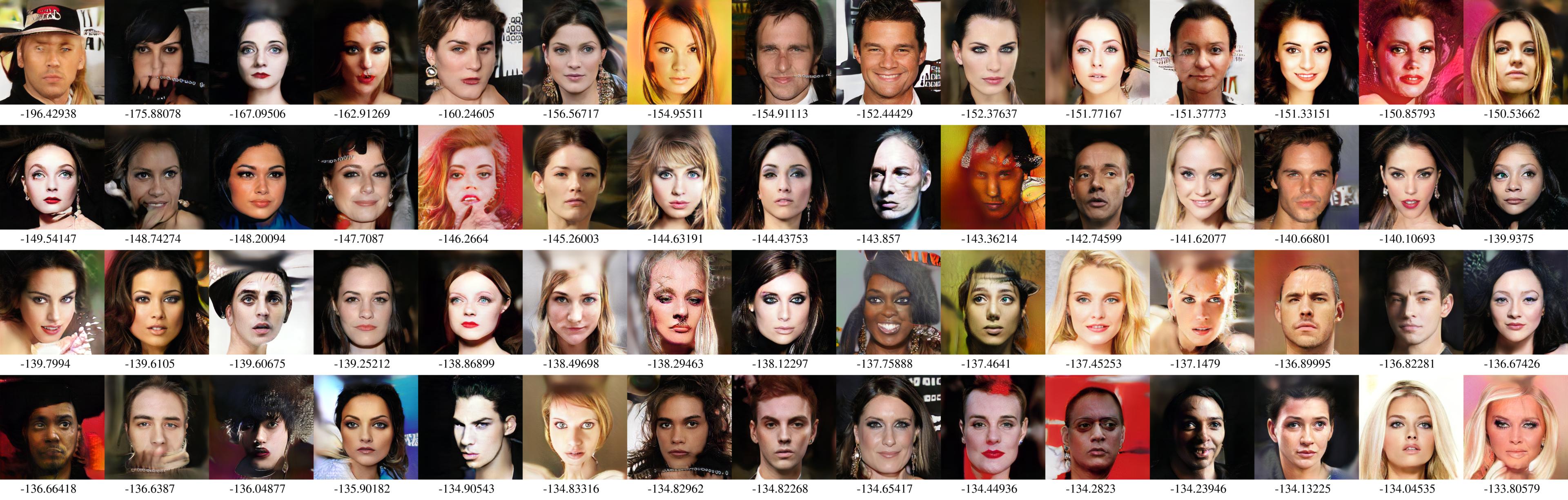}
\caption{Worst-case images (among 20k random samples) according to our fine-tuned ProGAN discriminator network.} \label{fig:worst_grid}
\end{figure}

\begin{figure}[H]
\centering
  \includegraphics[width=0.99\linewidth]{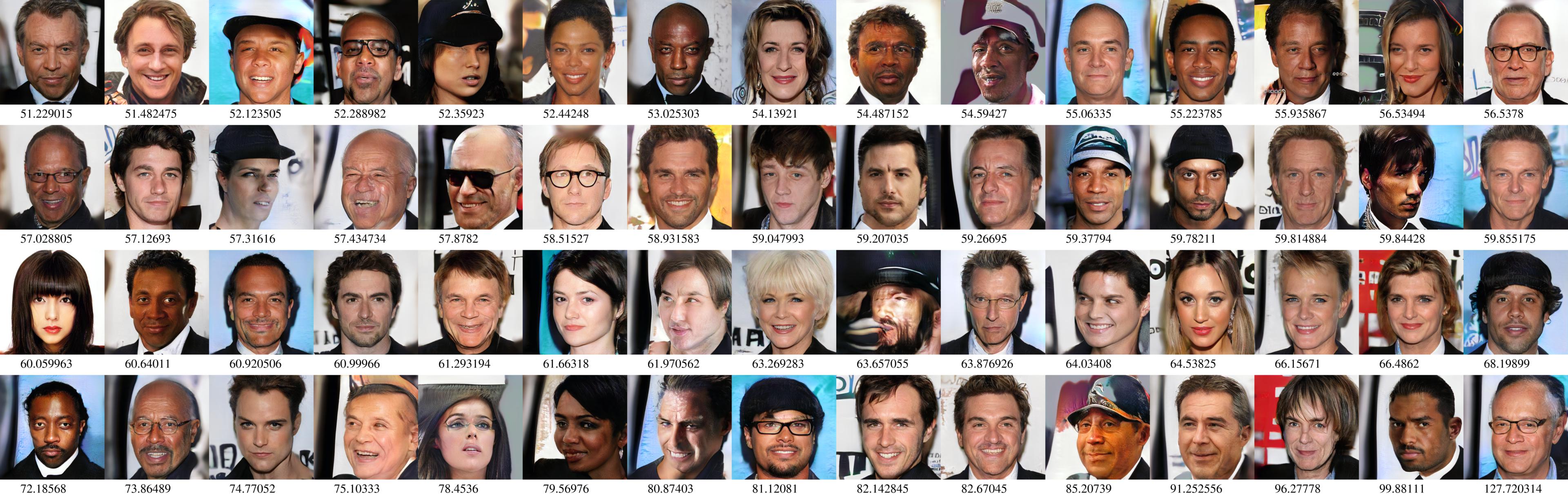}
\caption{Best-case images (among 20k random samples) according to our fine-tuned ProGAN discriminator network.} \label{fig:best_grid}
\end{figure}

In Figure \ref{fig:random_grid}, we examine a set of synthetic CelebA-HQ images (from our fine-tuned ProGAN) and corresponding critic values, sampled using latent codes drawn from $p_Z$, i.e., randomly from the generator manifold. It is apparently clear that many images are very realistic, meaning that they would fool a human examiner, but that a surprising amount (perhaps 20 \% in this set) are still quite unrealistic, exhibiting either unnatural facial distortions, or show a varying amount of artifacts in either the foreground or in the background, which wouldn't fool the human examiner. Indeed, to sucessfully utilize this generator in real-world applications, some points in the latent space will need to be avoided.  
To that end, figures \ref{fig:worst_grid} and \ref{fig:best_grid} showcase the discriminator's ability to separate the most realistic from the least realistic generations, by displaying a set of images with low and high discriminator values, respectively. These images form the two tails of the discriminator value distribution, sampled from a collection of 20k generated images. Considering the number of unrealistic examples in each set, they are not all and none, unfortunately, but they are in majority among the worst, and in clear minority among the best. 

\subsubsection{Additional geodesics}\label{app:moreCelebA}
Below follows some more examples of trained paths for ProGAN trained on CelebA-HQ for (a) \textit{Linear}, (b) \textit{sqDiff}, (c) \textit{sqDiff+D}, (d) \textit{VGG} and (e) \textit{VGG+D} on CelebA.

\begin{figure}[H]
\centering
  \includegraphics[width=0.8\linewidth]{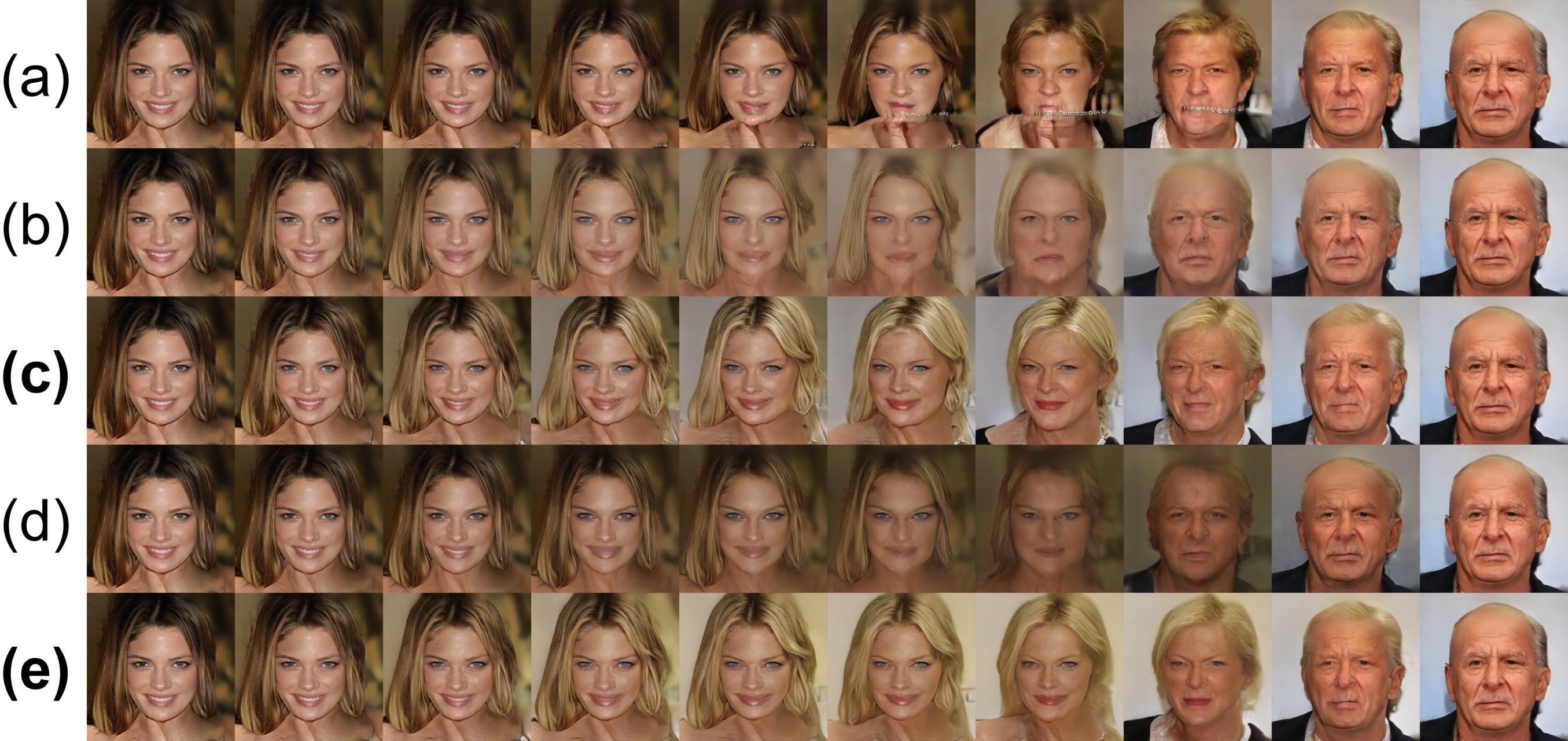}
%\caption{Comparison of (a) \textit{Linear}, (b) \textit{sqDiff}, (c) \textit{sqDiff+D}, (d) \textit{VGG} and (e) \textit{VGG+D} on CelebA.} 
\end{figure}

\begin{figure}[H]
\centering
  \includegraphics[width=0.8\linewidth]{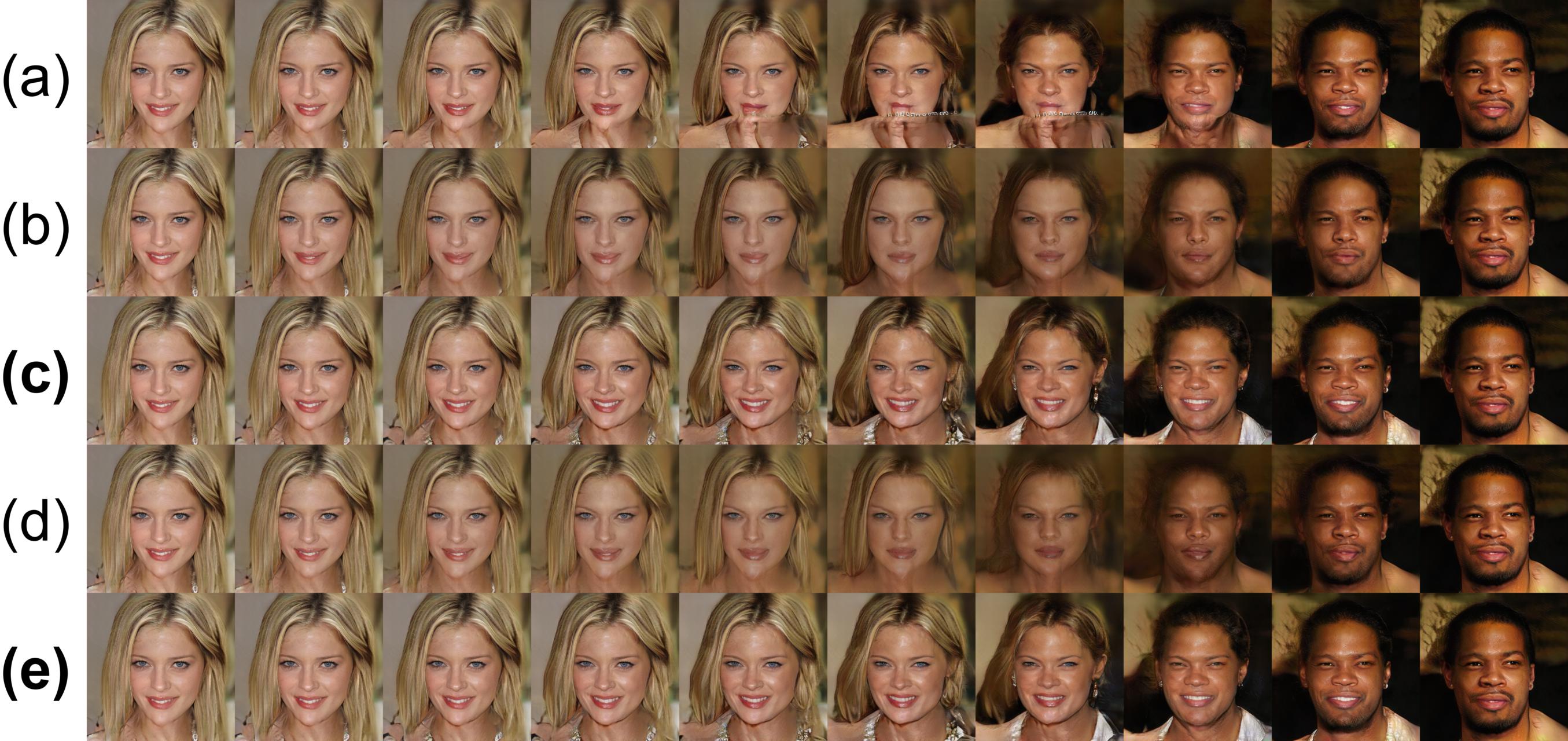}
%\caption{Comparison of (a) \textit{Linear}, (b) \textit{sqDiff}, (c) \textit{sqDiff+D}, (d) \textit{VGG} and (e) \textit{VGG+D} on CelebA.} 
\end{figure}

\begin{figure}[H]
\centering
  \includegraphics[width=0.8\linewidth]{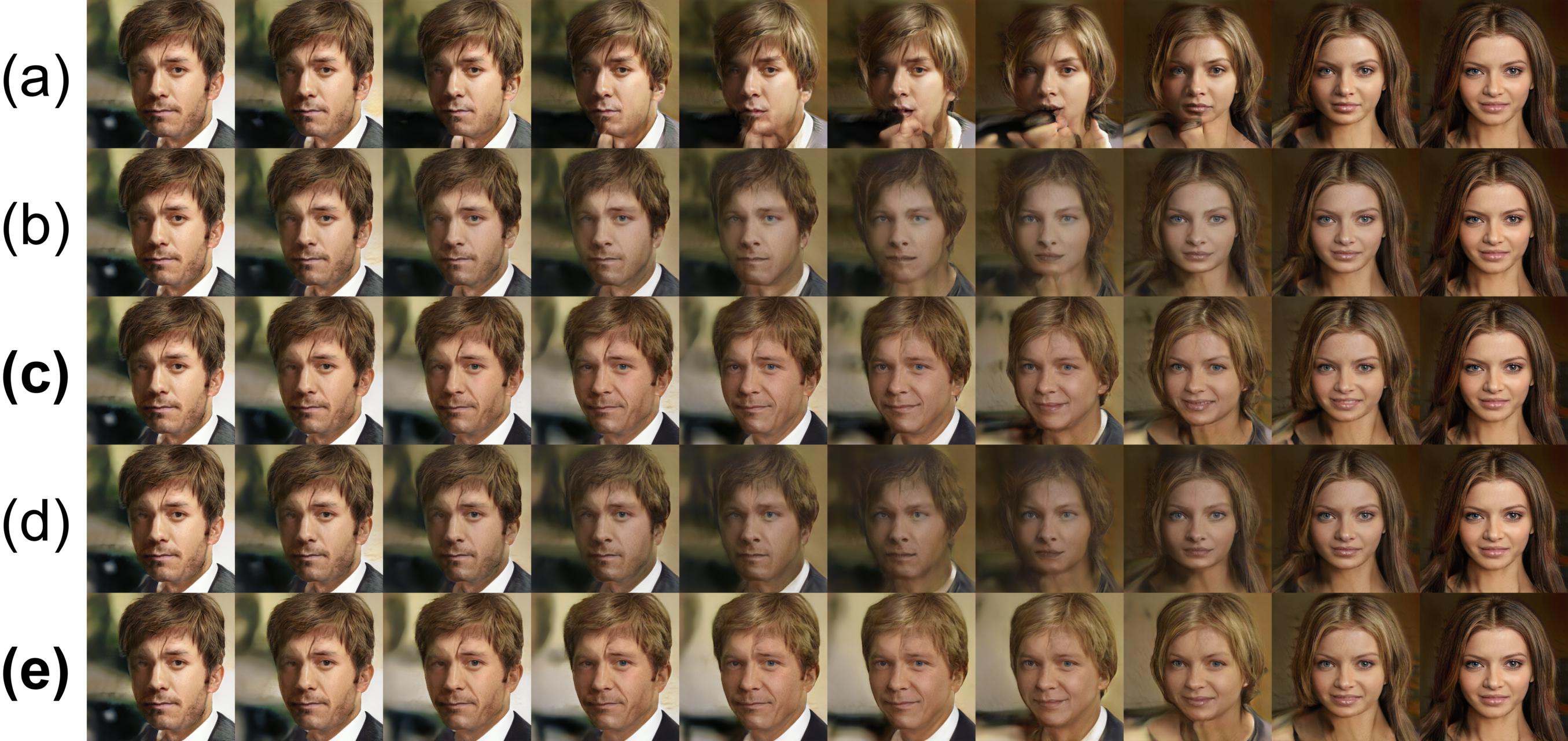}
%\caption{Comparison of (a) \textit{Linear}, (b) \textit{sqDiff}, (c) \textit{sqDiff+D}, (d) \textit{VGG} and (e) \textit{VGG+D} on CelebA.} 
\end{figure}

\begin{figure}[H]
\centering
  \includegraphics[width=0.8\linewidth]{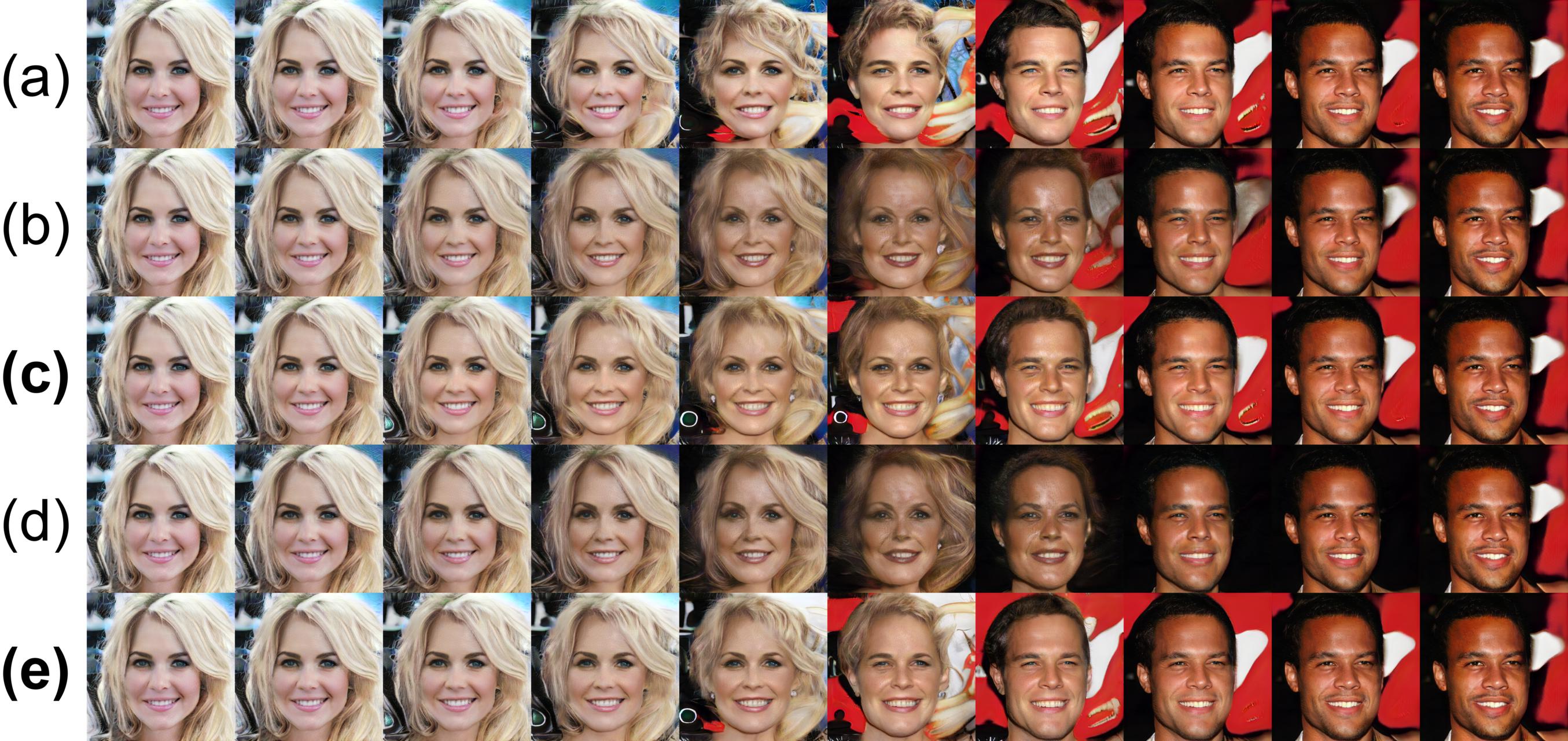}
%\caption{Comparison of (a) \textit{Linear}, (b) \textit{sqDiff}, (c) \textit{sqDiff+D}, (d) \textit{VGG} and (e) \textit{VGG+D} on CelebA.} 
\end{figure}

\begin{figure}[H]
\centering
  \includegraphics[width=0.8\linewidth]{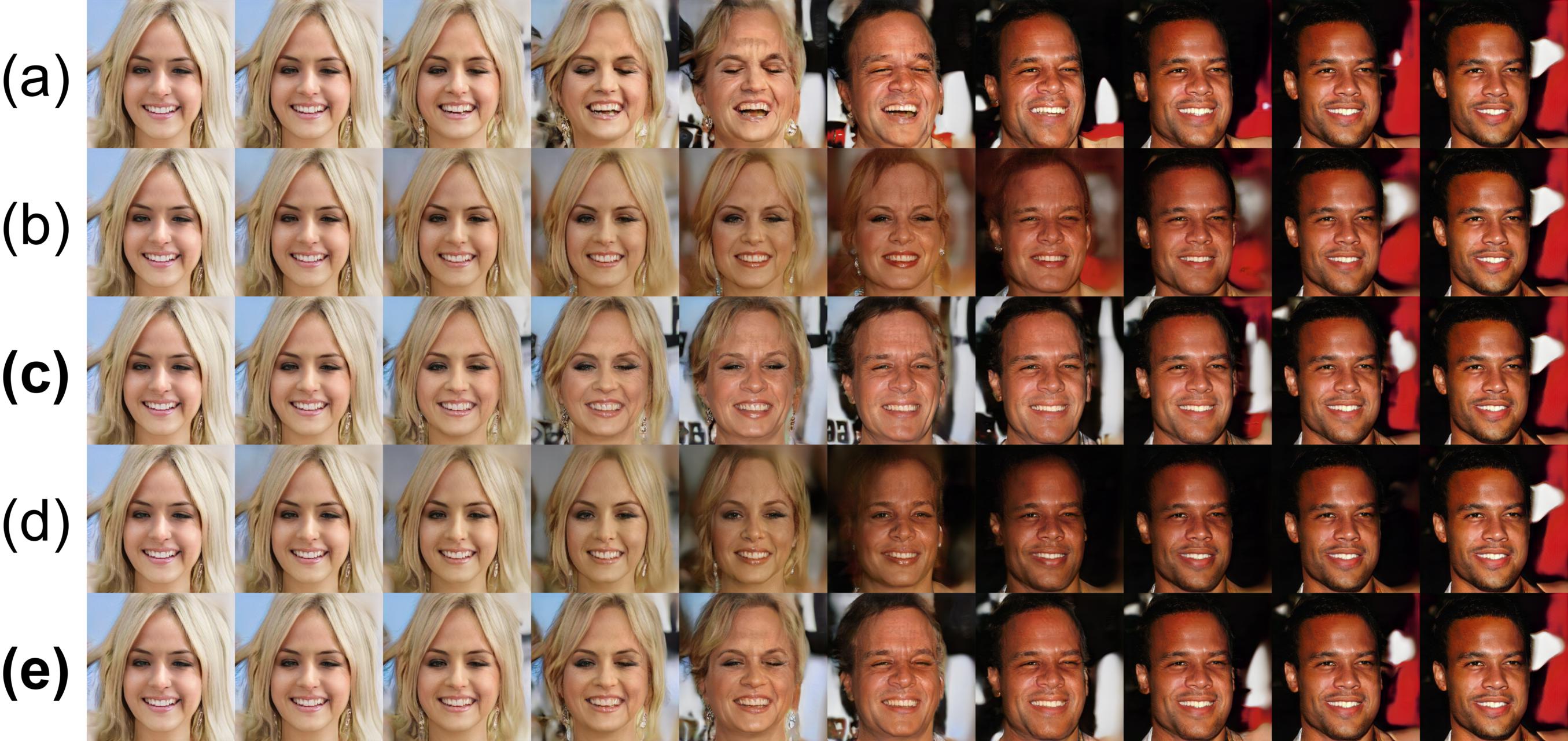}
%\caption{Comparison of (a) \textit{Linear}, (b) \textit{sqDiff}, (c) \textit{sqDiff+D}, (d) \textit{VGG} and (e) \textit{VGG+D} on CelebA.} 
\end{figure}

\begin{figure}[H]
\centering
  \includegraphics[width=0.8\linewidth]{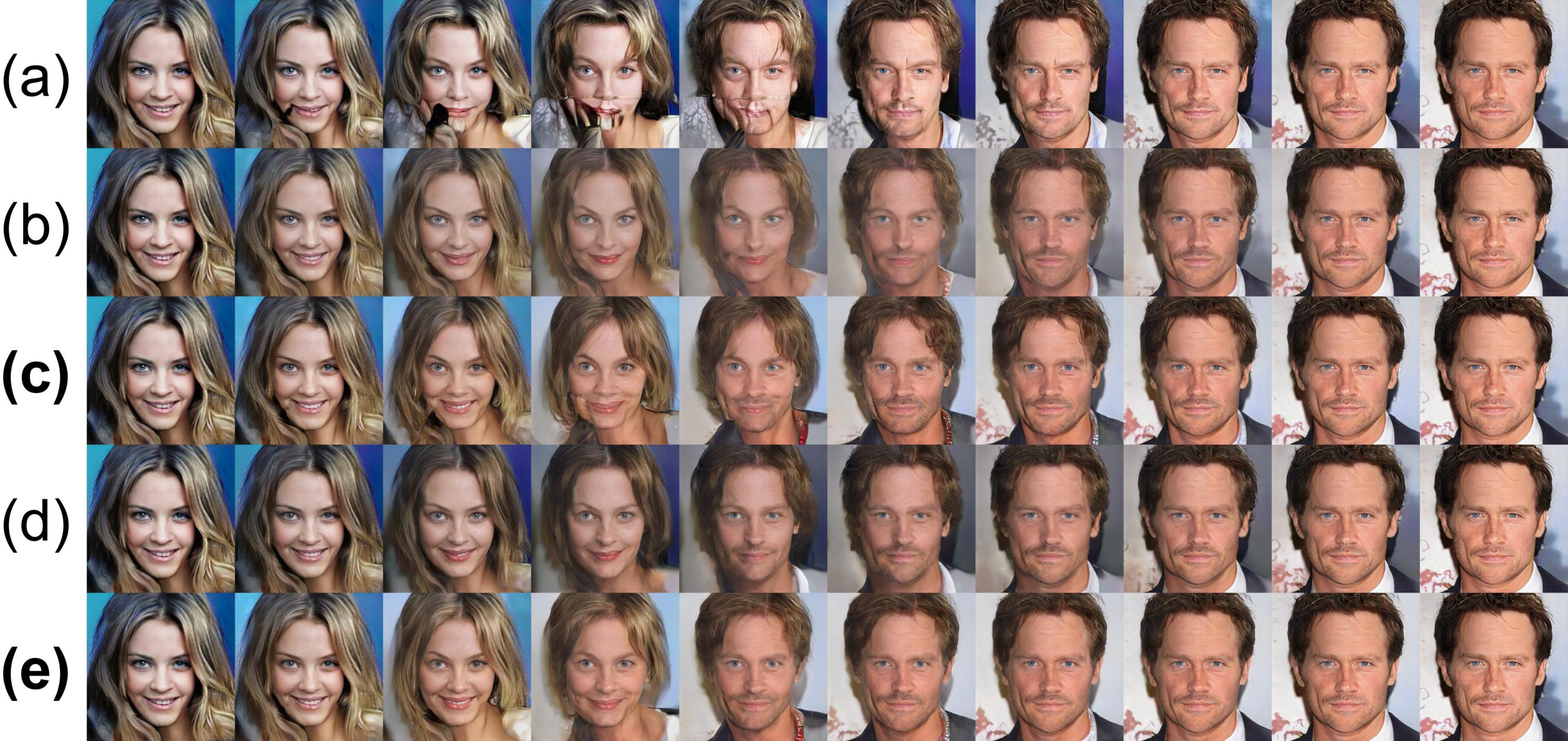}
%\caption{Comparison of (a) \textit{Linear}, (b) \textit{sqDiff}, (c) \textit{sqDiff+D}, (d) \textit{VGG} and (e) \textit{VGG+D} on CelebA.} 
\end{figure}

\begin{figure}[H]
\centering
  \includegraphics[width=0.8\linewidth]{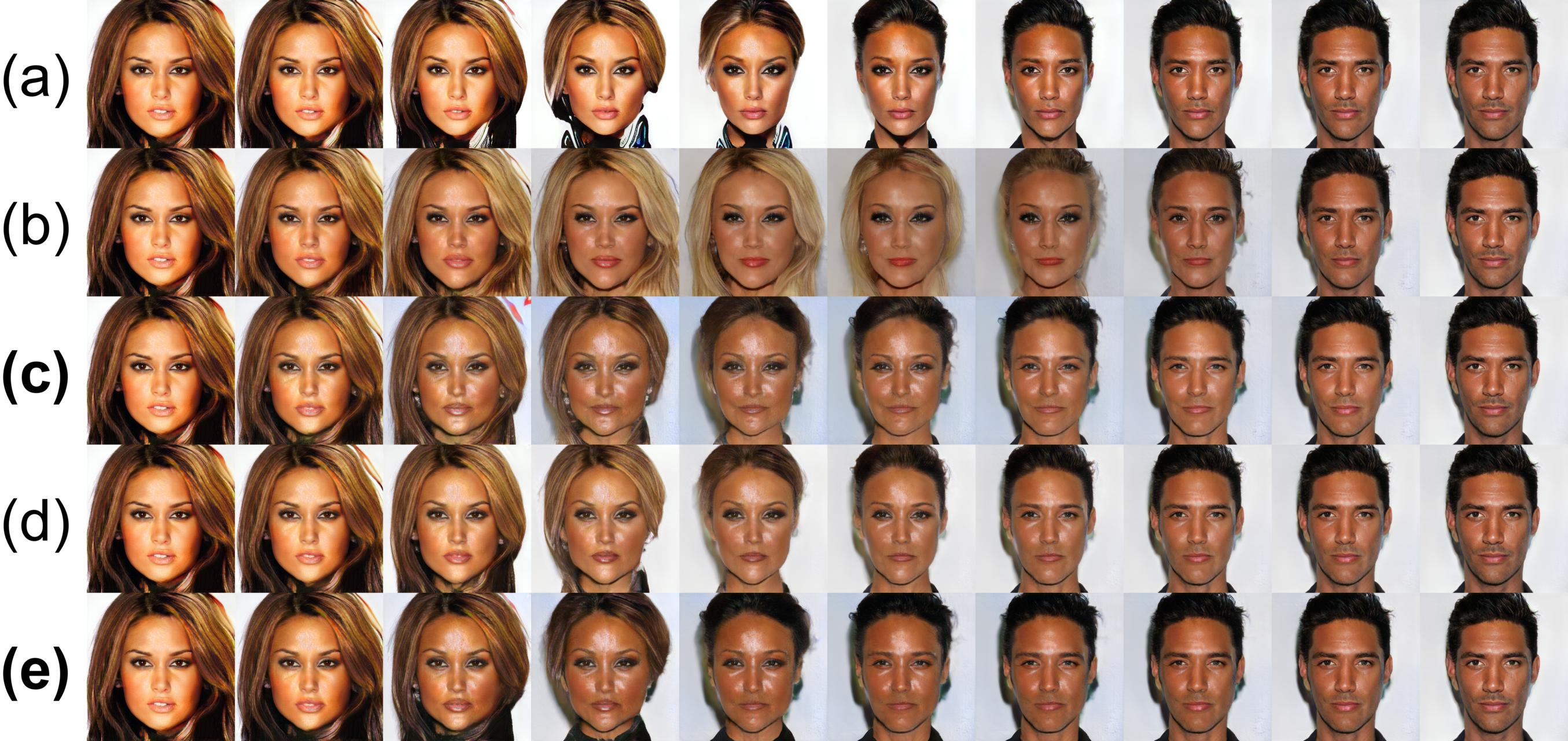}
%\caption{Comparison of (a) \textit{Linear}, (b) \textit{sqDiff}, (c) \textit{sqDiff+D}, (d) \textit{VGG} and (e) \textit{VGG+D} on CelebA.} 
\end{figure}

\begin{figure}[H]
\centering
  \includegraphics[width=0.8\linewidth]{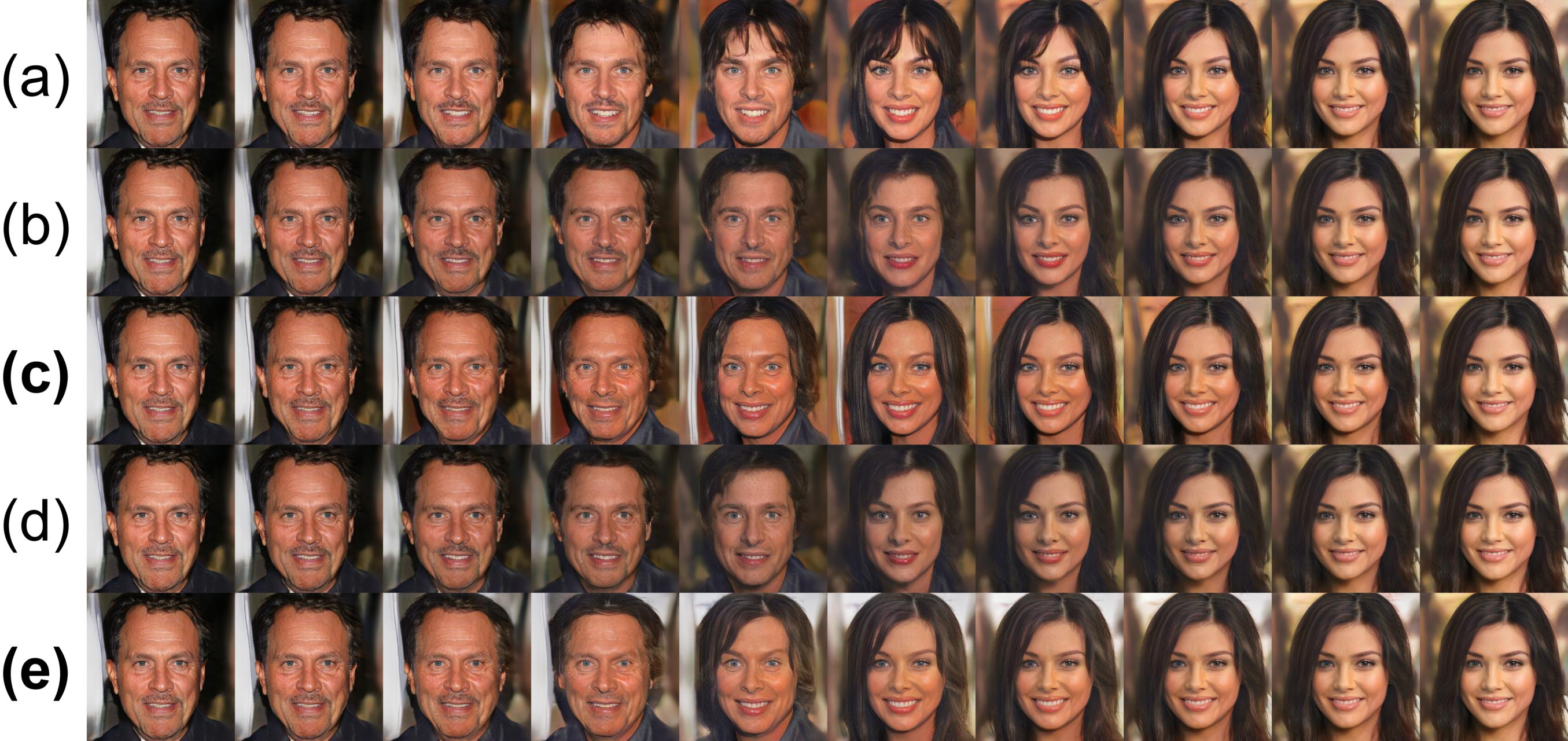}
%\caption{Comparison of (a) \textit{Linear}, (b) \textit{sqDiff}, (c) \textit{sqDiff+D}, (d) \textit{VGG} and (e) \textit{VGG+D} on CelebA.} 
\end{figure}

\begin{figure}[H]
\centering
  \includegraphics[width=0.8\linewidth]{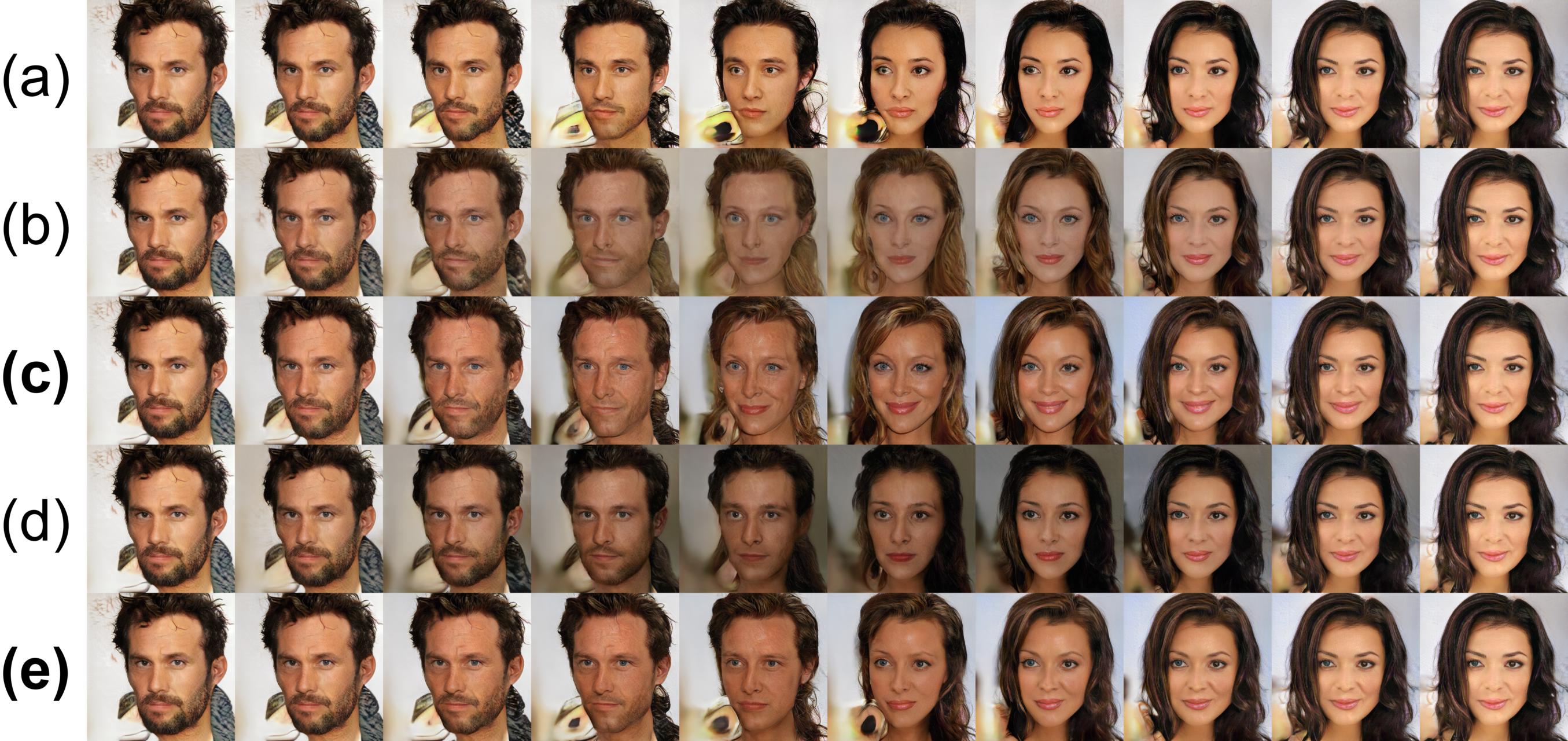}
%\caption{Comparison of (a) \textit{Linear}, (b) \textit{sqDiff}, (c) \textit{sqDiff+D}, (d) \textit{VGG} and (e) \textit{VGG+D} on CelebA.} 
\end{figure}

\begin{figure}[H]
\centering
  \includegraphics[width=0.8\linewidth]{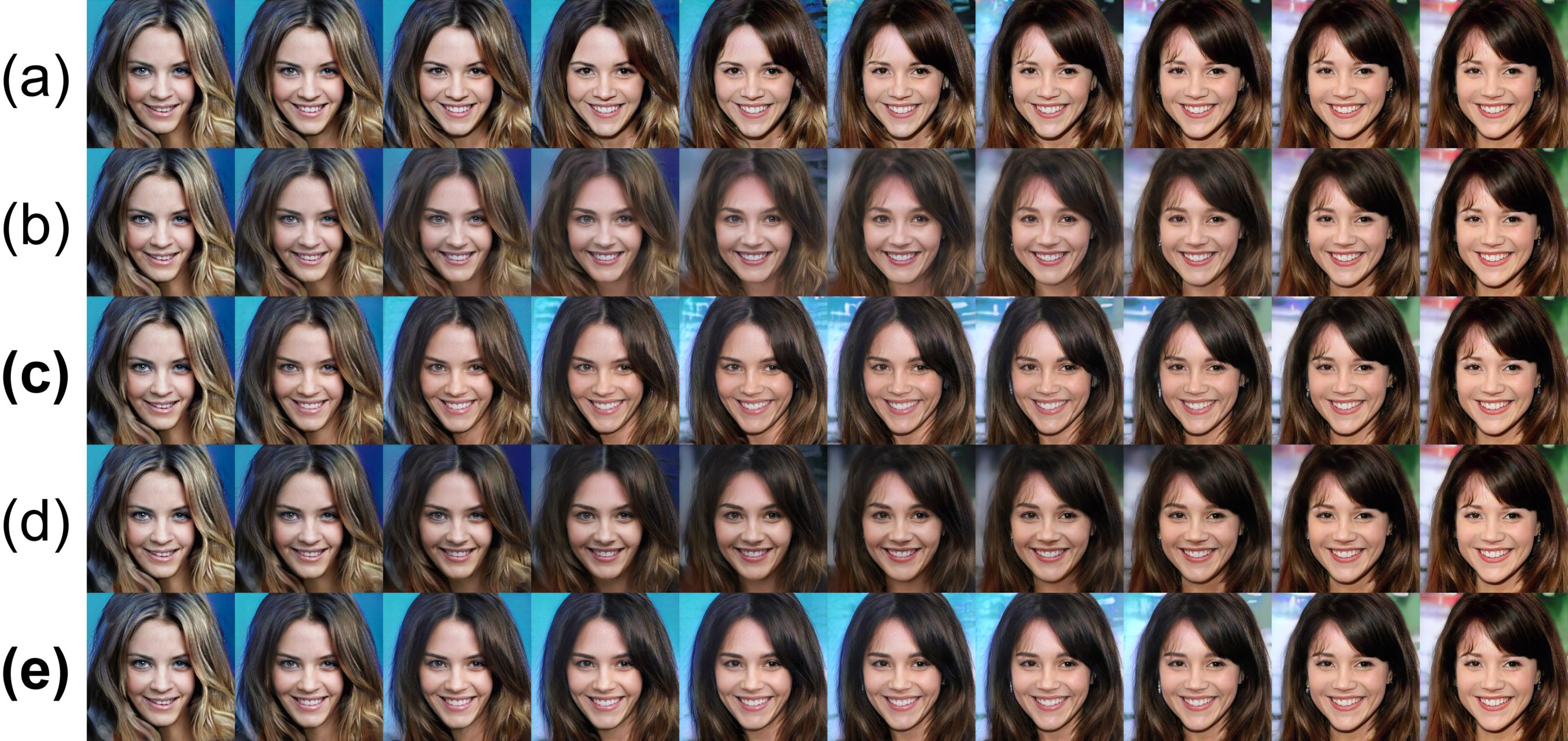}
%\caption{Comparison of (a) \textit{Linear}, (b) \textit{sqDiff}, (c) \textit{sqDiff+D}, (d) \textit{VGG} and (e) \textit{VGG+D} on CelebA.} 
\end{figure}

\begin{figure}[H]
\centering
  \includegraphics[width=0.8\linewidth]{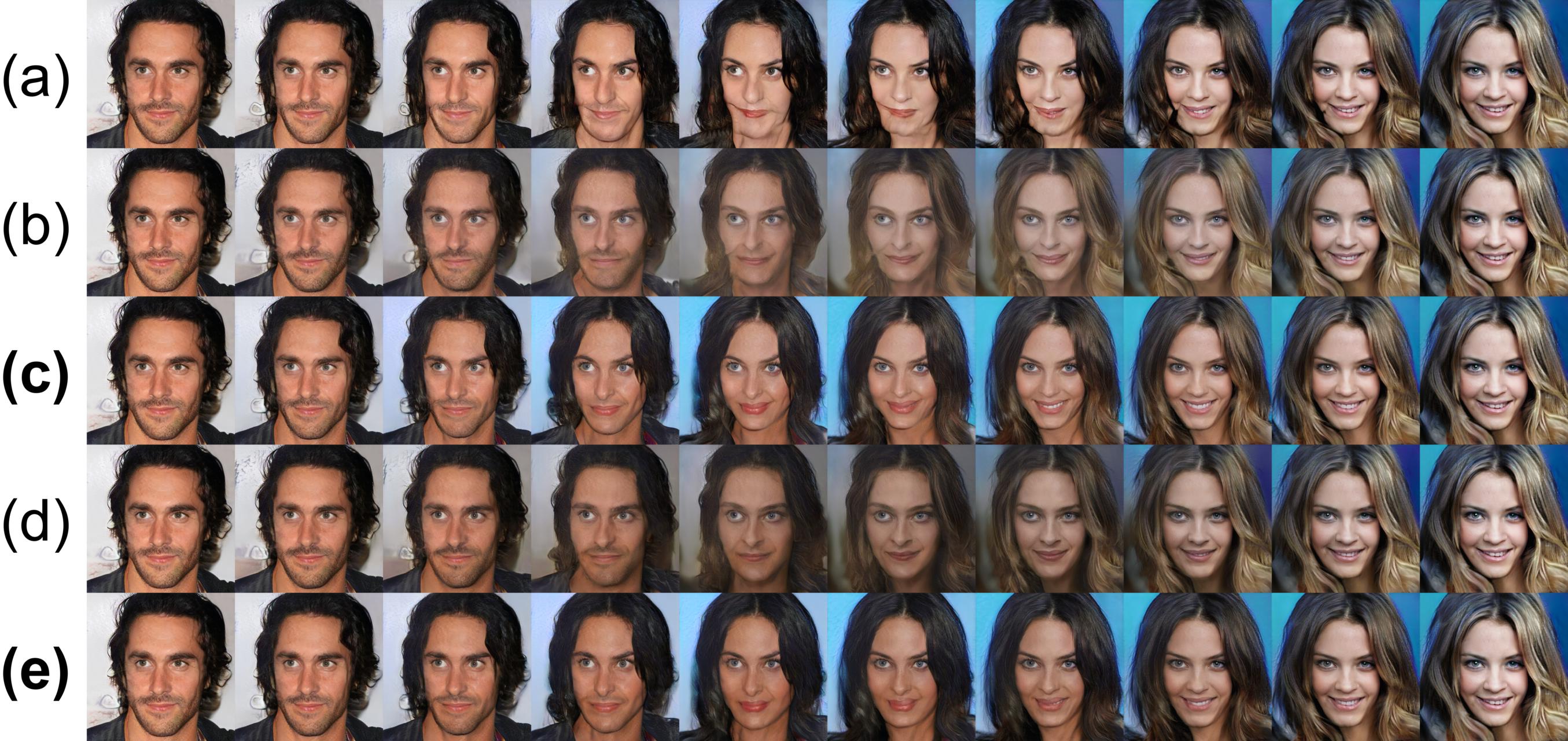}
%\caption{Comparison of (a) \textit{Linear}, (b) \textit{sqDiff}, (c) \textit{sqDiff+D}, (d) \textit{VGG} and (e) \textit{VGG+D} on CelebA.} 
\end{figure}

\begin{figure}[H]
\centering
  \includegraphics[width=0.8\linewidth]{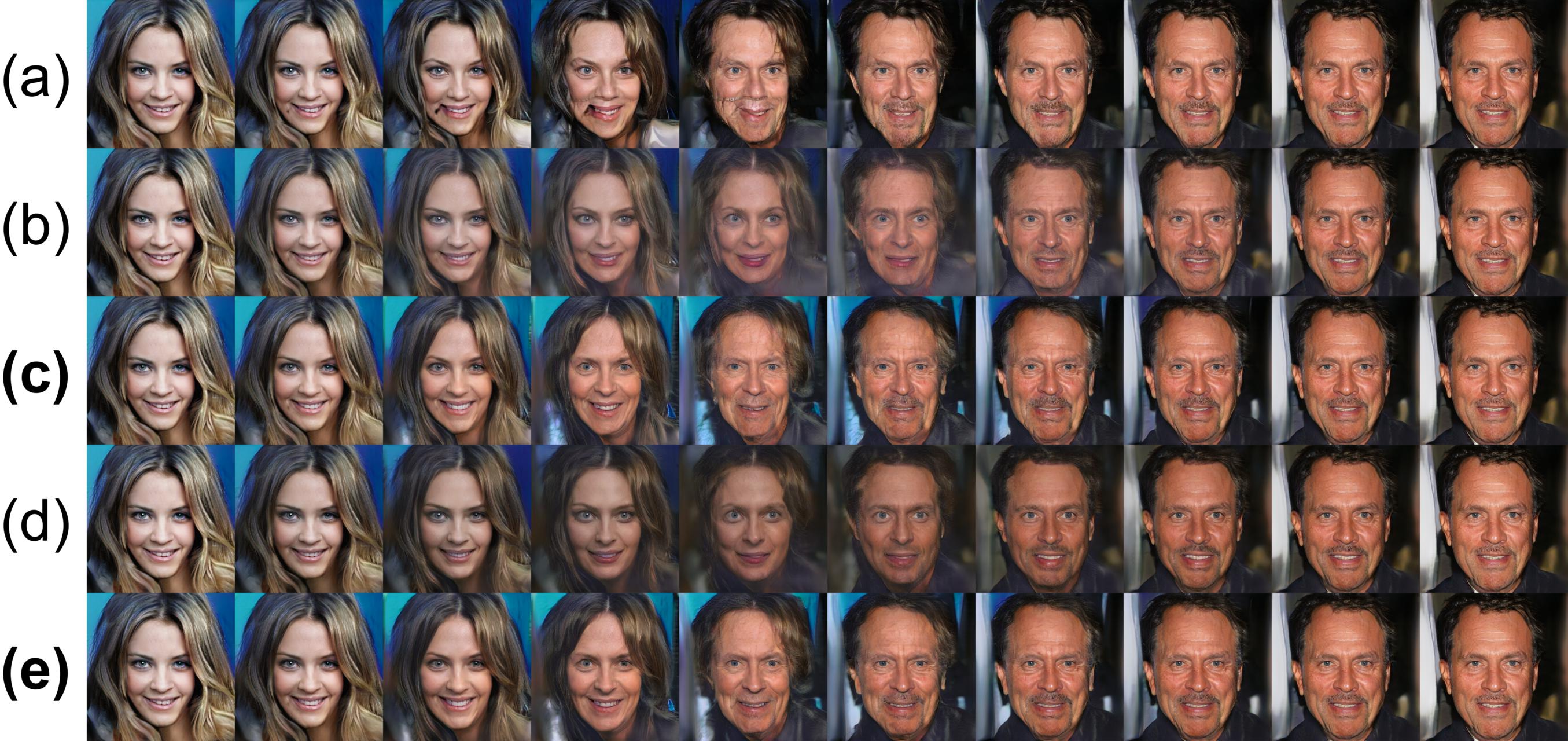}
%\caption{Comparison of (a) \textit{Linear}, (b) \textit{sqDiff}, (c) \textit{sqDiff+D}, (d) \textit{VGG} and (e) \textit{VGG+D} on CelebA.} 
\end{figure}

\begin{figure}[H]
\centering
  \includegraphics[width=0.8\linewidth]{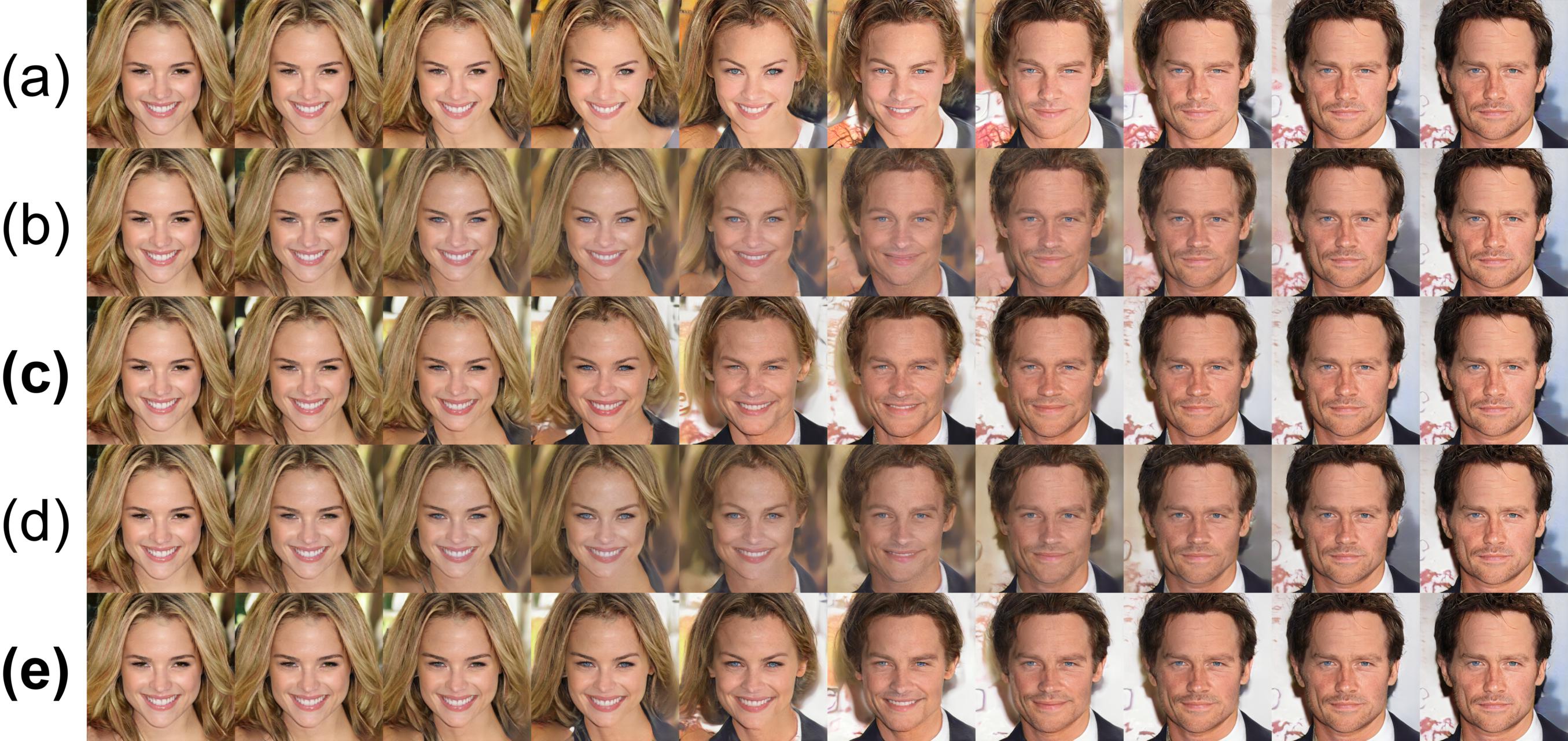}
%\caption{Comparison of (a) \textit{Linear}, (b) \textit{sqDiff}, (c) \textit{sqDiff+D}, (d) \textit{VGG} and (e) \textit{VGG+D} on CelebA.} 
\end{figure}

\newpage 
\paragraph{Squared differences along paths}
We show  mean squared differences 
$ \left |\left | G(\gamma(t_{i+1}))- G(\gamma(t_i)) \right |\right |^2_2 $
along 100 interpolation points for the twelve geodesics used for Figure~\ref{fig:criticAlongGeodesics}, confirming that those are minimized for the method \textit{sqDiff}. Similarly, \textit{VGG} leads to small squared differences in image space, explaining the perceived blurriness. While \textit{Linear} leads to large squared differences along the geodesic, our proposed methods \textit{sqDiff+D} and \textit{VGG+D} result in shorter paths (as measured by the squared differences in sample space). We explain the short distances close to the endpoints by the fact that, in the implementation of \cite{KarrasALL18}, latent vectors are projected onto a $512$-dimensional sphere of radius $\sqrt{512}$ before an application of the generator, so that larger distances on the sphere in latent space appear when coordinates cross the zero point.

\begin{figure}[H]
\centering
  \includegraphics[width=0.5\linewidth]{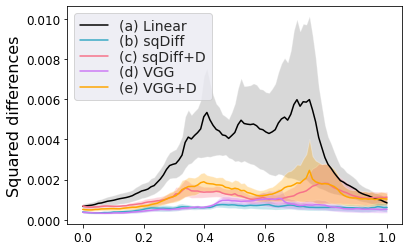}
\caption{Squared Differences along paths. Mean $\pm 1$ standard error of mean (12 paths - the same paths as were used in Figure~\ref{fig:criticAlongGeodesics}). The proposed methods \textit{sqDiff+D} and \textit{VGG+D} fall between the shortest possible path found by \textit{sqDiff} and the \textit{Linear} method.}
\end{figure}

\paragraph{Squared VGG feature differences along paths}
We show squared feature differences of the VGG19 network as a measure of distance covered along paths. Following \cite{Laine18}, we consider a weighted mean (by reciprocal of layer dimensions) of the layers \textit{conv\_1\_2},\textit{conv\_2\_2}, \textit{conv\_3\_2}, \textit{conv\_4\_2} and \textit{conv\_5\_2}. We find that \textit{VGG} minimizes this measure close to the endpoints, but obtains slightly larger values than \textit{sqDiff} in the middle of the path. In general, \textit{sqDiff} and \textit{VGG} are similar in distance, confirming visual evaluation that both result in blurry paths. Apparently, the way to minmize $||H(G(\gamma(t_i)))-H(G(\gamma(t_{i-1})))||$ in feature spaces defined by a continuous feature map $H$ is partially achieved by minimizing distances in image space $||G(\gamma(t_i))-G(\gamma(t_{i-1})||$, which is not entirely surprising. We also note that we do not train until convergence, which may explain the parts where \textit{VGG} obtains slightly larger values than \textit{sqDiff}. The comparison to the other curves is similar as above. \textit{Linear} covers the largest distance, while the proposed methods \textit{sqDiff+D} and \textit{VGG+D} settle in the middle.

\begin{figure}[H]
\centering
  \includegraphics[width=0.5\linewidth]{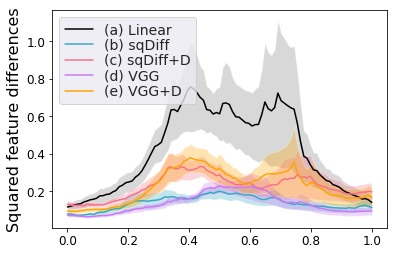}
\caption{Squared Differences in VGG feature spaces along paths. Mean $\pm 1$ standard error of mean (12 paths - the same paths as were used in Figure~\ref{fig:criticAlongGeodesics}).}
\end{figure}

\newpage
\subsubsection{Biased discriminators}\label{app:failure}

Starting from the model trained by \cite{KarrasALL18} and finetuning the discriminator only (on 2000 images) resulted in a critic that produced high quality images, but with a bias toward white people and bright illumination. All hyperparameter setting are unchanged from the images above in the following biased interpolations:

\begin{figure}[H]
\centering
  \includegraphics[width=0.8\linewidth]{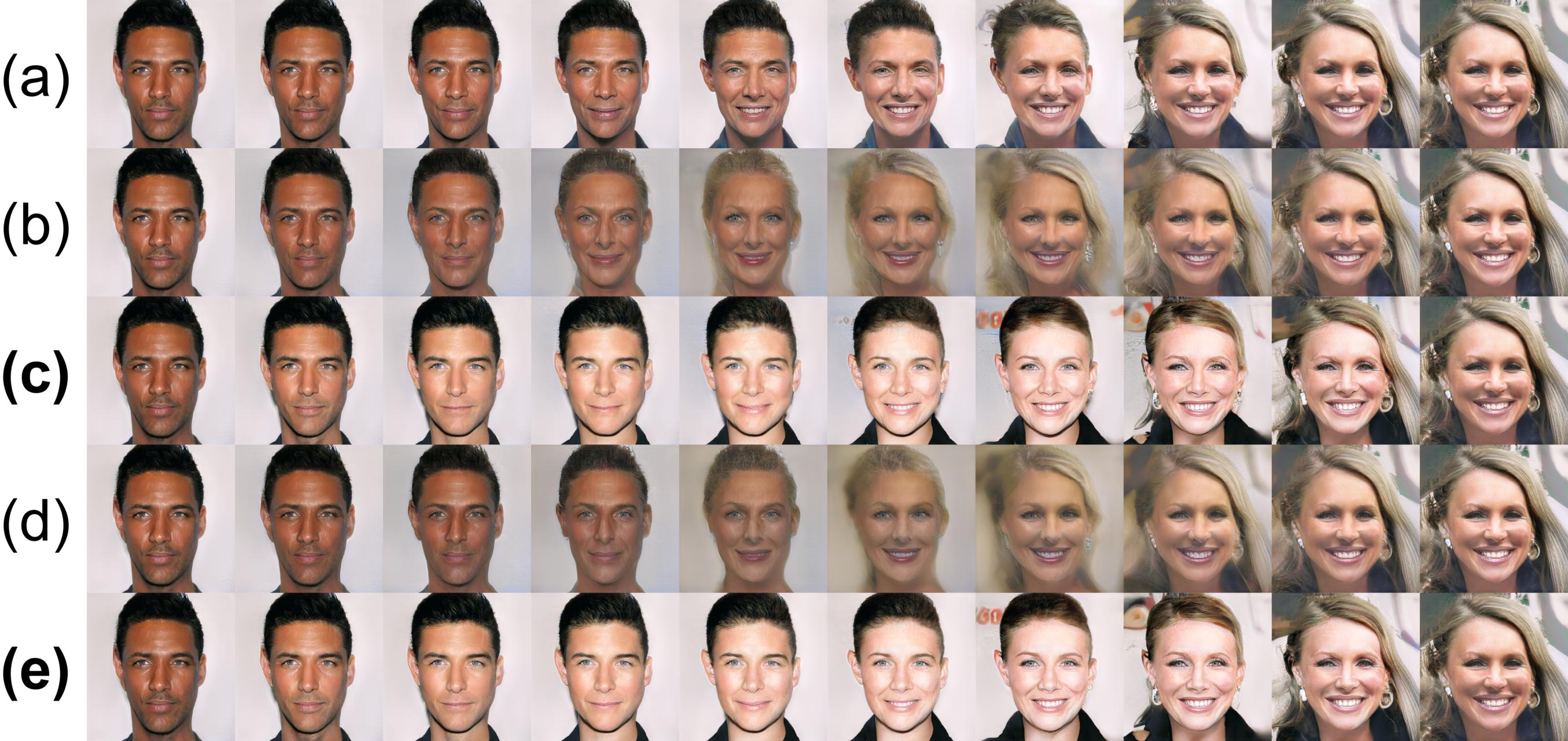}
%\caption{Comparison of (a) \textit{Linear}, (b) \textit{sqDiff}, (c) \textit{sqDiff+D}, (d) \textit{VGG} and (e) \textit{VGG+D} on CelebA.} 
\end{figure}

\begin{figure}[H]
\centering
  \includegraphics[width=0.8\linewidth]{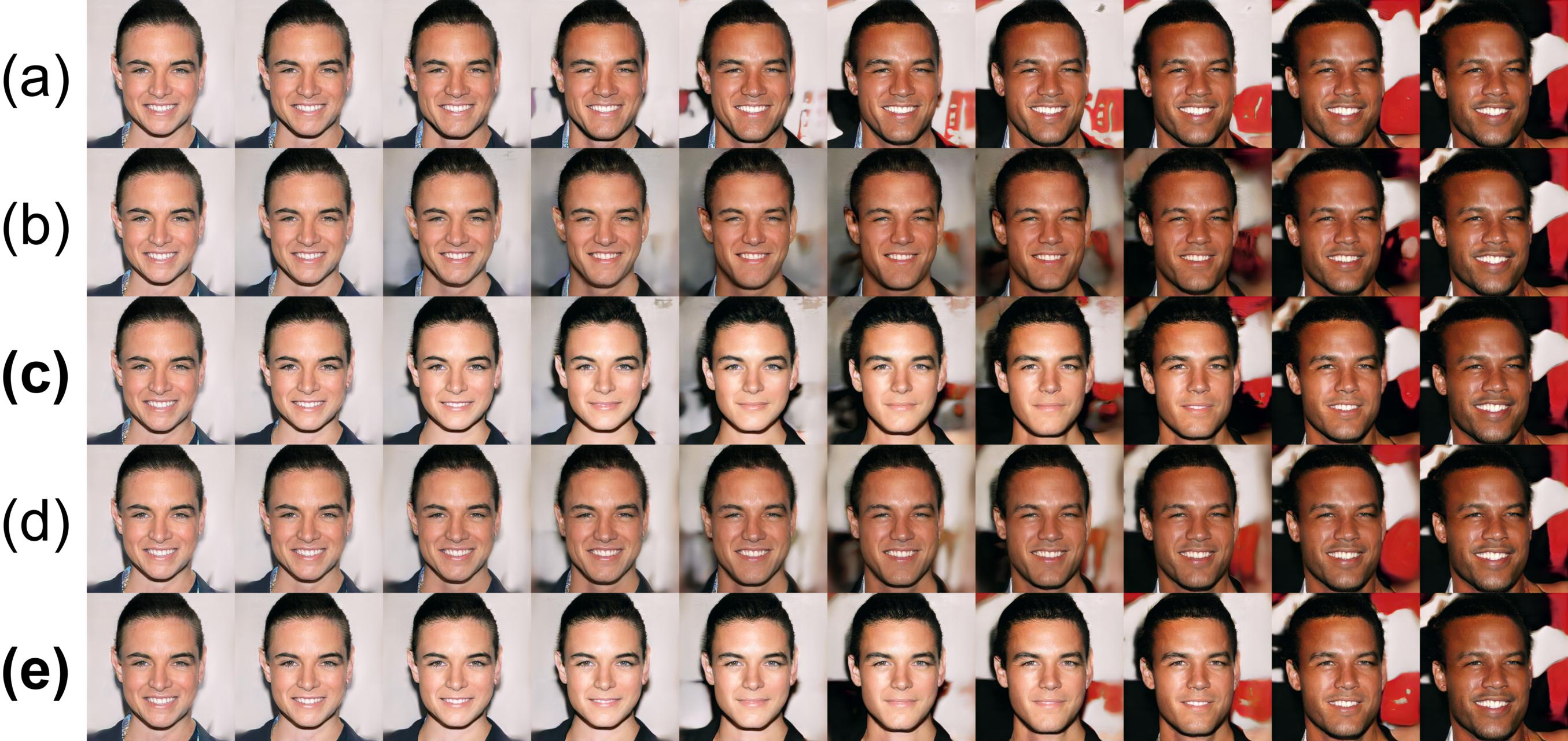}
%\caption{Comparison of (a) \textit{Linear}, (b) \textit{sqDiff}, (c) \textit{sqDiff+D}, (d) \textit{VGG} and (e) \textit{VGG+D} on CelebA.} 
\end{figure}

\newpage 
\subsubsection{Local perturbations in latent space}\label{app:localPerturbations}

\begin{figure*}[ht]
  \centering
  \includegraphics[width=.95\textwidth]{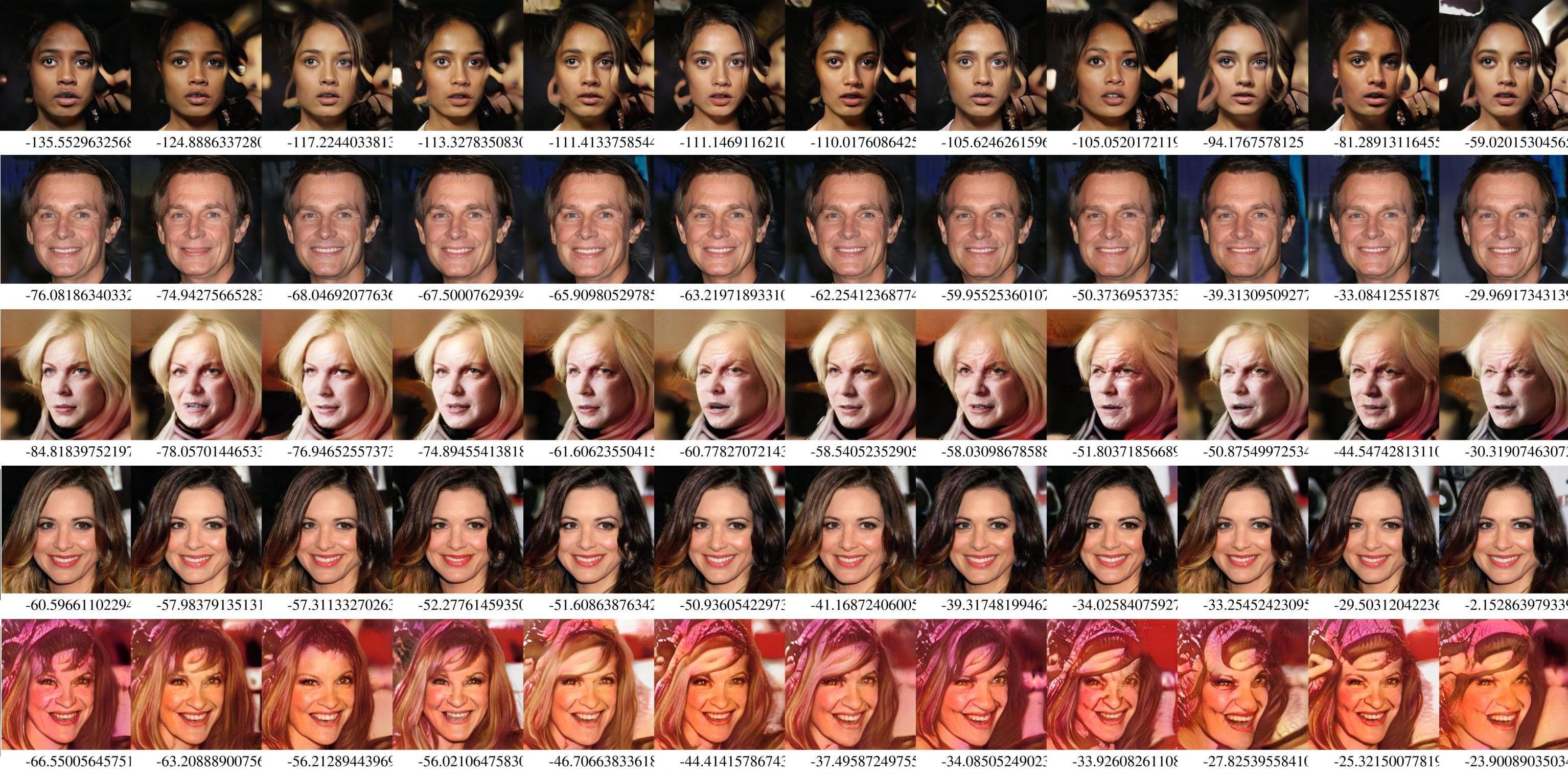}
\caption{Critic values of ProGAN on generated images from small perturbations of a fixed latent vector. While the generated samples do not change significantly, the critic values cover a wide range.}
\label{fig:localPerturbations}
\vspace{-3mm}
\end{figure*}

Figure~\ref{fig:localPerturbations} shows generated samples and their critic values. Each row corresponds to images generated from small, local perturbations to the latent vector. While the generated samples do not change significantly, the critic values cover a large part of the range of the critic values. This showcases that critic values on single images may not always provide useful information for improvements of the sample quality.

\newpage

\subsection{MNIST}
In the following figures, we illustrate some additional examples of trained paths for the ProGAN trained on MNIST using the same settings as above. Rows illustrate (a) \textit{Linear} on top, (b) \textit{sqDiff} in the middle, and (c) \textit{sqDiff+D} at the bottom.
\begin{figure}[H]
\centering
  \includegraphics[width=0.8\linewidth]{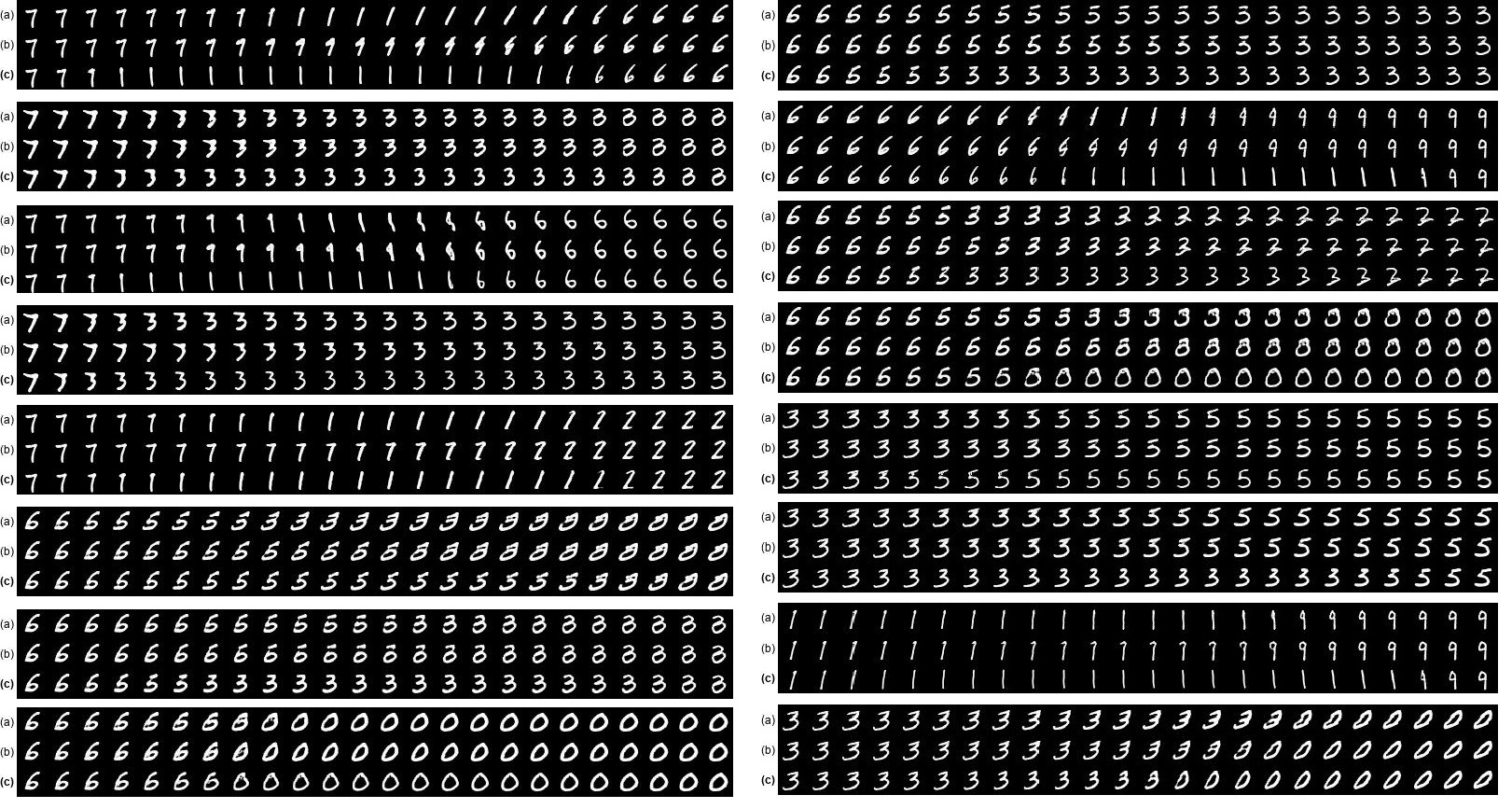}
\caption{Comparison of (a) \textit{Linear}, (b) \textit{sqDiff}, and (c) \textit{sqDiff+D} %, (d) \textit{VGG} and (e) \textit{VGG+D} 
on MNIST.
} 
\label{fig:supplMNIST}
\end{figure}

\begin{figure}[H]
\centering
  \includegraphics[width=0.8\linewidth]{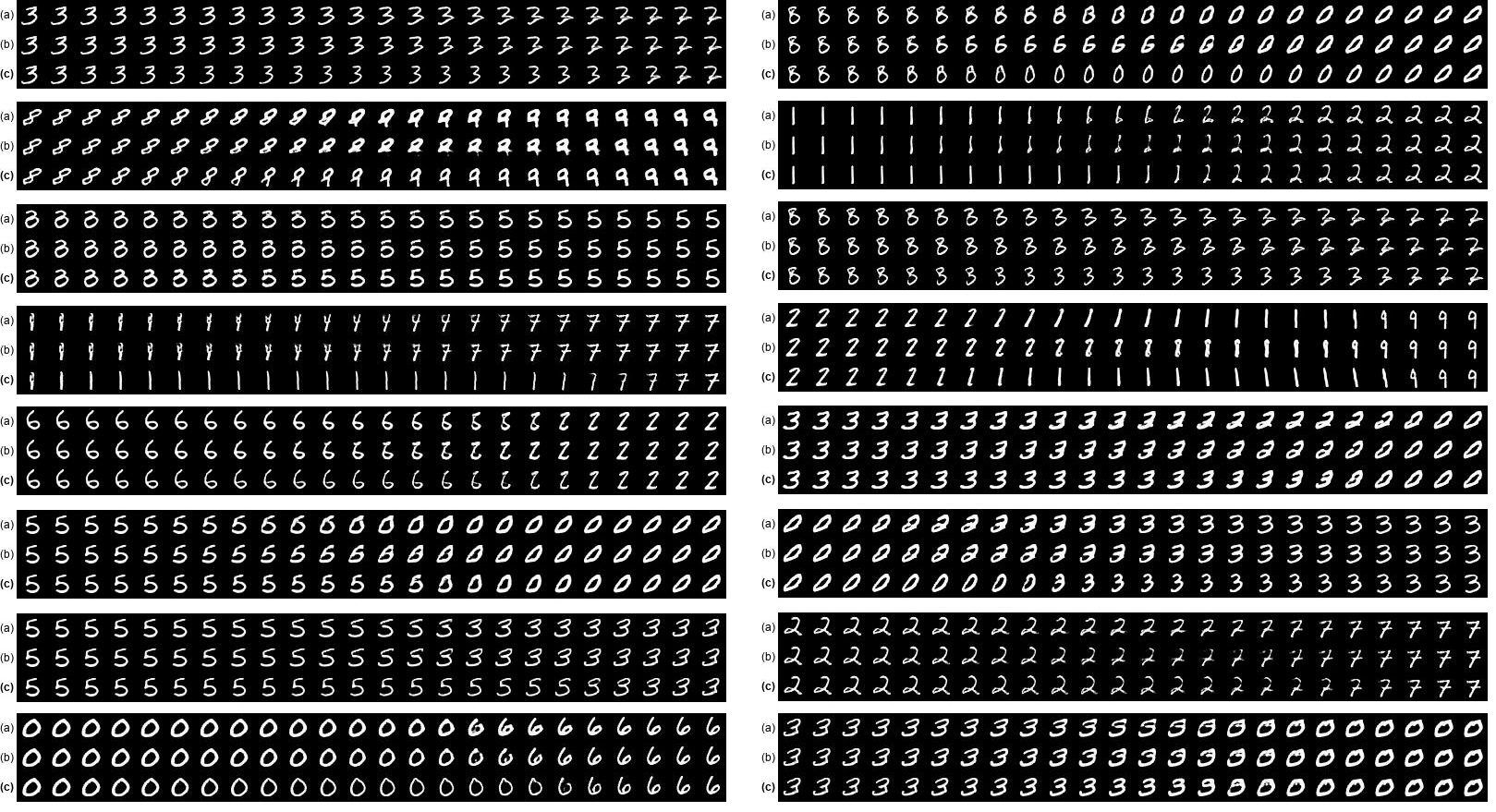}
\caption{Comparison of (a) \textit{Linear}, (b) \textit{sqDiff}, and (c) \textit{sqDiff+D} %, (d) \textit{VGG} and (e) \textit{VGG+D} 
on MNIST.
}
\label{fig:supplMNIST2}
\end{figure}

\newpage

\subsection{LSUN-Cars}
We additionally experimented on the LSUN-Car dataset. Both methods \textit{sqDiff} and \textit{VGG} to result in very blurry images on this dataset when trying to minimize the path length in sample space or VGG feature space. This is detrimental to our method that tries to improve upon the short path by incorporating critic values into the metric. We also found the \textit{Linear} method to work well when the endpoints were chosen similarly, but otherwise linear fails to find a good interpolations. Whenever linear failed, we were not able to find short and realistic paths using the proposed method \textit{VGG+D}, but the experiments show the possibility to improve the image quality significantly over \textit{VGG} by using discriminator information. This functionality is demonstrated by the following two Figures, where the first choice of hyperparameter $\lambda$ balances shortest path and discriminator information, while the second puts emphasis on discriminator information leaving the shortest path significantly.

\begin{figure}[H]
\centering
  \includegraphics[width=0.8\linewidth]{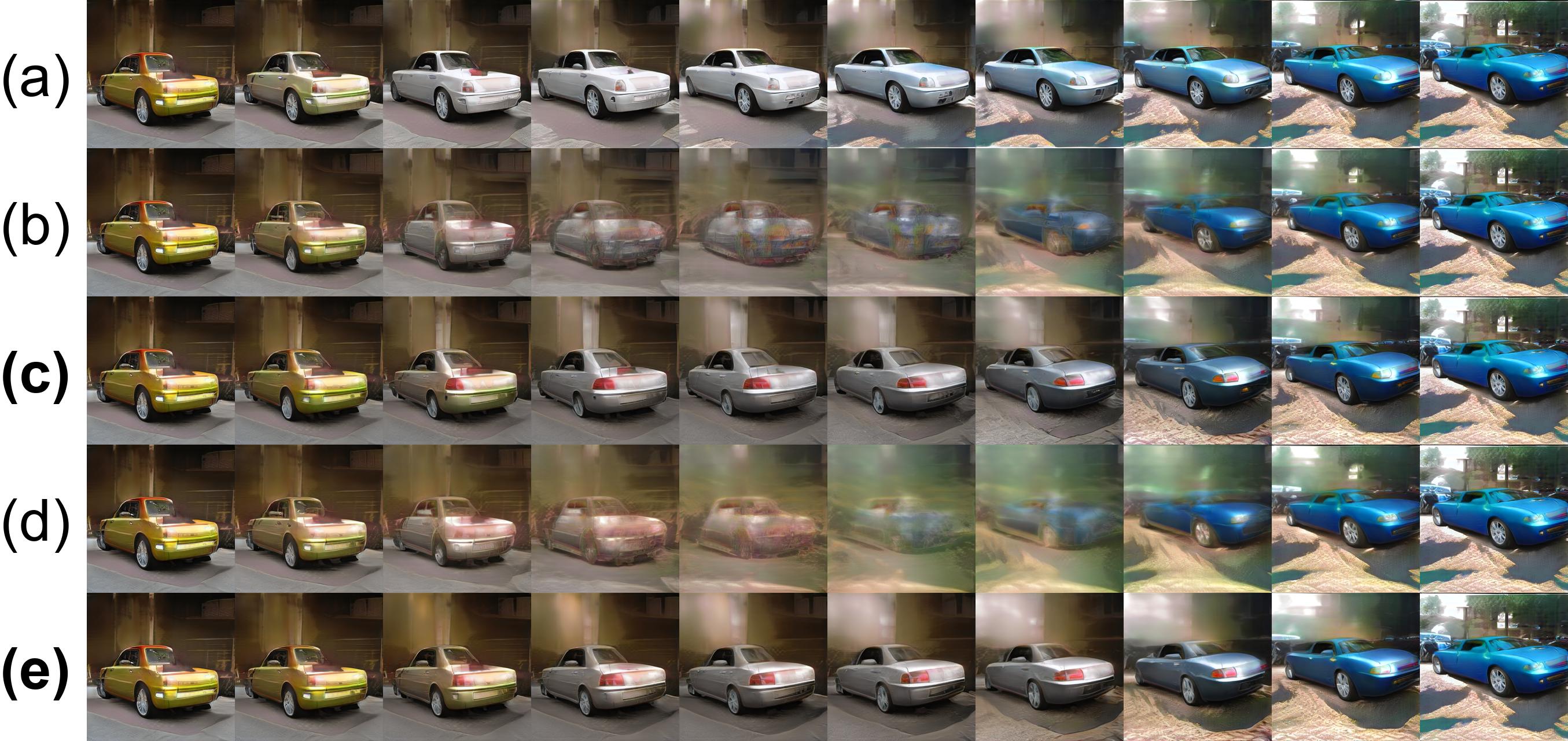}
\caption{Comparison of (a) \textit{Linear}, (b) \textit{sqDiff}, (c) \textit{sqDiff+D}, (d) \textit{VGG} and (e) \textit{VGG+D} on MNIST. \textit{sqDiff} and \textit{VGG} result in blurry images. Enforcing high discriminator in (c) and (e) improves image quality. } 
\label{fig:supplCar}
\end{figure}

\begin{figure}[H]
\centering
  \includegraphics[width=0.8\linewidth]{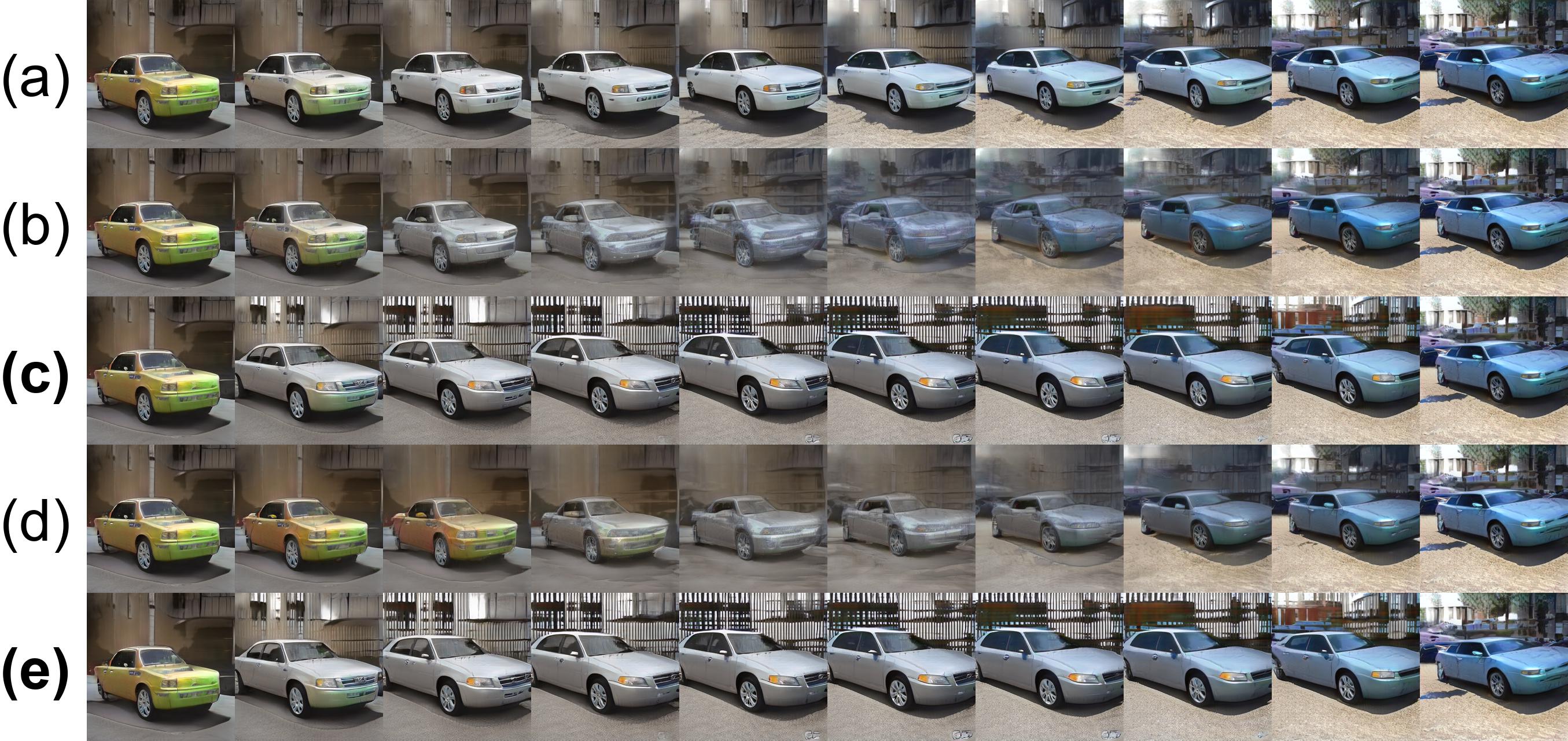}
\caption{Comparison of (a) \textit{Linear}, (b) \textit{sqDiff}, (c) \textit{sqDiff+D}, (d) \textit{VGG} and (e) \textit{VGG+D} on MNIST. The same endpoints as in Fig.~\ref{fig:supplCar}, but with larger hyperparameter $\lambda$ putting more emphasis on high discriminator values than the objective to find shortest paths. While image quality improves, the path deviates from short paths significantly.}
\label{fig:supplCar2}
\end{figure}

\newpage
More examples comparing (a) \textit{Linear}, (d) \textit{VGG} and (e) \textit{VGG+D}.

\begin{figure}[H]
\centering
  \includegraphics[width=0.78\linewidth]{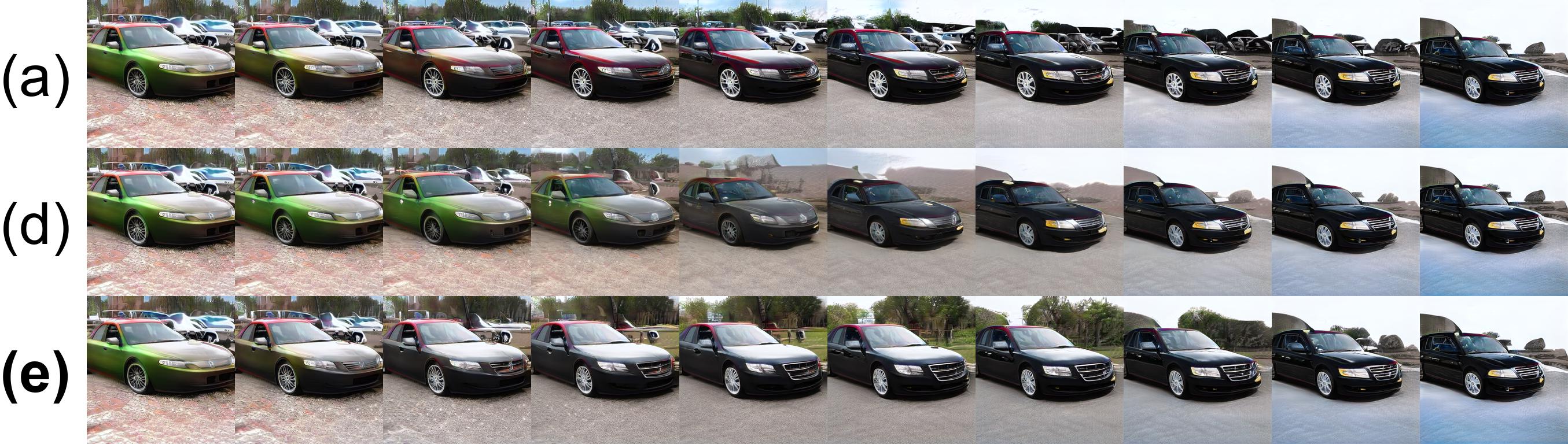}
  \vspace{0.3cm}
\label{fig:supplCar3}
  \includegraphics[width=0.78\linewidth]{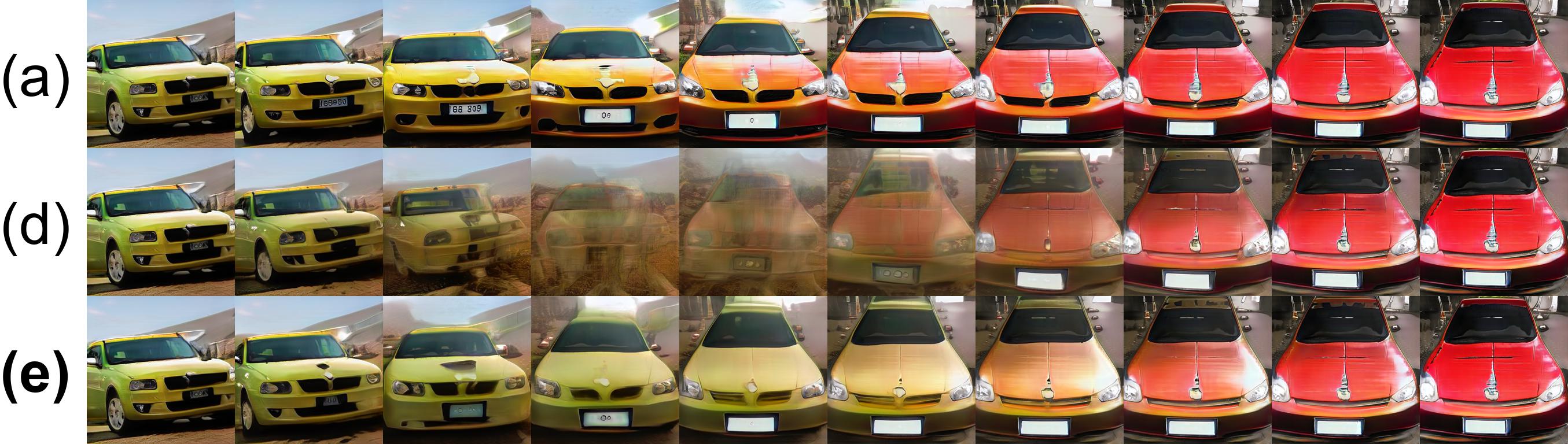}
\vspace{0.3cm}
\includegraphics[width=0.78\linewidth]{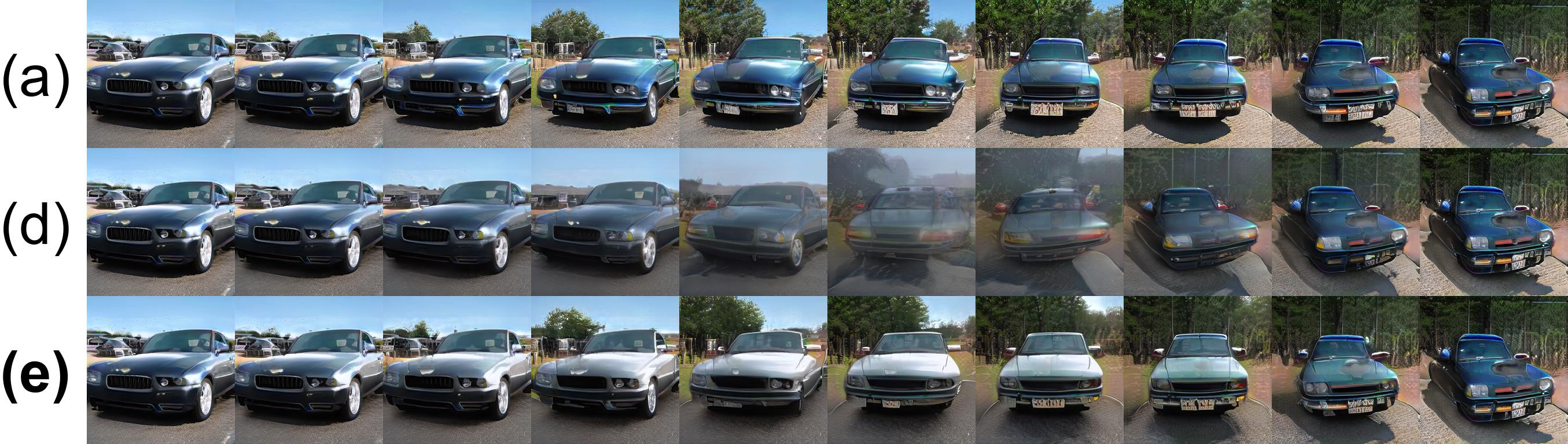}
\vspace{0.3cm}
\includegraphics[width=0.78\linewidth]{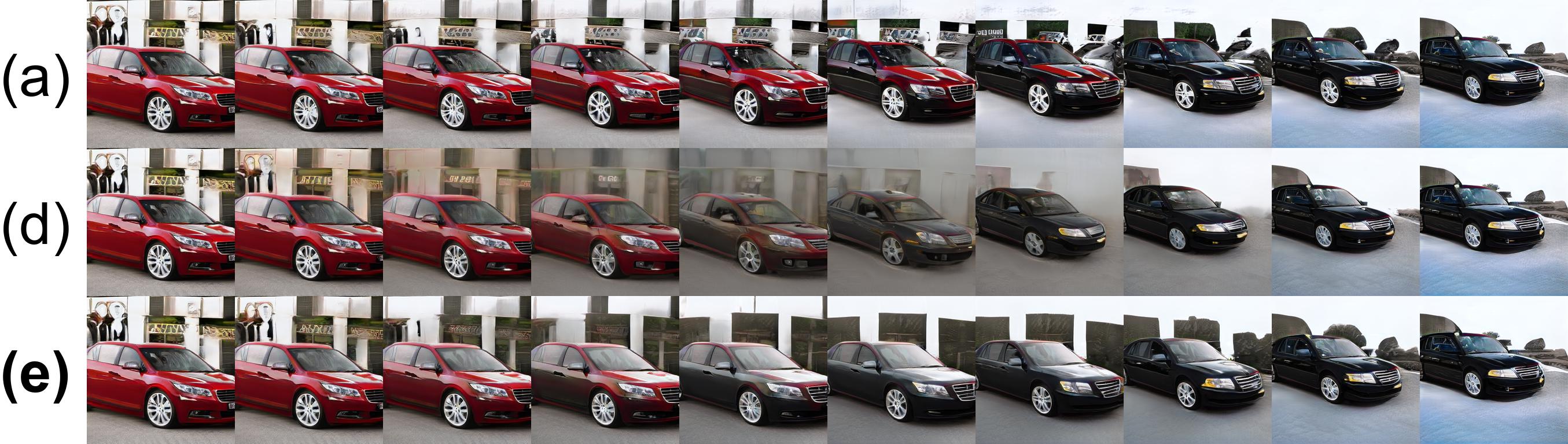}
\vspace{0.3cm}
\includegraphics[width=0.78\linewidth]{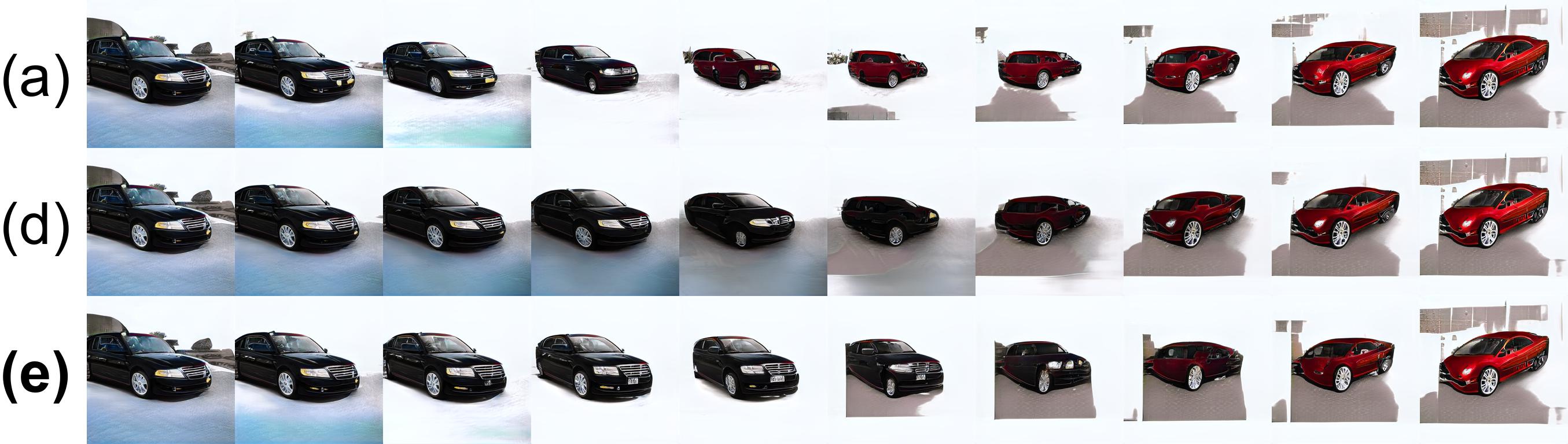}
\end{figure}

\newpage

\subsection{LSUN-Bedrooms}

For the LSUN dataset of bedrooms, we exemplify here that our proposed method again recovers sharp images from blurry images along paths using \textit{VGG}, and that the paths found by \textit{VGG+D} can be more direct (see in particular the change of color in the third exaple).

\begin{figure}[H]
\centering
  \includegraphics[width=0.8\linewidth]{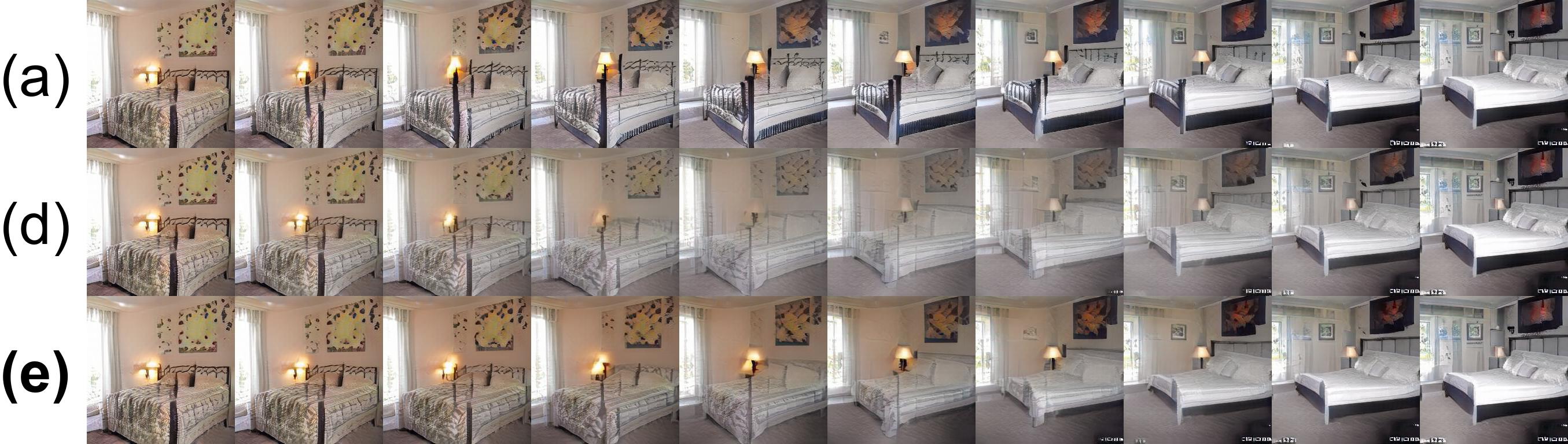}
  \vspace{0.5cm}
\label{fig:supplBed}
  \includegraphics[width=0.8\linewidth]{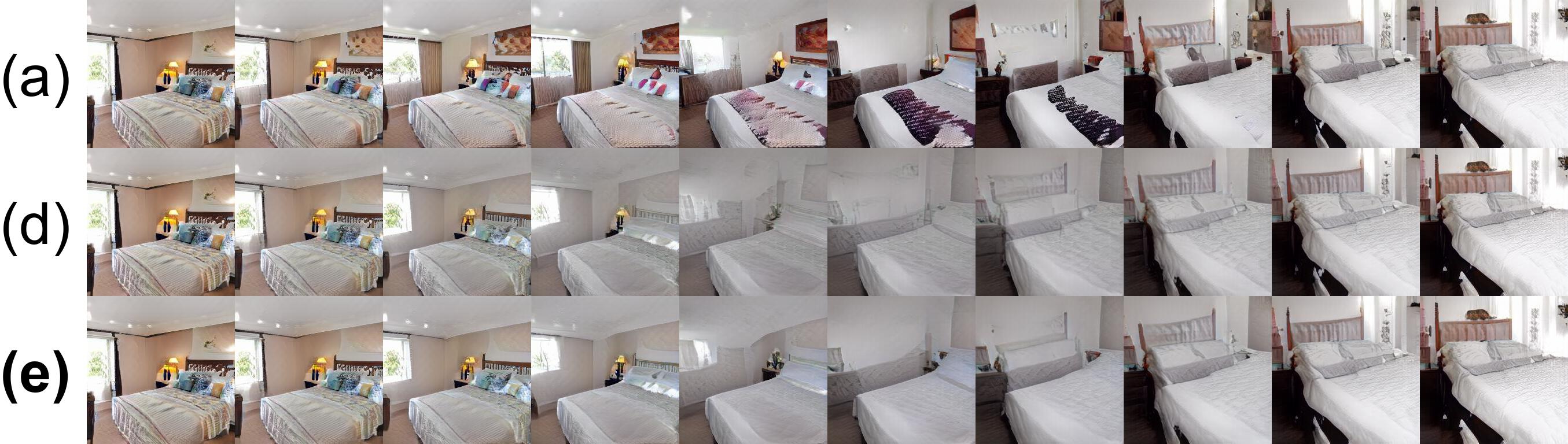}
\vspace{0.5cm}
\includegraphics[width=0.8\linewidth]{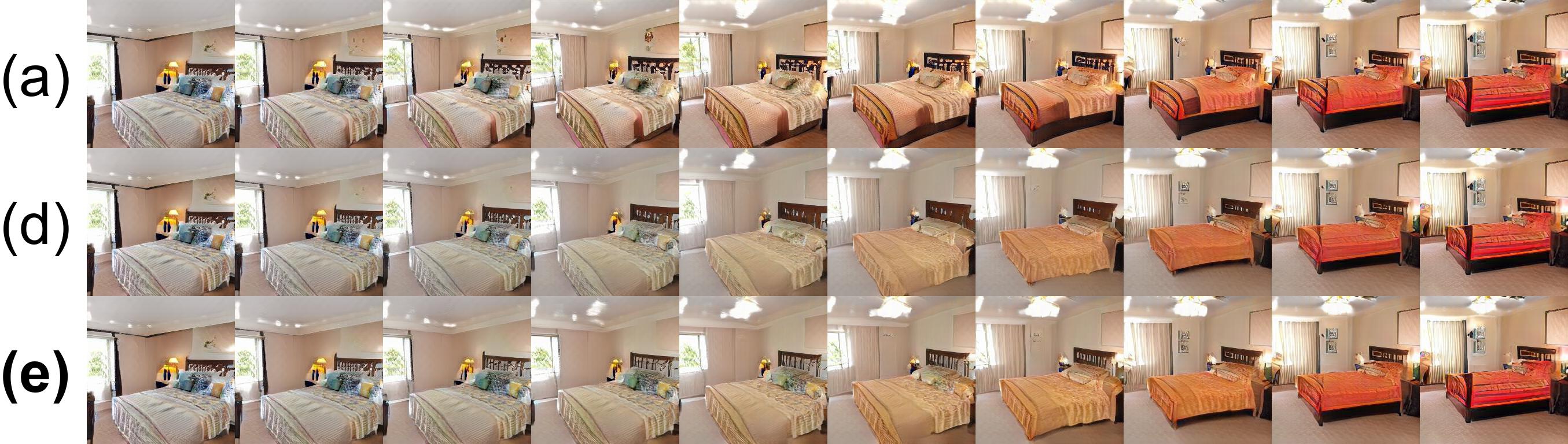}
\vspace{0.5cm}
\includegraphics[width=0.8\linewidth]{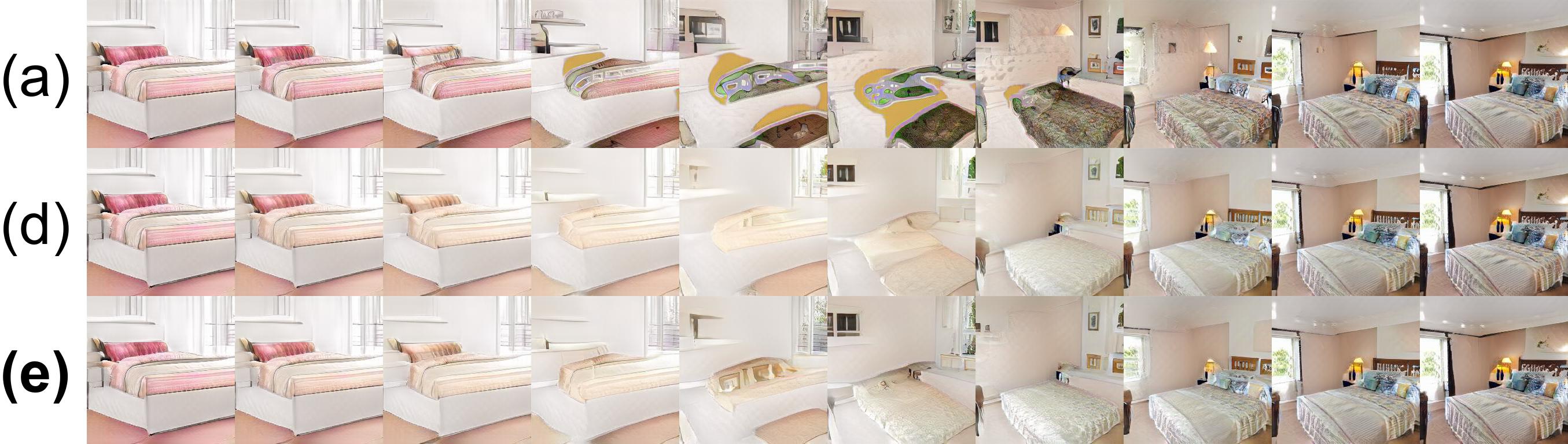}
\caption{Comparison of (a) \textit{Linear}, (d) \textit{VGG} and the proposed method (e) \textit{VGG+D} on LSUN-bedrooms.}
\end{figure}

\end{document}